\documentclass[acmsmall]{acmart}

\AtBeginDocument{%
  }

\usepackage{amsthm}
\DeclareMathOperator*{\argmax}{arg\,max}
\DeclareMathOperator*{\argmin}{arg\,min}
\usepackage{enumitem}
\usepackage{adjustbox}
\usepackage{makecell}
\usepackage{multirow}
\usepackage{tikz}
\theoremstyle{definition} 

\usepackage[edges]{forest}
\usepackage{pgfplots}
\usepackage{tabularx}
\usepackage{colortbl}
\usepackage{url}
\usepackage[most]{tcolorbox}
\usepackage {soul, color}
\definecolor{BBlue}{RGB}{157,195,230}
\definecolor{YYellow}{RGB}{255,217,102}
\definecolor{GGreen}{RGB}{169,209,142}
\definecolor{OOrange}{RGB}{244,177,131}
\definecolor{GGray}{RGB}{166,166,166}
\definecolor{PPink}{RGB}{255,166,166}
\usepackage{xspace}
\newcommand{\etal}{\emph{et al.}\xspace}
\newcommand{\eg}{\emph{e.g.,}\xspace}
\newcommand{\ie}{\emph{i.e.,}\xspace}
\newcommand{\etc}{\emph{etc.}\xspace}

\begin{document}

\title{Efficient Prompting Methods for Large Language Models: A Survey}

\author{Kaiyan Chang}
\email{changkaiyan\_neu@outlook.com}
\affiliation{%
  \institution{Northeastern University}
  \streetaddress{NO. 195, Chuangxin Road, Hunnan District}
  \city{Shenyang}
  \state{Liaoning}
  \country{China}
  \postcode{110819}
}

\author{Songcheng Xu}
\email{winsome.xsc@gmail.com}
\affiliation{%
  \institution{Northeastern University}
  \streetaddress{NO. 195, Chuangxin Road, Hunnan District}
  \city{Shenyang}
  \state{Liaoning}
  \country{China}
  \postcode{110819}
}

\author{Chenglong Wang}
\affiliation{%
  \institution{Northeastern University}
  \streetaddress{NO. 195, Chuangxin Road, Hunnan District}
  \city{Shenyang}
  \state{Liaoning}
  \country{China}
  \postcode{110819}
}
\author{Yingfeng Luo}
\affiliation{%
  \institution{Northeastern University}
  \streetaddress{NO. 195, Chuangxin Road, Hunnan District}
  \city{Shenyang}
  \state{Liaoning}
  \country{China}
  \postcode{110819}
}
\author{Xiaoqian Liu}
\affiliation{%
  \institution{Northeastern University}
  \streetaddress{NO. 195, Chuangxin Road, Hunnan District}
  \city{Shenyang}
  \state{Liaoning}
  \country{China}
  \postcode{110819}
}

\author{Tong Xiao}\authornote{Corresponding Author.}
\email{xiaotong@mail.neu.edu.cn}
\affiliation{%
  \institution{Northeastern University \& NiuTrans Research}
  \streetaddress{NO. 195, Chuangxin Road, Hunnan District}
  \city{Shenyang}
  \state{Liaoning}
  \country{China}
  \postcode{110819}
}
\author{Jingbo Zhu}
\email{zhujingbo@mail.neu.edu.cn}
\affiliation{%
  \institution{Northeastern University \& NiuTrans Research}
  \streetaddress{NO. 195, Chuangxin Road, Hunnan District}
  \city{Shenyang}
  \state{Liaoning}
  \country{China}
  \postcode{110819}
}

\renewcommand{\shortauthors}{Kaiyan Chang \etal}

\begin{abstract}
Prompting is a mainstream paradigm for adapting large language models to specific natural language processing tasks without modifying internal parameters.
Therefore, detailed supplementary knowledge needs to be integrated into external prompts, which inevitably brings extra human efforts and computational burdens for practical applications.
As an effective solution to mitigate resource consumption, Efficient Prompting Methods have attracted a wide range of attention.
We provide mathematical expressions at a high level to deeply discuss \textit{Automatic Prompt Engineering} for different prompt components and \textit{Prompt Compression} in continuous and discrete spaces.
Finally, we highlight promising future directions to inspire researchers interested in this field.
\end{abstract}




\maketitle

\section{Introduction}
\tikzstyle{my-box}=[
    rectangle,
    draw=GGray!50,
    rounded corners,
    text opacity=1,
    minimum height=1.5em,
    minimum width=5em,
    inner sep=2pt,
    align=center,
    fill opacity=.5,
    line width=1pt,
]
\tikzstyle{Ins_leaf}=[my-box, minimum height=1.5em,
    fill=BBlue!25, text=black,
    align=left,font=\normalsize,
    inner xsep=2pt,
    inner ysep=4pt,
    line width=1pt,
]
\tikzstyle{CoT_leaf}=[my-box, minimum height=1.5em,
    fill=YYellow!25, text=black,
    align=left,font=\normalsize,
    inner xsep=2pt,
    inner ysep=4pt,
    line width=1pt,
]
\tikzstyle{T2V_leaf}=[my-box, minimum height=1.5em,
    fill=OOrange!25, text=black,
    align=left,font=\normalsize,
    inner xsep=2pt,
    inner ysep=4pt,
    line width=1pt,
]
\tikzstyle{T2T_leaf}=[my-box, minimum height=1.5em,
    fill=GGreen!25, text=black,
    align=left,font=\normalsize,
    inner xsep=2pt,
    inner ysep=4pt,
    line width=1pt,
]
\begin{figure*}[t]
    \centering
    \resizebox{1.0\textwidth}{!}{
        \begin{forest}
            forked edges,
            for tree={
                grow'=0,
                draw,
                reversed=true,
                anchor=base west,
                parent anchor=east,
                child anchor=west,
                base=left,
                font=\large,
                rectangle,
                rounded corners,
                align=left,
                minimum width=4em,
                edge+={black, line width=1pt},
                s sep=4pt,
                inner xsep=2pt,
                inner ysep=3pt,
                line width=1pt,
                ver/.style={rotate=90, child anchor=north, parent anchor=south, anchor=center},
            },
            where level=1{text width=7.7em,font=\normalsize,}{},
            where level=2{text width=7.7em,font=\normalsize,}{},
            where level=3{text width=6.2em,font=\normalsize,}{},
            where level=4{text width=30em,font=\normalsize,}{},
		[
                Efficient Prompting Methods, ver, color=PPink!100, fill=PPink!25, text=black
                [
                    Automatic Prompt\\Engineering (\S \ref{Automatic_prompt_engineering}), color=BBlue!100, fill=BBlue!50, text=black
                    [
                        Instruction Design, color=BBlue!100, fill=BBlue!40, text=black
                        [
                            Sampling, color=BBlue!100, fill=BBlue!30, text=black
                            [
                                Self-instruct~\cite{wang2022self}{, }APE~\cite{Zhou2022LargeLM}{, }GPO~\cite{Li2023RobustPO}{, }OPRO~\cite{Yang2023LargeLM}{, }\\PromptAgent~\cite{Wang2023PromptAgentSP}{, }MoP~\cite{wang2024one}{, }GPS~\cite{xu2022gps}{, }EvoPrompting~\cite{Chen2023EvoPromptingLM}{, }\\EvoPrompt~\cite{Guo2023ConnectingLL}{, }Promptbreeder~\cite{Fernando2023PromptbreederSS}{, }AELP~\cite{hsieh2023automatic}{, }PhaseEvo~\cite{cui2024phaseevo}
                                , Ins_leaf
                            ]
                        ]
                        [
                            Feedback, color=BBlue!100, fill=BBlue!30, text=black
                            [
                                RLPrompt~\cite{Deng2022RLPromptOD}{, }DSP~\cite{li2024guiding}{, }PACE~\cite{dong2023pace}{, }PRewrite~\cite{kong2024prewrite}{, }SCULPT~\cite{kumar2024sculpt}{, }\\ProTeGi~\cite{Pryzant2023AutomaticPO}{, }PE2~\cite{ye2023prompt}{, }PREFER~\cite{zhang2024prefer}{, }AutoHint~\cite{sun2023autohint}{, }UniPrompt~\cite{juneja2024task}{, }\\GPO~\cite{tang2024unleashing}{, }AMPO~\cite{yang2024ampo}{, }APOHF~\cite{lin2024prompt}{, }BPO~\cite{Cheng2023BlackBoxPO}{, }APEER~\cite{jin2024apeer}{, }FIPO~\cite{lu2024fipo}
                                , Ins_leaf
                            ]
                        ]
                        [
                            Editing, color=BBlue!100, fill=BBlue!30, text=black
                            [
                                GrIPS~\cite{Prasad2022GrIPSGE}{, }Plum~\cite{Pan2023PlumPL}{, }SPRIG~\cite{zhang2024sprig}{, }TEMPERA~\cite{Zhang2022TEMPERATP}
                                , Ins_leaf
                            ]
                        ]
                    ]
                    [
                        Chain-of-Thought\\Optimization, color=YYellow!100, fill=YYellow!40, text=black
                        [
                            Sampling, color=YYellow!100, fill=YYellow!30, text=black
                            [
                                LMSI~\cite{huang2022large}{, }Auto-CoT~\cite{Zhang2022AutomaticCO}{, }Boosted Prompting~\cite{pitis2023boosted}{, }COSP~\cite{Wan2023BetterZR}{, }\\USP~\cite{wan2023universal}{, }Meta-CoT~\cite{Zou2023MetaCoTGC}{, }Reprompting~\cite{Xu2023RepromptingAC}
                                , CoT_leaf
                            ]
                        ]
                        [
                            Feedback, color=YYellow!100, fill=YYellow!30, text=black
                            [
                                Self-refine~\cite{madaan2024self}{, }PromptPG~\cite{lu2022dynamic}{, }Prompt-OIRL~\cite{sun2023query}{, }Reflexion~\cite{Shinn2023ReflexionLA}{, }\\PROMST~\cite{Chen2024PRomptOI}{, }DTG~\cite{Li2023DeliberateTG}{, }Reprompt~\cite{Chen2024RePromptPB}
                                 , CoT_leaf
                            ]
                        ]
                        [
                            Interaction, color=YYellow!100, fill=YYellow!30, text=black
                            [
                                ReAct~\cite{Yao2022ReActSR}{, }Verify-and-Edit~\cite{zhao2023verify}{, }ART~\cite{Paranjape2023ARTAM}{, }Self-ask~\cite{press2022measuring}{, }\\ToolLLM~\cite{qin2023toolllm}{, }ATC~\cite{shi2024chain}
                                , CoT_leaf
                            ]
                        ]
                    ]  
                ]
                [
                    Prompt\\Compression(\S \ref{Prompt_compression}), color=GGreen!100, fill=GGreen!50, text=black
                    [
                        Text-to-Vector\\Compression, color=OOrange!100, fill=OOrange!40, text=black
                        [
                            Internalization, color=OOrange!100, fill=OOrange!30, text=black
                            [
                                Context Distillation~\cite{Askell2021AGL}{, }Prompt Injection~\cite{Choi2022PromptIP}{, }Distilling Context~\cite{Snell2022LearningBD}{, }\\Instruction Distillation~\cite{Sun2023InstructionDM}{, }Distilling Step-by-Step~\cite{Hsieh2023DistillingSO}{, }xRAG~\cite{Cheng2024xRAGEC}
                                , T2V_leaf
                            ]
                        ]
                        [
                            Encoding, color=OOrange!100, fill=OOrange!30, text=black
                            [
                                Prompt Compression~\cite{Wingate2022PromptCA}{, }Gist~\cite{Mu2023LearningTC}{, }Gist-COCO~\cite{Li2024SayMW}{, }UltraGist~\cite{Zhang2024CompressingLC}{, }\\AutoCompressor~\cite{Chevalier2023AdaptingLM}{, }LLoCO~\cite{Tan2024LLoCOLL}{, }ICAE~\cite{Ge2023IncontextAF}{, }500xCompressor~\cite{Li2024500xCompressorGP}{, }\\POD~\cite{Li2023PromptDF}{, }RDRec~\cite
                                {Wang2024RDRecRD}{, }SelfCP~\cite{Gao2024SelfCPCO}
                                , T2V_leaf
                            ]
                        ]
                    ]
                    [
                        Text-to-Text\\Compression, color=GGreen!100, fill=GGreen!40, text=black
                        [
                            Pruning, color=GGreen!100, fill=GGreen!30, text=black
                            [
                                DynaICL~\cite{Zhou2023EfficientPV}{, }FilCo~\cite{Wang2023LearningTF}{, }CPC~\cite{Liskavets2024PromptCW}{, }AdaComp~\cite{zhang2024adacomp}{, }\\LLMLingua~\cite{Jiang2023LLMLinguaCP}{, }LongLLMLingua~\cite{Jiang2023LongLLMLinguaAA}{, }CoT-Influx~\cite{Huang2023FewerIM}{, }\\Selective Context~\cite{Li2023CompressingCT}{, }PROMPT-SAW~\cite{Ali2024PROMPTSAWLR}{, }PCRL~\cite{Jung2023DiscretePC}{, }LLMLinga-2~\cite{Pan2024LLMLingua2DD}
                                , T2T_leaf
                            ]
                        ]
                        [
                            Summarization, color=GGreen!100, fill=GGreen!30, text=black
                            [
                                RECOMP~\cite{Xu2023RECOMPIR}{, }Nano-Capsulator~\cite{Chuang2024LearningTC}{, }MEMWALKER~\cite{Chen2023WalkingDT}{, }\\CompAct~\cite{Yoon2024CompActCR}{, }Style-Compress~\cite{pu2024style}
                                , T2T_leaf
                            ]
                        ]
                    ]
                ]
		]
        \end{forest}
    }
    \caption{Taxonomy of efficient prompting methods.}
    \label{Fig_Taxonomy}
\end{figure*}



As hundreds of billions of breakthroughs on the parameter scale, Large Language Models (LLMs) acquire emergent abilities~\cite{Wei2022EmergentAO}, especially in-context learning ability~\cite{brown2020language} that promote rapid advancement in prompting techniques. 
Prompting stands out as a lightweight promising solution for controlling LLMs without tuning parameters, having received widespread attention within the Natural Language Processing (NLP) community. 
There are two types of prompts, where the hard prompt is discrete natural language descriptions and the soft prompt is continuous vector representations.
In particular, the hard prompt has become a crucial bridge for human-machine interaction relying on its improved interpretability, controllability and flexibility compared to the soft prompt.
Therefore, we mainly focus on the hard prompt that covers all the components of LLM input scaling from concise instructions to long context with demonstrations (Chain-of-Thought (CoT)~\cite{Wei2022ChainOT}, role-playing system prompts, \etc).

At present, it is common to unlock the potential of LLMs in specific domains by prompting. For example, the CoT series of studies~\cite{Yao2023TreeOT, Besta2023GraphOT, Chen2022ProgramOT} have progressively enhanced LLM reasoning capability by thinking aloud. Furthermore, OpenAI recently introduced the reasoning LLM o1~\cite{o1, o1-mini} trained with reinforcement learning to break down more difficult problems and produce a long internal CoT before responding. The excellent performance of these methods promotes increasing research on optimized prompting methods, which has gradually formed a brand new area in the NLP landscape. Meanwhile, several challenges related to application efficiency come one after another: more and more complex prompt design makes manual prompt optimization time-consuming and labor-intensive; more and more detailed prompt content inevitably consumes significant computational resources when applied to large-scale models. Such prohibitive overheads have become a major barrier to the practical deployment of LLMs, so we define ``Efficient Prompting Methods'' as prompting language models to achieve comparable or even better performance with fewer human or computational resources in this paper.

We narrow this survey to efficient prompting methods in the era of LLMs. To the best of our knowledge, this is the \textit{\textbf{first survey}} to summarize LLM prompting methods from the point of ``Efficient''. It is remarkable that we model the core concepts of each category of resource-saving methods from a \textit{\textbf{mathematical perspective}} in \S\ref{Mathematical_modeling}. Following this, we propose a \textit{\textbf{novel taxonomy}} as shown in Fig. \ref{Fig_Taxonomy} to comprehensively review efficient prompting methods based on their consistent optimization strategies. To avoid human resources, we introduce automatic prompt engineering efforts based on LLM empowerment in \S\ref{Automatic_prompt_engineering}, including iterative design and optimization of different prompt components. To save computational resources, we organize prompt compression efforts into two categories based on prompt types in \S\ref{Prompt_compression}: Text-to-Vector (T2V) compression in continuous space and Text-to-Text (T2T) compression in discrete space. Additionally, we provide sufficient schematic diagrams depicting the basic pipeline of each category of methods, as well as tables representing differences in the details of the same category of methods. There is also a list of open-source projects in Appendix \ref{Open_Resources} as a quick access for NLP practitioners efficiently prompting LLMs in both scientific research and commercial deployment. Finally, we analyze the challenges of existing methods and discuss promising \textit{\textbf{future directions}} in \S\ref{Future_directions}. We hope this survey can provide a clear picture of the efficient prompting topic and contribute to convenient human-machine interaction in the progress of Artificial General Intelligence (AGI).
\section{Preliminary}
\subsection{Origin of Prompting}
There have been two significant paradigm shifts from \textit{supervised learning} to \textit{fine-tuning} to \textit{prompting} in the field of NLP, where the ``paradigm'' refers to the way to \textit{train} a language model (LM) and \textit{apply} it to specific downstream tasks. Currently, prompting has become the dominant paradigm for LLMs. In the following, we elicit the origin of prompting in the context of the development of language models.

With the rise of Pre-trained Language Models (PLMs) based on the Transformer~\cite{Vaswani2017AttentionIA} architecture, the first major shift transformed from a one-stage (fully supervised learning) to a two-stage process (pre-train + fine-tune). We have embraced the era of PLM that avoids the computational cost of training from scratch. The first stage is pre-training an LM with a wealth of unlabeled data, and the second stage is fine-tuning the PLM with a small amount of labeled data for specific tasks.

Notable generative PLM GPT-3~\cite{brown2020language} with 175B parameters has presented impressive in-context learning ability to generate expected responses from few-shot demonstrations, which triggered the second major shift (pre-train, prompt and predict). It is feasible for prompting to guide a single generalist model to perform multiple tasks through appropriate prompts rather than updating parameters, saving substantial computational costs required to fine-tune separate models for each task. As a result, the parameter scale of language models continues to expand to a ten-billion level, opening the era of LLM that have set a solid stage for further development of prompting. In conclusion, prompting has dominated the interaction way between humans and AI systems, with great potential to open the door to AGI.

\subsection{Prompt Type} \label{Prompt_type}
Prompt as the model input to describe human intention can be various modalities such as text, images, audio, \etc, its quality directly determines the accuracy of specific tasks. This paper only discusses the prompt in text form in the field of NLP, referred to as ``hard prompt''. While the text is processed into vectors by neural networks, the prompt in vector form is also known as ``soft prompt''. Considering the close relationship between prompt type and model architecture, the following introduction will be presented from the perspectives of PLM and LLM eras, respectively.

\subsubsection{Hard Prompt} \label{Hard_Prompt}
\begin{table*}[t]
    \centering
    \caption{Three shapes of the hard prompt, where the \textcolor{GGreen!150}{green} represents hard prompts and the \hl{highlighted} represents CoT. (a) Cloze prompt is usually a prompt template with an empty space to be filled. (b) Prefix prompt uses concise instructions and input-output examples to activate the memory of LLMs during pre-training. (c) Detailed description contains task-specific instruction, context, input, and output indicators, which makes better use of LLM emergent abilities in various downstream tasks.}
    \label{Tab_hard_prompt}
	\begin{adjustbox}{width=\textwidth}
		\begin{tabular}{llcl}
			\toprule[1.5pt]
			Prompt Shape & Task & Illustration & Instantiation\\
			\midrule
			a. Cloze & NLU &
                \begin{minipage}[c][1.2cm][c]{5cm}
				\centering
				\includegraphics{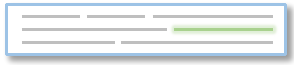}
			\end{minipage}
                & How was the movie? \textcolor{GGreen!150}{[MASK]}, I was surprised.
			\\
			\midrule 
			b. Prefix & NLG & 
                \begin{minipage}[c][1.2cm][c]{5cm}
				\centering
				\includegraphics{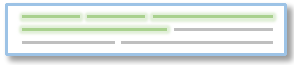}
			\end{minipage}
                & \color{GGreen!150} \makecell[l]{Translate English to French: \\ sea otter => loutre de mer \\ cheese =>}
			\\
			\midrule 
			\makecell[l]{c. Detailed\\description} & NLP &
			\begin{minipage}[c][4cm][c]{5cm}
				\centering
				\includegraphics{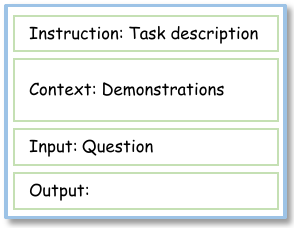}
			\end{minipage}
                & \makecell[l]{Assuming you are a mathematician, help me solve the\\following math problem.\\
                Question: John writes 20 pages a day. How long will it\\take him to write 3 books that are 400 pages each? \\
                Answer: \hl{He wants to write 3*400=<<3*400=1200>>1200 pages.}\\\hl{So it will take him 1200/20=<<1200/20=60>>60 days.}\#\#\#\# 60. 
                \\Question: Joy can read 8 pages of a book in 20 minutes.\\How many hours will it take her to read 120 pages? \\
                Answer: }
			\\
			\bottomrule[1.5pt]
		\end{tabular}
	\end{adjustbox}
\end{table*}
\begin{itemize}[leftmargin=*]
    \item \textbf{The PLM Era}.
    Transformer-based PLMs can be categorized into three types according to their architecture: Encoder-only, Decoder-only, and Encoder-Decoder. The input formats suitable for each model largely depend on the choice of pre-training tasks, so different prompt shapes are designed for specific architectures. For example, Encoder-only models~\cite{Devlin2019BERTPO, Liu2019RoBERTaAR, Lan2019ALBERTAL} pre-trained with Masked Language Modeling (MLM) task are good at predicting the [MASK] token in the prompt template, as shown in Table \ref{Tab_hard_prompt}-a. While Decoder-only models~\cite{Radford2018ImprovingLU, Radford2019LanguageMA, brown2020language} pre-trained with auto-regressive language modeling task do well in generating the next token after the prefix prompt from left to right, as depicted in Table \ref{Tab_hard_prompt}-b.
    
    \item \textbf{The LLM Era}.
    At present, most LLMs adopt a Decoder-only architecture that is fine-tuned with instructions and aligned with human preferences, so they can generate desired responses instructed by natural language prompts. Furthermore, as scaling up LLMs perform excellently across a variety of general tasks, prompts are designed to be more detailed and comprehensive. The basic components of a prompt can be summarized into four parts, as illustrated in Table \ref{Tab_hard_prompt}-c. 
\end{itemize}

\subsubsection{Soft Prompt} \label{Soft_Prompt}
\begin{table*}[t]
    \centering
    \caption{Three shapes of the soft prompt, where circles represent continuous vectors and squares represent discrete tokens, the \textcolor{OOrange!150}{orange} represents soft prompts, the \textcolor{GGray!150}{gray} represents frozen parameters and the \textcolor{GGreen!150}{green} represents hard prompts. (a) Embeddings refer to trainable parameters added to the prompt template. (b) Prefixes are prepended to each layer of the neural network to indicate the downstream task types. (c) Compact vectors are hard prompts compressed by open-source LLMs.} 
    \label{Tab_soft_prompt}
	\begin{adjustbox}{width=\textwidth}
		\begin{tabular}{l|ccc}
			\toprule[1.5pt]
			\makecell[l]{Prompt\\Shape} & a. Embedding & b. Prefix & c. Compact vector \\
                \midrule 
			    Task & NLU & NLG & NLP
                \\		
                \midrule
                Illustration &
       		\begin{minipage}[c][3cm][c]{4.5cm}
    				\centering
    				\includegraphics{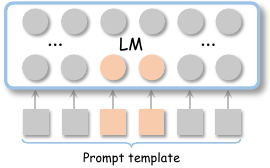}
    			\end{minipage} &
                \begin{minipage}[c][3cm][c]{4.5cm}
    				\centering
    				\includegraphics{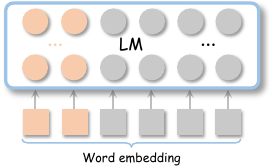}
    			\end{minipage} &
                \begin{minipage}[c][3cm][c]{4.5cm}
    				\centering
    				\includegraphics{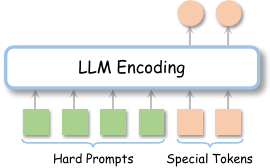}
    			\end{minipage}
                \\
			\bottomrule[1.5pt]
		\end{tabular}
	\end{adjustbox}
\end{table*}
\begin{itemize}[leftmargin=*]
    \item \textbf{The PLM Era}.
    Early work has investigated parameter-efficient fine-tuning using soft prompts, including prompt tuning~\cite{Lester2021ThePO}, prefix tuning~\cite{Li2021PrefixTuningOC}, \etc As can be seen in Table \ref{Tab_soft_prompt}-a \& b, these methods freeze original parameters and fine-tune only a small number of additional soft prompts to achieve comparable performance to full-parameter fine-tuning. This phenomenon becomes more evident as the model scale increases. However, with the advent of the LLM era, many LLMs are closed-source with inaccessible parameters, leading to a relative stagnation in follow-up research on soft prompts.
    
    \item \textbf{The LLM Era}.
    In this survey, we refer to the vector representations of hard prompts inside language models as soft prompts as well. The reason soft prompts are still critical is that they can be trained together with the language model to achieve effective model adjustments. Soft prompts in the LLM era as shown in Table \ref{Tab_soft_prompt}-c are typically more compact than hard prompts, storing valid information from hard prompts with less space to accelerate inference. 
\end{itemize}

\subsection{Mathematical Modeling} \label{Mathematical_modeling}
To facilitate the broad advantages of LLMs across various research and applications, we review efficient prompting methods from the point of reducing resource consumption in real-world applications. We divide the primary sources into two main categories: human resources and computational resources. Our analysis reveals that although there are many commonalities in the optimization objectives within each category, there is a lack of connections between them. To bridge this gap, we attempt to model the optimization process of each category of efficient prompting methods into mathematical formulas and shed light on how they serve as the theoretical foundation for this survey. In this way, not only can beginners clarify the essence of efficient prompting methods at a glance, but researchers also can be inspired to rationally integrate the two strategies into win-win solutions that simultaneously reduce resources in both areas.

\textbf{Automatic Prompt Engineering} (\S\ref{Automatic_prompt_engineering}): The optimization object is mainly oriented to the instruction $p_{ins}$ and demonstration $p_{dem}$ of the prompt. The optimization objective is to search for the optimal natural language prompt $p^*$ that maximizes the performance of the target model $\text{M}$ in the discrete prompt space $P_\text{Hard}$: 

\begin{equation}\label{eq_Prompt_Design}
p^* = \argmax _{p \in P_\text{Hard}} \mathcal{P} [f(p_{ins}, p_{dem}; \text{M})]
\end{equation}

\noindent where $f(\cdot; \text{M})$ denotes the output of the target model and $\mathcal{P}$ denotes the performance measured by an evaluation metric comparing the output with the ground truth label, \eg accuracy in NLU or math reasoning tasks.

\textbf{Prompt Compression} (\S\ref{Prompt_compression}): There are two prompt compression methods performed in different optimization spaces. Text-to-Vector (T2V) compression in continuous space compresses long text $p^h_o$ (original hard prompt) into short vectors $p^s_c$ (compressed soft prompt) through fine-tuning the LLM. Text-to-Text (T2T) compression in discrete space compresses long text $p^h_o$ into short text $p^h_c$ (compressed hard prompt) by filtering redundant information, or summarizing until information is sufficient. The informativeness $I(p)$ (such as self-information~\cite{Li2023CompressingCT}, perplexity~\cite{Jiang2023LLMLinguaCP, Jiang2023LongLLMLinguaAA}, \etc) can be measured by a Small Language Model (SLM) or an LLM. The optimization objective is to minimize the difference in model output distributions before and after the prompt compression:

\begin{equation}\label{eq_Soft_Prompt_Compression}
p^{s*}_{c} = \argmin_{p^s \sim \{\theta_\text{M},\; \theta_\text{Soft}\}} \mathcal{D} [q(y_\text{M} \;|\; p^h_{o}) \parallel q(y_\text{M} \;|\; p^s_{c})]
\end{equation}

\begin{equation}\label{eq_Hard_Prompt_Compression}
p^{h*}_{c} = \argmin_{I(p^h) \geq \lambda} \mathcal{D} [q(y_\text{M} \;|\; p^h_{o}) \parallel q(y_\text{M} \;|\; p^h_{c})]
\end{equation}

\noindent where $\theta$ denotes trainable parameters, $\lambda$ denotes minimum threshold of informativeness, $q(y_\text{M} \;|\; \cdot)$ denotes the output distribution of the target model $\text{M}$ and $\mathcal{D}$ denotes the distance metric between the output distributions, \eg Kullback-Leibler divergence.

\begin{figure}[t!]
\centering
\includegraphics[scale=0.75]{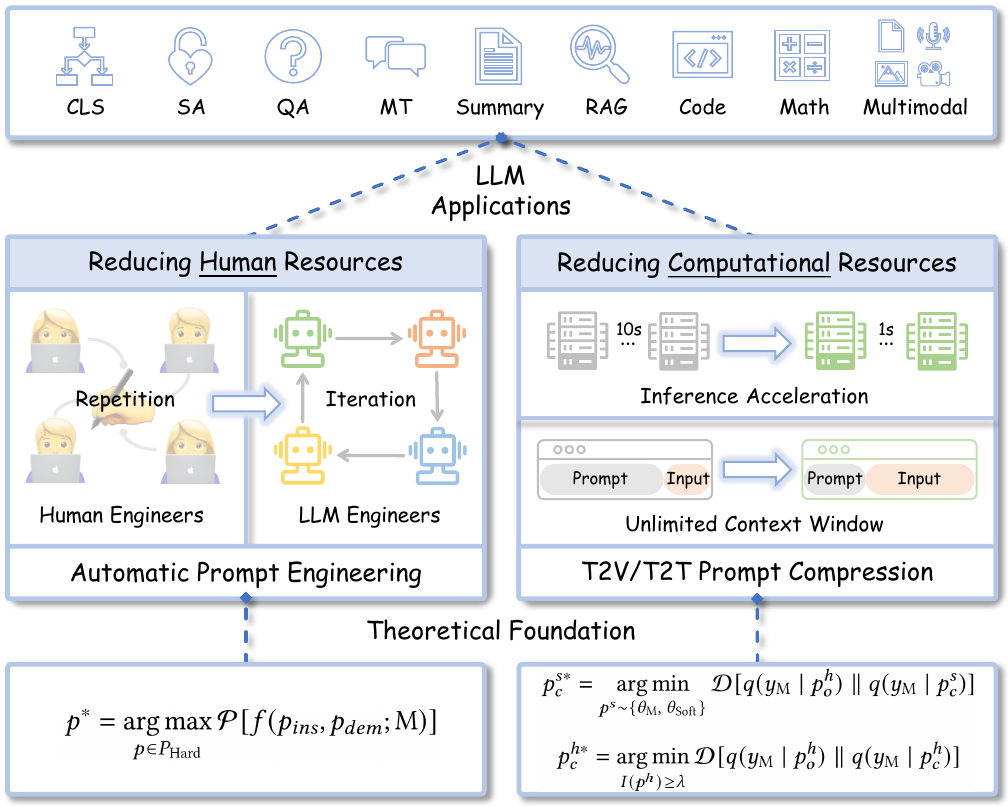}
\caption{An overview of efficient prompting methods.}
\label{Fig_Overview}
\end{figure}

Based on the above theoretical foundation, we provide a detailed presentation of the representative efficient prompting methods from two categories in the following, which will further enhance the understanding of these mathematical formulas. Figure \ref{Fig_Overview} shows an overview of this survey.
The left side expresses that automatic prompt engineering (\S\ref{Automatic_prompt_engineering}) takes advantage of LLMs' powerful generative capabilities through meta-prompts to iteratively exploit high-quality instructions (\S\ref{Instruction_design}) and optimize CoT prompting frameworks (\S\ref{CoT_optimization}) for better performance, which substitutes repetitive labor of human prompt engineers.
The right side discusses prompt compression (\S\ref{Prompt_compression}) in both continuous and discrete spaces to reduce prompt length for more input space and conserve computational resources for inference acceleration. T2V compression (\S\ref{Text-to-vector_compression}) utilizes vectors to cache key information of hard prompts by training open-source LLMs. T2T compression (\S\ref{Text-to-text_compression}) preserves only valid information from hard prompts without performance loss.
\section{Automatic Prompt Engineering} \label{Automatic_prompt_engineering}

\begin{figure}[t]
\centering
\includegraphics[scale=0.8]{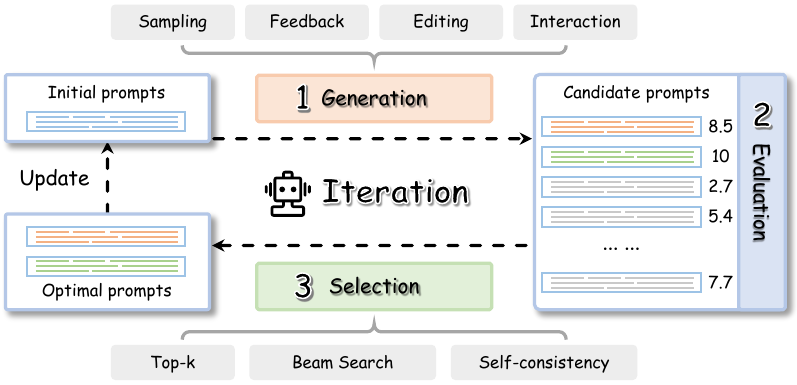}
\caption{The basic pipeline of automatic prompt engineering. Step 1: The discrete prompt space is expanded according to the customized optimization direction. Step 2: Candidate prompts are reasonably evaluated based on target model performance. Step 3: The optimal prompts are selected from the prompt pool using appropriate sampling strategies.}
\label{Fig_Prompt_optimization}
\end{figure}

Prompt engineering refers to designing effective prompts to make better use of specific language models in complex scenarios.

\textbf{Challenges}.
It is intuitive for prompt engineering to manually design elaborate prompts that make language models fully understand human intentions and mimic human behaviors. However, due to the limited comprehension competency of early language models, prompts were usually concise instructions and left little room for manual improvement. Later on, since the revolutionary breakthrough in the general capabilities of scaling language models, humans began to frequently use more detailed natural language prompts to interact with LLMs. A large number of experiments have found that the upper boundary of prompting LLMs becomes ambiguous, so human expertise cannot quickly and accurately navigate novel tasks. We summarize several challenges of manual-designed prompts for LLMs: (1) \textit{\textbf{Sensitivity}}: Especially in zero-shot settings, subtle differences in prompt content can lead to significant differences in model performance~\cite{jiang2020can, zhao2021calibrate, sclar2023quantifying}, so customized prompts are needed for different models and tasks. (2) \textit{\textbf{Suboptimality}}: Although well-crafted prompts effectively improve task accuracy, the internal compatibility of LLMs with particular prompts remains uncertain. The missing optimization direction means that human engineers can only rely on trial and error to explore a wider prompt space in search of relatively optimal solutions. (3) \textit{\textbf{Discrepancy}}: LLMs may interpret natural language differently from humans~\cite{Deng2022RLPromptOD}, sometimes gibberish prompts inconsistent in grammar and syntax may be more effective, which is obviously beyond the scope of human-designed prompts. To conclude, designing natural language prompts is an art that demands a lot of time and experience in the LLM era, which inevitably over-consumes human resources.

\textbf{Solutions}.
LLMs may exactly know what kinds of prompts they want, so a wide range of LLM-driven prompt optimization methods have been proposed to mitigate human biases and adaptively search out high-quality prompts to maximize LLM performance by Eq. \ref{eq_Prompt_Design}. This line of work is collectively referred to as automatic prompt engineering, which essentially mimics \textit{\textbf{search algorithms in discrete space}} consisting of three iterative steps, as shown in Fig. \ref{Fig_Prompt_optimization}.

According to the different components of prompts, automatic prompt engineering is divided into two parts: Instruction Design in \S\ref{Instruction_design} and CoT Optimization in \S\ref{CoT_optimization}. Both approaches primarily employ search algorithm principles to identify optimal solutions, with meta-prompts playing a crucial role in guiding LLMs (refer to as optimizers) to perform generation, evaluation, and selection instead of human engineers, effectively automating prompt design in a simple and efficient way. The difference between them is that the former focuses on expanding the instruction space, while the latter is more inclined to construct comprehensive problem-solving frameworks.

\subsection{Instruction Design}\label{Instruction_design}
Instructions as the most direct way of human-computer interaction are the first choice for in-context learning. Typically, instructions are concise and simple task descriptions that guide LLMs in performing specific tasks, similar to activating certain knowledge in a specific domain. As the core component of prompts, instruction design has been the most prevalent subject of numerous discussions in the field of automatic prompt engineering. The premise for imitating search algorithms is to create a sufficient search space, which is perhaps a big challenge for human engineers. However, it is a piece of cake for LLMs with  extensive language knowledge accumulated from pre-training and excellent generative capability mastered during supervised fine-tuning, leading to creativity superior to that of humans. In the following, we will organize existing works from three perspectives.

\subsubsection{Sampling-based methods}\label{Ins_Sampling}
Self-instruct~\cite{wang2022self} has indicated that LLMs can generate applicable instructions from a small set of seed human-written instructions automatically. To fully leverage the generative capability of LLMs in constructing a diverse prompt space, the most straightforward approach is iteratively sampling during inference. But how does the LLM generate instructions? It can either match the most suitable instructions based on input-output pairs in the prompt or imitate a well-crafted prompt. Subsequently, the meta-prompt steers the LLM to perform each step in automatic prompt engineering.

APE (Automatic Prompt Engineer)~\cite{Zhou2022LargeLM} is a pioneering work that treats the human-intractable question as a black-box optimization process guided by LLMs. ALL prompting optimization processes can be executed by ONE human-aligned instruction-generation LLM. The LLM respectively acts as different prompt engineers to optimize the instruction in the prompt. Step 1: The inference LLM generates instructions according to demonstrations (input-output pairs). Step 2: The scoring LLM evaluates these candidate instructions by predicting the probability of the next token and selects high-quality instructions whose scores exceed a certain threshold. Step 3 (Optional): The resampling LLM performs Monte Carlo search for semantically similar instructions around the best candidates. Such an optimization pipeline is equivalent to compressing the demonstrations into a single instruction, which greatly outperforms the prior LLM baseline by a large margin in zero/few-shot learning and zero-shot CoT settings. Moreover, APE achieves better or comparable performance to the instructions generated by human annotators on 24/24 Instruction Induction tasks and 17/21 curated Big-Bench~\cite{Suzgun2022ChallengingBT} tasks.
However, prompt optimization methods face challenges of significant performance gaps under data distribution shifts. GPO framework~\cite{Li2023RobustPO} first utilizes a meta-prompt to ask the LLM to generate multiple corresponding outputs for a single unlabeled input. Then, the prompt ensemble labeling strategy selects the output with the highest consistency as the label. Finally, demonstrations among different distributions are mixed to run APE for joint prompt optimization and obtain the final optimized prompt with high robustness and generalizability.

Based on the exploration and exploitation concept, OPRO~\cite{Yang2023LargeLM} uses meta-prompt to instruct the LLM respectively as an optimizer to exploit plenty of instructions and a scorer to evaluate their scores. Then, instructions paired with their scores iteratively form an optimization trajectory to guide the optimizer to explore new instructions for higher task accuracy.
Due to the lack of human trial-and-error processes and deeper domain expertise in the aforementioned methods, PromptAgent~\cite{Wang2023PromptAgentSP} automatically design expert prompts rich in domain-specific details and structured guidance through Monte Carlo Tree Search (MCTS) based on the self-reflection ability of LLMs. It systematically augments the prompt space in a tree structure and opts for the best node in the optimal path with the highest reward.
Inspired by Mixture of Experts (MoE), Mixture-of-Prompt (MoP)~\cite{wang2024one} expands the instruction optimization space by clustering demonstrations for different experts and then identifies the best instruction for each cluster by a Region-Based Joint Search (RBJS) algorithm.


GPS~\cite{xu2022gps} has verified the feasibility of automatically searching for high-performing prompts based on the Evolutionary Algorithm (EA). EA is a highly robust and widely applicable cluster of optimization algorithms that simulate the principle ``survival of the fittest'' of natural selection, including Genetic Algorithm (GA), Differential Evolution (DE), \etc In the field of prompt optimization, some efforts adhere to the EA concept to enhance the diversity of the discrete prompt space. The LLM serves as the evolutionary operator to perform four critical operations guided by meta-prompts. Step 1: Initializing the prompt population with appropriate seeds. Step 2: Evolving initial prompts into candidate prompts by selection, mutation, and crossover. Step 3: Evaluating candidates based on a proper metric. Step 4: Updating the population with the fittest samples. Population initialization may vary from different methods while the iterative evolution process is basically the same.

EvoPrompting~\cite{Chen2023EvoPromptingLM} introduces LLMs to be evolutionary operators for the first time, which avoids resource wastage and human bias of manual-designed discrete search space. The code-pretrained LLM is used to perform the end-to-end meta-learning algorithm for Neural Architecture Search (NAS) task. Parent samples are initialized by a handful of well-designed program seeds to warm-start the iteration. EvoPrompting outperforms both manual-design and naive few-shot prompting methods in terms of model performance and computational efficiency.
EvoPrompt~\cite{Guo2023ConnectingLL} is the first attempt to synergize LLM and EA algorithms through natural language descriptions, which achieves an exploration and exploitation trade-off in discrete prompt optimization. The initial population includes both manual prompts to leverage human prior knowledge and diverse generated prompts selected by the roulette wheel method to avoid local optimum. Experiments suggest choosing GA when several high-quality prompts already exist, otherwise DE. 
Promptbreeder~\cite{Fernando2023PromptbreederSS} is a general-purpose self-referential self-improvement mechanism to evolve task-prompts (origin-prompts) and mutation-prompts (meta-prompts) simultaneously for a given domain.
AELP~\cite{hsieh2023automatic} optimizes long prompts by a simple greedy algorithm with beam search and utilizes search history to enhance the effectiveness of LLM-based mutation.
\citet{cui2024phaseevo} introduce a multi-phase exploration-exploitation strategy PhaseEvo to achieve joint optimization of instructions and demonstrations. Firstly, both available input-output pairs and human-designed prompts are used to explore a vast joint initialization population. Then, an LLM Improver generates candidates based on the feedback offered by an LLM Examiner. Notably, global and local LLM-based mutations are applied for evolution diversity and rapid convergence in succession.

\subsubsection{Feedback-based methods}\label{Ins_Feedback}
\begin{figure}[t]
\centering
\includegraphics[scale=0.85]{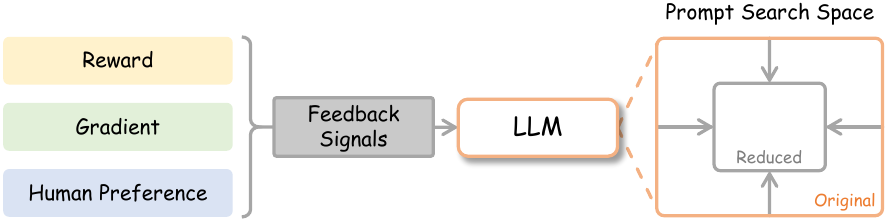}
\caption{Various feedback signals contribute to specifying the optimization space of automatic prompt engineering, where the \textcolor{OOrange!150}{orange} dashed line indicates the original prompt search space and the \textcolor{GGray!150}{gray} solid line represents the reduced search space that benefits from a more clear optimization direction.}
\label{Fig_Feedback}
\end{figure}

Accurate feedback is a key factor in supervising iterative updates of the optimization object to better versions, which can take various forms as shown in Fig. \ref{Fig_Feedback}, such as reward signals in reinforcement learning (RL), gradients in gradient descent, or even human preference. The difference between this line of work and the previous one is that the feedback signal indicates a more clear optimization direction so that the search space is relatively small.

Discrete prompt optimization for black-box LLMs can be formulated as the RL problem at an early stage. Typically, the policy combines a frozen PLM with a trainable neural network with a relatively small amount of parameters, and the reward reflects comprehensive feedback comparing the prediction of the target LLM with the ground truth.

RLPrompt~\cite{Deng2022RLPromptOD} consists of a frozen PLM (RoBERTa-large~\cite{liu2019roberta}) and a learnable task-specific MLP (Multilayer Perceptron) to generate tokens of prefix prompt one by one to maximize the reward, leading to a semantically incoherent prompt.
DSP~\cite{li2024guiding} fine-tunes a small policy LM (T5~\cite{raffel2020exploring}) using only minimal labeled data to generate nuanced, instance-specific directional stimulus prompts for original inputs, such as keywords added after summarization instructions.
PACE~\cite{dong2023pace} leverages LLMs as the dual roles of actors and critics for iterative prompt refinement.
PRewrite~\cite{kong2024prewrite} trains a prompt rewriter LLM (PaLM~\cite{anil2023palm}) to optimize input-independent instructions by RL and designs two rewrite strategies PRewrite-I with greedy decoding and PRewrite-S with search algorithm.
SCULPT~\cite{kumar2024sculpt} is a novel framework that systematically refines long prompts in the form of a hierarchical tree through an actor-critic mechanism of RL. There are two complementary feedback mechanisms: Preliminary Assessment evaluates the prompt structure before execution and Error Assessment diagnoses model errors after execution.

In fact, gradient descent is the most common algorithm for exploring optimal solutions in continuous space, where the gradient is a form of negative feedback. In the context of the widespread prevalence of close-resource LLMs, meta-prompts are commonly used to guide LLMs to implement critical steps of automatic prompting optimization by themselves, especially gradient generation can be seen as a kind of self-feedback.

ProTeGi~\cite{Pryzant2023AutomaticPO} proposes a simple and nonparametric prompt optimization solution with the help of API-based LLMs. The non-parametric nature is reflected in the following aspects: (1) Textual gradient: LLM summarizes the flaws of the initial prompt as the textual gradient based on mini-batches of training data. (2) Backpropagation: Another LLM fixes flaws by editing the initial prompt into candidate prompts in the opposite semantic direction of the textual gradient. (3) Expansion: The paraphrasing LLM performs Monte Carlo search to explore the prompt space within a scope semantically similar to the candidates. (4) Beam search: The best arm identification algorithm in bandit optimization chooses the most promising candidates for the next iteration.
Interestingly, as the meta-prompt plays an important role in manipulating LLM engineers, PE2~\cite{ye2023prompt} focuses on prompt engineering the meta-prompt following concepts commonly used in gradient-based optimization.
To optimize search space for ProTeGi, PREFER~\cite{zhang2024prefer} designs a feedback-reflect-refine framework for prompt ensemble learning, where the feedback reflects on the inadequacies of each weak prompt and works in tandem to cover multiple sub-spaces during boosting. There is also an effective bagging method to enhance the stability of the prompt effect evaluation, which is superior to regular beam search or Monte Carlo search.
Based on gradient accumulation, AutoHint~\cite{sun2023autohint} deduces a hint for each sample (prompt-input pair) with incorrect output and summarizes the gradient from per-sample hints. UniPrompt~\cite{juneja2024task} generates optimized prompts that capture multiple facets of a task by clustering samples with similar task facets and combining feedback from each cluster like mini-batch.
Without summarizing gradient information, GPO~\cite{tang2024unleashing} directly guides the LLM to realize the update direction from the meta-prompt including task prompts, wrong demonstrations and prompt-score trajectory, and perform the update method with edit distance.
To effectively adapt to diverse data distributions, AMPO~\cite{yang2024ampo} automatically constructs a multi-branched prompt to efficiently handle multiple patterns in complex tasks. LLM-Analyzer identifies the cause for each failed case and LLM-Summarizer summarizes all the causes into different patterns iteratively to reduce repetitive and redundant branches. LLM-Revisor finally optimizes the prompt either in depth (by adding more details) or in breadth (by adding more branches).

Human preference alignment benefits LLMs in learning human intelligence, where the effect of alignment depends largely on the quality of the preference data.

Similar to RLHF~\cite{ouyang2022training}, APOHF~\cite{lin2024prompt} automates prompt optimization by training a neural network to predict the best prompt based on human preference.
BPO~\cite{Cheng2023BlackBoxPO} constructs paired original and optimized prompts based on human preference feedback to train a 7B seq2seq model as the prompt preference optimizer. The optimized prompts are accommodated LLMs’ understanding and applicable to various LLMs, which further eliminate the Human-LLM alignment gap compared with RL based on PPO~\cite{Schulman2017ProximalPO} and DPO~\cite{Rafailov2023DirectPO}. 
APEER~\cite{jin2024apeer} employs the LLM to iteratively refine prompts, especially for passage relevance ranking tasks based on self-feedback and high-quality prompt preferences.
FIPO~\cite{lu2024fipo} introduces the first large-scale Prompt Optimization Preference dataset (POP) that collects rejected data from a suboptimal GPT-3.5 and chosen data from an optimal GPT-4 and undergoes rigorous cross-validation by human experts and analytical models. POP is used to fine-tune offline local LLM-based optimizers to improve free-form instruction-oriented prompts.

However, whether LLM is a good prompt optimizer is a question worth exploring~\cite{ma2024large}.

\subsubsection{Editing-based methods}\label{Editing}
In addition to the aforementioned methods, it is also feasible to expand the prompt space without deviating from the original prompt by editing the prompt directly with delete, swap, paraphrase, add, \etc operations at specific lexical unit levels.

GrIPS~\cite{Prasad2022GrIPSGE} is a gradient-free, edit-based search method consisting of three steps. Step 1 involves two operations: slicing and editing. Firstly, base instructions are split into shorter lexical units (\ie word, phrase, or sentence) using CRF-based constituency parser~\cite{Zhang2020FastIB}, with phrase-level slices proven to be the most helpful. Then, the certain slice is edited by one of four operations (delete, swap, paraphrase, addition) to expand the instruction space. Step 2 evaluates candidate instructions on a score set and selects the best instructions based on greedy search, beam search, and simulated annealing (SA) algorithms. Step 3 conducts multiple search iterations until the score no longer improves or a maximum threshold is reached. GrIPS has been shown to work for many prompt modes (with/without demonstrations) especially with instruction fine-tuned LLMs.
Plum~\cite{Pan2023PlumPL} as a general prompt learning method that satisfies automatic, discrete, black-box, gradient-free, and interpretable all at once combines GrIPS editing operations with various metaheuristic algorithms (including hill climbing, simulated annealing, genetic algorithms with/without crossover, tabu search, and harmony search) to expand the discrete prompt space.
SPRIG~\cite{zhang2024sprig} especially optimizes system prompts with GrIPS editing operations.

Editing operations can also be used to explore optimization space in the RL process. TEMPERA~\cite{Zhang2022TEMPERATP} sequentially edits query-dependent prompts during test time. Editing actions for prompts include Swap, Add and Delete for instructions; Permute and Swap for examples; Change for verbalizers. According to the step reward proposed in RLPrompt~\cite{Deng2022RLPromptOD}, TEMPERA uses the score difference between successive edits as the immediate reward and attention-based policy architecture to choose possible actions. Distinct from prior work, this formulation strikes a good balance between human prior knowledge and prompt performance.

\subsection{CoT Optimization}\label{CoT_optimization}
Given the brevity of instructions, the information they contain is predictably insufficient. To better inspire the potential of LLMs in solving complex tasks, task-related demonstrations can be added to enrich the context of the prompt. Among these techniques, Chain-of-Thought (CoT)~\cite{Wei2022ChainOT} was introduced as a pivotal prompting strategy that significantly enhances LLMs' reasoning capability. The core idea of CoT is ``divide and conquer'' which breaks down complex problems into finer-grained subproblems and explicitly deduces step by step in a complete framework of mind. As reasoning is one of the most meaningful tasks for AI systems, numerous optimization efforts for prompting frameworks represented by CoT have emerged. We incorporate them into CoT optimization in automatic prompt engineering and provide three categories similar to the instruction design classification criteria in \S\ref{Instruction_design}. The final category involves the use of external tools to interact with broader knowledge bases.

\subsubsection{Sampling-based methods}\label{CoT_Sampling}
Zero-Shot-CoT~\cite{Kojima2022LargeLM} has first validated LLMs can sample diverse CoT demonstrations (rationale-answer pairs) with a piece of instructive prompt ``Let's think step by step'', which greatly reduces the human resource requirements of manual-designed few-shot CoTs.
Following this, a new decoding strategy, self-consistency~\cite{wang2022self2}, was proposed to replace the naive greedy decoding used in CoT sampling. Specifically, it first samples diverse rationale-answer pairs by adjusting the temperature coefficient and then selects the most consistent answer by majority voting.
These two studies significantly unleash the reasoning ability of LLMs and establish a reliable foundation for CoT prompting optimization.

At first, LMSI~\cite{huang2022large} has demonstrated that LLMs can self-improve by training with mixed formats of rationale-augmented answers using CoT prompting and self-consistency.
Instead of augmenting training data, a lot of work has begun to focus on improving prompting input for LLMs based on their own output. To mitigate LLM hallucinations in Zero-Shot-CoT prompting, Auto-CoT~\cite{Zhang2022AutomaticCO} utilizes Sentence-BERT~\cite{Reimers2019SentenceBERTSE} to cluster questions with diversity and generate corresponding CoT demonstrations for more flexible and task-adaptive few-shot prompting.
Boosted Prompting~\cite{pitis2023boosted} encourages the LLM to sample multiple rationales for different problems. The problems with minimal accuracy or consistency and their correct rationales are considered as hard examples to form informative prompts that iteratively comprise a boosted ensemble, increasing overall coverage of the problem space.
COSP~\cite{Wan2023BetterZR} specially designs a scoring function that incorporates consistency, diversity and repetitiveness to select outstanding outputs generated by Zero-Shot-CoT prompting LLMs in stage 1, and self-adaptively determines the number of demonstrations through outcome entropy in stage 2. Finally, majority voting over outputs from both stages forms the final prediction.
USP~\cite{wan2023universal} is an improved, universal version of COSP~\cite{Wan2023BetterZR} not limited to reasoning tasks. It employs the same candidate prompt generation method but differs in using a Task-Specific Selector, which designs scoring functions tailored to three classic NLP task types (classification, short-form generation, and long-form generation) to select the appropriate pseudo-demos. Finally, greedy decoding is used to replace majority voting.
Facing mixed-task scenarios, Meta-CoT~\cite{Zou2023MetaCoTGC} employs a routing mechanism to match the question type with an off-the-shelf Demo Pool. If unmatched, Zero-Shot-CoT prompting is used to generate CoT demonstrations that automatically update the data cache with similar-question clusters based on density like Auto-CoT~\cite{Zhang2022AutomaticCO}.
Reprompting~\cite{Xu2023RepromptingAC} generates step-by-step CoT demonstrations by Gibbs Sampling. It deduces the joint distribution of CoT demonstrations by sampling from the conditional distribution of the training data, which can also be viewed as a variant of evolutionary algorithms that iteratively finds effective CoT for each model given a few question-answer pairs without human intervention.

\subsubsection{Feedback-based methods}\label{CoT_Feedback}
Self-refine~\cite{madaan2024self} has verified that LLMs possess the competency for iterative refinement across various tasks based on self-feedback, where meta-prompts instruct the same LLM for initial generation, feedback, and refinement. As a result, many optimized prompting frameworks include iterative feedback, which is usually the LLM's deep reflection on its own behavior that may be completely wrong or suboptimal. We summarize related work involving the nature of step-by-step reasoning in this category.

Reinforcement learning still works in reasoning tasks, PromptPG~\cite{lu2022dynamic} trains a small linear layer on top of the frozen PLM (BERT~\cite{Reimers2019SentenceBERTSE}) to select performing in-context examples based on the consistency between LLM predictions and labels, ensuring the integrity of prompts. 
Prompt-OIRL~\cite{sun2023query} optimizes query-dependent prompts on an instance level rather than a distributional level by offline inverse RL based on a proxy reward model (XGBoost~\cite{chen2015xgboost}). The proxy fed with query-prompt pairs is trained to approximate the true online reward calculated by black-box LLMs and golden labels.
Reflexion framework~\cite{Shinn2023ReflexionLA} designs a special memory module to transform scalar rewards into natural language feedback. The interaction history (call it trajectory) between the actor and the environment serves as the short-term memory while a verbal experience feedback summarized from the trajectory with its reward scores is stored in long-term memory. The Reflexion agent outperforms other LLMs based on the synergy of the two memory components.
In contrast to Reflexion, which executes a task multiple times for online dynamic feedback, TaskLLM in the PROMST framework~\cite{Chen2024PRomptOI} executes a task at once to automatically synthesize offline feedback about errors based on human-designed feedback rules. SumLLM summarizes feedback as context for GenLLM to generate new prompt candidates. To mitigate the evaluation cost, the score prediction model is fine-tuned to heuristically sample a subset of candidates for evaluation. 

Negative feedback may be more helpful for LLMs to deliberate. DTG~\cite{Li2023DeliberateTG} designs templates to prompt LLMs to detect error types in irrelevant system output and triggers the deliberation ability of LLMs from the negative feedback. The prompt templates require only minimal adjustments to guide the LLM in performing a simple single-step inference applicable to a wide range of text generation tasks. 
Reprompt~\cite{Chen2024RePromptPB} optimizes the step-by-step instructions of CoT across various reasoning tasks. If there is no such part, the additional checker will convert the CoT into one with a step-by-step instruction. Inspired by gradient descent, the loss is the focus point summarized from a batch of interaction history between the target LLM and the feedback generator. Then, the prompt optimizer mainly updates the common prompt part based on the loss. This act loop iterates until the prompt has converged.

\subsubsection{Interaction-based methods}\label{Interaction}
LLMs have been shown to often experience issues like hallucinations and error propagation when addressing reasoning tasks. To this end, it is increasingly common for LLMs to interact with external resources to refine their internal capabilities. For example, intervening with external tools to check for errors at intermediate steps in a prompting framework has proven to be highly effective.

Improved ReAct~\cite{Yao2022ReActSR} prompt trajectories are formed by sparsely synergizing CoT (reason-only) prompts and action-only prompts, which guide the LLM to generate reasoning traces and task-specific actions in an interleaved manner and overcome prevalent reasoning issues by interacting with a simple Wikipedia API.
It is more natural and conversational for Verify-and-Edit framework~\cite{zhao2023verify} to produce verifying questions for less consistent CoT demonstrations and post-edit rationales with external knowledge for more factually aligned predictions.
ART framework~\cite{Paranjape2023ARTAM} retrieves similar tasks from the task library to craft a few-shot prompt that decomposes the task into sub-steps and calls tools provided in the tool library such as search engine and code generation when necessary.
In self-ask framework~\cite{press2022measuring}, the LLM continues to explicitly ask follow-up questions and answer with a search engine until there is sufficient information to solve the original task. 
ToolLLM~\cite{qin2023toolllm} is a general tool-use framework that includes ToolBench constructed with API collection, instruction generation and solution annotation, ToolLLaMA supervised fine-tuned by ToolBench, and ToolEval for multi-round interaction evaluation.
ATC~\cite{shi2024chain} shows that LLMs can not only utilize a chain of tools through programming until solving the complex tasks (multi-tool user) but also master fast-paced new tools automatically by black-box probing (multi-tool learner).
\section{Prompt Compression} \label{Prompt_compression}

\begin{figure}[t]
\centering
\includegraphics[scale=0.85]{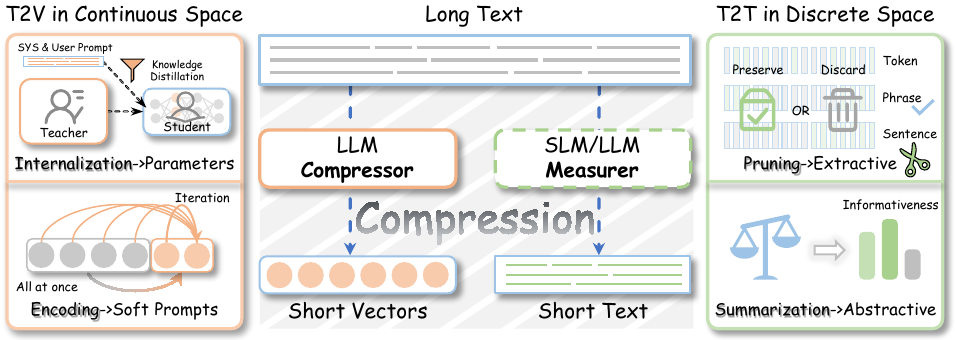}
\caption{Prompt compression in continuous and discrete space. T2V compression includes \textbf{Internalizing} system or user prompts into model parameters based on KD, and \textbf{Encoding} key information of hard prompts into soft prompts in an iterative or one-off way. T2T compression contains extractive and abstractive methods respectively are \textbf{Pruning} in various granularities, and \textbf{Summarization} for sufficient informativeness.}
\label{Fig_Prompt_compression}
\vspace{-0.6em}
\end{figure}

Prompt compression refers to distilling long text prompts into the shortest possible text or vectors without sacrificing LLM performance.

\textbf{Challenges}.
With the increasing generative capability of LLMs, researchers have found that prompting LLMs with more specific contexts in few-shot settings can improve task accuracy. Meanwhile, the applications of LLMs in long-context scenarios, such as multi-turn dialogue, Retrieval-Augmented Generation (RAG), multi-document summarization, \etc, are being actively examined. Theoretically, the more comprehensive task-relevant knowledge covered in the context, the better LLMs perform. However, the unlimited input length will bring difficulties for the practical deployment: (1) \textit{\textbf{Limited Context Window}}: Each LLM has a fixed input length identified during pre-training, any text exceeding this length will be truncated so that losing this part of contextual semantics. (2) \textit{\textbf{Catastrophic Forgetting}}: When there is no sufficient cache space, the LLM may forget previously learned knowledge when modeling long sequences. (3) \textit{\textbf{Slow Inference Speed}}: The large scale of LLMs is a double-edged sword, while LLMs with extensive parameters can store rich knowledge to be generalist, the over-consumption of computational resources is inevitable during inference.

\textbf{Solutions}.
To alleviate limitations of the context window, reusable system prompts can be encapsulated within the model itself, thereby conserving more adequate space for user inputs. Following earlier parameter-efficient prompt tuning efforts \cite{Lester2021ThePO, Li2021PrefixTuningOC}, a small number of learnable parameters (soft prompts) can carry task-specific information so that task descriptions in lengthy hard prompts can be transformed into more compact soft prompts to address catastrophic forgetting. Both approaches leverage open-source LLMs for training by Eq. \ref{eq_Soft_Prompt_Compression}, in the case of closed-source LLMs, the information density of hard prompts can be increased with the help of the SLM for information measurement or the LLM for summarization as described in Eq. \ref{eq_Hard_Prompt_Compression} to speed up inference.

According to the type of compressed prompts, we divide prompt compression into Text-to-Vector (T2V) compression in continuous space and Text-to-Text (T2T) compression in discrete space. Figure \ref{Fig_Prompt_compression} depicts the four main types of methods, T2V compression considers the LLM as a compressor that compresses essential text information into more compact vector representations. T2T compression directly measures the token informativeness and compresses as much as possible redundant information with acceptable performance loss.

\subsection{Text-to-Vector Compression} \label{Text-to-vector_compression}
When a language model processes text prompts, converting them into vectors is a necessary step. Since vectors can be denser representations than text, it is natural to consider compressing discrete text prompts outside the model into continuous vectors inside the model. In this context, continuous vectors can be either internal parameters of the model referred to as internalization, or additional soft prompts known as encoding. This kind of prompt compression not only expands the context window but also accelerates inference speed especially when the same prompts are frequently reused.

\subsubsection{Internalization} \label{Internalization}
Here, we introduce the concept of Knowledge Distillation (KD)~\cite{Hinton2015DistillingTK}, where the knowledge from a larger teacher model is transferred into a smaller student model. Inspired by this, the language model can be fine-tuned with Kullback-Leibler (KL) divergence of output distribution between original and compressed prompts to distill the knowledge from text prompts into model parameters, then achieve the internalization of prompt functionality.

The initial research efforts attempted to internalize the system prompts. To develop a versatile AI system aligned with human values, Context Distillation~\cite{Askell2021AGL} internalizes an HHH (helpful, honest, and harmless) prompt within the language model by KD between the original model with the context and input and the distilled model with the input alone.
PI (Prompt Injection)~\cite{Choi2022PromptIP} proposes the Pseudo-INput Generation (PING) method to internalize system prompts, which avoids the repeated computation of reused prompts during inference. The first stage generates pseudo-inputs based on prompts using an input generator trained on task-specific data. The second stage performs KD between the teacher model with the prompt and pseudo-input and the student model only with the pseudo-input.

Later on, researchers start compressing the context of the user prompts inside the model. \citet{Snell2022LearningBD} propose to distill three specific types of context in the prompt: abstract task instructions, complex reasoning, and concrete training examples. Based on teacher and student prompt templates, the original input is transformed into two kinds of prompts with significant distribution differences. The language model is trained with different prompts to produce the same response for internalization. Three distillation methods are tested to demonstrate that once the model learns the knowledge of task-specific prompts, it can perform the corresponding task without explicit prompts.
Instruction Distillation~\cite{Sun2023InstructionDM} internalizes three complex instruction techniques: pointwise ranking, pairwise ranking, and listwise ranking. The goal is to address the inefficiency of zero-shot relevance ranking by enabling LLMs to rank efficiently with simpler instructions. 
Especially for reasoning, Distilling Step-by-Step~\cite{Hsieh2023DistillingSO} not only enables student SLM to learn labels from the teacher LLM but also extracts rationales as additional supervisory signal for internalizing LLM's reasoning capability inside the SLM under a multi-task framework, which achieves better performance than standard fine-tuning with smaller model sizes and less training data.
As for retrieval-augmented generation task, inspired by the modality encoder CLIP~\cite{Linearprobe2021LearningTV} extracting modality features, xRAG~\cite{Cheng2024xRAGEC} redefines document embeddings in dense retrieval as features from the retrieval modality, which are seamlessly integrated into the language model representation space using a plug-and-play projector (modality fusion bridge). There are two training phases for xRAG: first, the LLM is internally compatible with compressed representations provided by the projector with paraphrase instructions, and then the frozen LLM with trainable projector is instruction fine-tuned to perform specific downstream tasks based on compressed representations. Self-distillation is also used to enhance xRAG's resilience in noisy contexts.

\subsubsection{Encoding} \label{Encoding}
In the above internalization methods, the compressed hard prompts lack explicit carriers, while encoding methods predefine LLM-adapted soft prompts to represent hard prompts such as multi-task instructions, lengthy context, \etc Normally, there is a compressor encoding hard prompts into soft prompts that can be applied for various downstream tasks efficiently.

\citet{Wingate2022PromptCA} first proposed to compress hard prompts into soft prompts, where hard prompts are prepended to the input sequence while soft prompts are prepended to the input embeddings. By minimizing both output distributions, important information from complex hard prompts is distilled into concise soft prompts. Although this training process is costly, it reduces inference costs when the same prompts are reused.

\begin{figure}[t]
\centering
\includegraphics[scale=0.8]{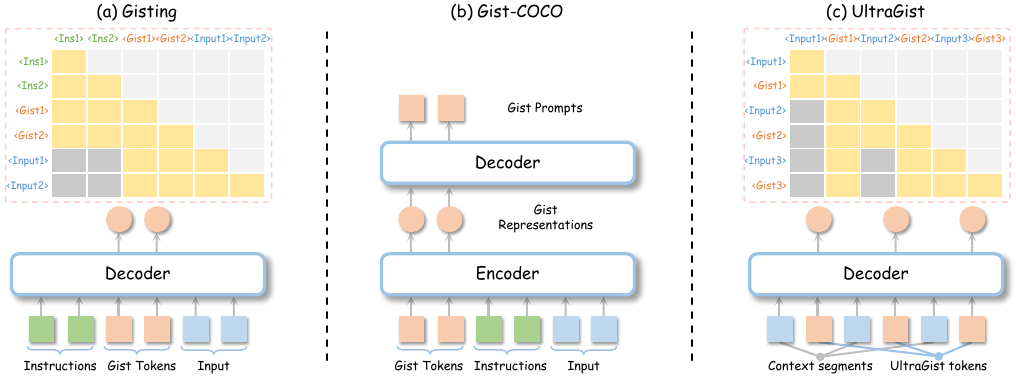}
\caption{Gisting series work compresses key information of hard prompts into gist tokens based on an encoder or a decoder trained with special attention mechanisms. The matrices in the upper half represent masking strategies, where the \textcolor{GGray!150}{gray} box indicates the standard mask and the \textcolor{YYellow!150}{yellow} box indicates the gist mask.}
\label{Fig_Gist}
\end{figure}

There are a series of works derived from Gisting~\cite{Mu2023LearningTC} as shown in Fig. \ref{Fig_Gist} with the goal of better generalization among diverse prompts. 

The goal of Gisting~\cite{Mu2023LearningTC} is encoding multi-task instructions into gist tokens that can be cached and reused for computation efficiency. Notably, it leverages meta-learning in HyperTuning~\cite{Phang2022HyperTuningTA} to predict gist tokens instead of gradient descent to update soft prompts. Specifically, gist tokens are first added to the vocabulary and embedding matrix, then concatenated after the instructions, and finally fine-tuned based on masking as depicted in Fig.\ref{Fig_Gist}-(a). Gisting can be applied to both decoder-only and encoder-decoder architecture, enabling up to 26x compression of prompts, a 40\% reduction in FLOPs, 4.2\% wall time speedups, and storage savings, all with minimal loss in output quality.
Inspired by the Minimum Description Length (MDL) principle~\cite{Grnwald2007TheMD} from information theory, Gist-COCO (Gist COnditioned COding)~\cite{Li2024SayMW} compresses original prompts into shorter gist prompts with an encoder-decoder architecture. The encoder is fine-tuned to compress gist tokens into gist representations to simulate the output distribution of the original prompts via KL divergence. The decoder verbalizes these gist representations into gist prompts to generalize across different LLMs with high compression rates.
In addition to compressing short instructions, Gisting can also compress long contexts. As the name suggests, UltraGist~\cite{Zhang2024CompressingLC} compresses ultra-long context with gist tokens. It employs a decoder-only architecture with an optimized cross-attention mechanism as shown in Fig.\ref{Fig_Gist}-(c) to progressively compress fine-grained context segments into UltraGist tokens that follows the next segment for subsequent compression. It distinctly allows randomly sampled compression ratios to deal with dynamic context while maintaining near-lossless performance.

\begin{figure}[t]
\centering
\includegraphics[scale=0.7]{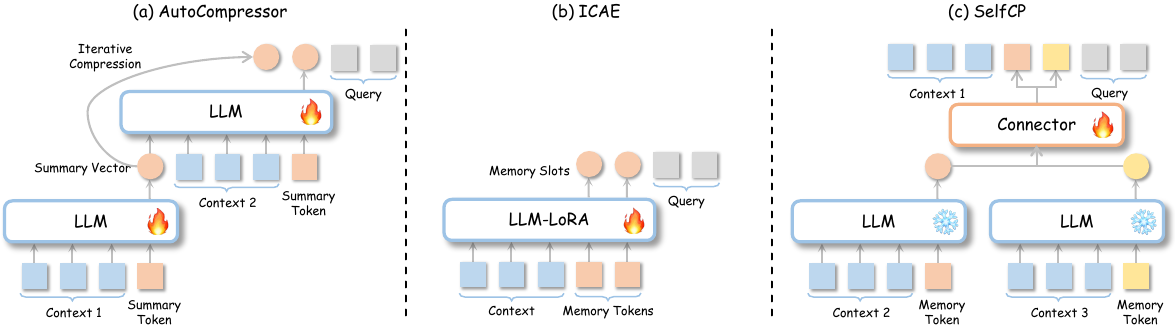}
\caption{Differences of representative encoding methods specially for long context. AutoCompressor iteratively compresses context segments with summary tokens. ICAE compresses the complete context all at once with memory tokens. SelfCP only compresses unlimited context segments based on a connector.}
\label{Fig_Context_compression}
\end{figure}

In the following, we introduce some research as shown in Fig. \ref{Fig_Context_compression} that specializes in compressing long context in the prompt, which realizes advantages in both inference acceleration and GPU memory reduction.

AutoCompressor~\cite{Chevalier2023AdaptingLM} employs a Recurrent Memory Transformer (RMT)~\cite{Bulatov2022RecurrentMT} architecture to iteratively compress arbitrary-length context segments with summary tokens into summary vectors. In subsequent iterations, summary vectors from the previous segment are concatenated in front of the next segment and summary tokens are concatenated at the end. Additionally, the backpropagation through time (BPTT) method~\cite{Chen2016TrainingDN} stops gradient updates on cached summary vectors after two steps of compression, further reducing computational load and GPU memory requirements.
In terms of the RAG system, LLoCO (Learning Long Contexts Offline)~\cite{Tan2024LLoCOLL} employs the AutoCompressor to compress long documents into multiple summary embeddings stored in the vector database for retrieval.

Another line of work compresses the complete context all at once instead of iterating in segments. There is usually a trainable LLM for encoding (refer to as compressor) and a frozen LLM for decoding (refer to as generator). ICAE (In-context Autoencoder)~\cite{Ge2023IncontextAF} fine-tunes a LoRA-adapted LLM to encode the context with memory tokens into memory slots based on the PWC (Prompt-with-Context) dataset. The generator with memory slots is pre-trained with autoencoding and text continuation tasks to reconstruct the original context and apply it to various downstream tasks. 
500xCompressor~\cite{Li2024500xCompressorGP} similarly fine-tunes LoRA-adapted LLM to compress prompts. The difference is that the generator learns to adapt KV values of compressed prompts as inputs and responses based on them by pre-training with the regeneration task and fine-tuning with the question-answering task. 500xCompressor can compress any text, answer various types of questions and achieve compression ratios ranging from 6x to 480x.
As for LLM-based recommendation, POD (Prompt Distillation)~\cite{Li2023PromptDF} is primarily used for three typical tasks: sequential recommendation, Top-N recommendation and explainable recommendation. The backbone model follows an encoder-decoder architecture where the encoder distills the discrete prompt templates into multiple continuous prompt vectors with an additional whole-word embedding to ensure the integrity of the item ID, the decoder generates recommendations based on these prompt vectors.
Following this, RDRec~\cite{Wang2024RDRecRD} framework consists of two stages to synthesize training data and internalize rationales into a smaller model. In the interaction rationale distillation, the LLM generates rationales including user preferences and item attributes as training labels. In the rationale-aware recommendation, prompt template vectors compressed by POD~\cite{Li2023PromptDF} are concatenated with user and item IDs as training inputs. RDRec achieves SOTA performance in both sequential and Top-N recommendations.

In order to balance training cost, inference efficiency, and generation quality, SelfCP (Self-Compressor)~\cite{Gao2024SelfCPCO} only compresses over-limit prompts to reduce the compression and generation difficulty. There are three components in SelfCP: the compressor and the generator are frozen LLMs, and a trainable connector between them is a 17M linear layer. Different from AutoCompressor with an inefficiency training process and ICAE with limited input length, SelfCP employs decoder-only compressors to compress each over-limit segment asynchronously and concatenates them as the AutoCompressor. Then, the connector is fine-tuned to project the hidden states on top of memory tags into LLM-acceptable memory tokens. As for in-context learning, it is more efficient to directly withdraw memory tokens compressed from target demonstrations.

\begin{table}[]
    \caption{Comparisons between different encoding methods.}
    \label{Tab_encoding}
    \resizebox{\textwidth}{!}{ 
        \begin{tabular}{l|l|l|l|l|ll}
        \toprule[1.5pt]
        \textbf{Encoding Methods} & \textbf{Target Model} & \textbf{Compressor Model} & \textbf{Soft prompt location} & \textbf{Hard prompt} & \multicolumn{2}{c}{\textbf{Soft prompt}} \\
        \midrule
        \makecell[l]{Prompt\\Compression} & Decoder-only & \makecell[l]{Bayesian attribute\\classifier framework} & Before context & Context & Embeddings & Vectors \\
        \midrule
        Gisting & \makecell[l]{Encoder-Decoder\\Decoder-only} & With gist masking & \makecell[l]{Between instruction\\and context} & Instruction & Gist tokens & Vectors\\
        Gist-COCO & Encoder-Decoder & Encoder & Before prompt & Prompt & Gist tokens & Gist representations \\
        UltraGist & Decoder-only & \makecell[l]{With optimized\\cross-attention} & After context segment & Context & Gist tokens & Vectors\\
        \midrule
        AutoCompressor & RMT & RMT & After context segment & Context & Summary tokens & Summary vectors \\
        ICAE & Encoder-Decoder & Encoder (LoRA) & After context & Context & Memory tokens & Memory Slots \\
        500xCompressor & Encoder-Decoder & Encoder (LoRA) & After context & Context & Compressed tokens & K V values \\
        POD & Encoder-Decoder & Encoder & Before context & Context & Embeddings & Vectors \\ 
        RDRec & Encoder-Decoder & Encoder & Before context & Rationale & Embeddings & Vectors \\ 
        \midrule
        SelfCP & Decoder-only & Decoder-only & After context segment & \makecell[l]{Over-limit\\context} & Memory tokens & Vectors \\
        \bottomrule[1.5pt]
        \end{tabular}
    }
\end{table}
Here, we provide comparisons between different encoding methods, as shown in Table \ref{Tab_encoding}. We find that soft prompts are usually prepended to the input sequence in the compressor with Decoder-only architecture while appended to that of Encoder architecture, which is related to the training task mentioned in \S\ref{Prompt_type}.

\subsection{Text-to-Text Compression} \label{Text-to-text_compression}
Text-to-Vector compression in continuous space focuses on the perspective of adapting language models to natural language. Nevertheless, the compressed soft prompts usually lack human readability and interpretability. To further facilitate seamless interaction between LLMs and humans, researchers consider compressing prompts only in discrete space from the perspective of human understanding natural language. The common practice is to utilize accessible language models to measure the information density of hard prompts and then shorten the prompt length as much as possible with stable LLM performance. We conclude two categories of methods as shown in the right half of Fig. \ref{Fig_Prompt_compression}, pruning can be defined as an \textit{extractive} compression method while summarization is an \textit{abstractive} compression method.

\subsubsection{Pruning} \label{Pruning}
Pruning indicates the direct removal of less informative lexical units in different granularity (\ie token, phrase, sentence) without changing the linguistic expression of the original prompt. It measures the amount of information by leveraging smaller language models to calculate information entropy or training discriminator models to determine whether to discard certain tokens in the prompt. According to different filter units, there are three kinds of pruning schemes: coarse-grained, coarse-to-fine grained and fine-grained.

\textbf{Coarse-grained}. \textit{Demonstration-level}: In order to achieve performance-efficiency trade-off of black-box LLMs, DynaICL~\cite{Zhou2023EfficientPV} trains a meta controller to dynamically allocate the number of in-context demonstrations according to the input complexity and the computational budget.
\textit{Sentence-level}: FilCo~\cite{Wang2023LearningTF} employs three lexical and information-theoretic methods—String Inclusion, Lexical Overlap, and Conditional Cross-Mutual Information (CXMI)—to distill retrieved documents into a useful context that trains a context filtering model as well as a generation model for RAG tasks. To maintain the semantic integrity of the context, CPC~\cite{Liskavets2024PromptCW} trains a context-aware sentence encoder to measure the embedding similarity (cosine distance) between the given question and each sentence in the context, and then removes irrelevant sentences for compression.
\textit{Document-level}: AdaComp~\cite{zhang2024adacomp} trains a compression-rate predictor to dynamically select optimal documents based on query complexity and retrieval quality.

\textbf{Coarse-to-fine grained}. Microsoft proposes the LLMLingua series of research to compress prompts for accelerating LLM inference. Instead of considering the prompt as a whole, LLMLingua~\cite{Jiang2023LLMLinguaCP} is the first to separately compress different components (\ie instruction, question, and demonstration) of the prompt in a coarse-to-fine way. First of all, an SLM aligned with the LLM by instruction fine-tuning is prepared to measure token informativeness based on perplexity (PPL). There is a budget controller dynamically allocating compression ratios for different components according to their importance, where the instruction and question are predefined a lower ratio while the higher ratios are left for demonstrations. Coarse-grained compression reduces the number of demonstrations and fine-grained compression iteratively filters tokens with lower PPL in the remained demonstrations.
As the name suggests, LongLLMLingua~\cite{Jiang2023LongLLMLinguaAA} further compress long documents on the basis of the LLMLingua framework. There is a linear scheduler adaptively controlling the fine-grained compression ratio based on the coarse-grained compression ratio. The significant difference is that LongLLMLingua measures the amount of information relevant to the given question, where a reordering mechanism is used to identify more question-aware documents and contrastive perplexity is proposed to retain more question-related tokens. Finally, the subsequence recovery method ensures the integrity of the key information (such as time and location) in the LLM response. With above tricks, compressed prompts can derive higher performance at lower costs while reducing end-to-end system latency.
CoT-Influx~\cite{Huang2023FewerIM} is a plug-and-play coarse-to-fine pruner to specially compress CoT prompts generated by GPT-4~\cite{achiam2023gpt}. The shot-pruner reduces useless CoT examples and the token-pruner filters redundant tokens. Both pruners are a two-layer feed-forward network (MLP) with GELU activation to be trained by RL with multi-objective rewards instead of backpropagation.

\textbf{Fine-grained}. Selective Context~\cite{Li2023CompressingCT} 
evaluates informativeness of lexical units with self- information~\cite{Shannon1948AMT} computed by a causal language model and filters redundant content by a percentile-based approach. It is worth mentioning that token self-information is calculated by predicting the next token probability, which is accumulated in the range of lexical units and then averaged to obtain phrase and sentence level self-information. Phrase has been proven to be the most effective filtering unit that can reduce inference memory usage by 36\% and inference time by 32\%, with negligible performance drop, striking a good balance between efficiency and performance.
In order to retain syntactic and semantic structure, PROMPT-SAW~\cite{Ali2024PROMPTSAWLR} extracts tokens with key information via relation-aware graphs and reinstates them to compressed prompts that achieve better readability and interpretability.
In addition to pruning redundant tokens by explicitly calculating the amount of information, the language model can be trained to determine whether to prune certain tokens or not by themselves. PCRL~\cite{Jung2023DiscretePC} considers prompt compression as a binary classification task, which is performed by a frozen pre-trained policy LM with trainable MLP layers based on RL. The compression policy assigns an \textit{include} or \textit{exclude} label for each token in the prompt and the reward function considers both faithfulness and reduced length of the compressed prompt.
Instead of calculating information entropy by a unidirectional decoder-only model, LLMLinga-2~\cite{Pan2024LLMLingua2DD} employs a bidirectional encoder-only model with a linear classification layer as a compressor to decide whether each token \textit{preserve} or \textit{discard}. The small but general compressor is trained 10 epochs on distilled data synthesized by GPT-4~\cite{achiam2023gpt} to significantly enhance performance, with a 3x-6x improvement in compression speedup, and a 1.6x-2.9x acceleration in end-to-end latency.

\begin{table}[t!]
  \caption{The LLM performance and prompt compression ratio (\textbf{shot} is the number of demonstrations and \textbf{×} is the ratio of the original prompt to the compressed prompt) of various pruning methods in commonly used reasoning tasks and long-context tasks. It is worth noting that, due to variations in the experimental setups of different methods, their performances may be not directly comparable.}
  \label{Tab_T2T_Pruning}
  \resizebox{\textwidth}{!}{ 
  \begin{tabular}{l|l|cc|cccc|ccccc}
  \toprule[1.5pt]
  \multirow{2}{*}{\textbf{Methods}} & \multirow{2}{*}{\makecell{\textbf{Compression}\\\textbf{Granularity}}}  & \multicolumn{2}{c|}{\textbf{NaturalQuestions}} & \multicolumn{2}{c}{\textbf{GSM8K}} & \multicolumn{2}{c|}{\textbf{BBH}} & \multicolumn{2}{c}{\textbf{ZeroSCROLLS}} & \multicolumn{3}{c}{\textbf{LongBench}} \\
  \cmidrule (lr){3-4}     \cmidrule (lr){5-6}     \cmidrule (lr){7-8}     \cmidrule (lr){9-10}     \cmidrule (lr){11-13}
  & & F1 & Ratio & EM & Ratio & EM  & Ratio  & Acc & Ratio & Acc & Ratio & Latency \\
  \midrule
  \multirow{2}{*}{DynalCL}  & \multirow{2}{*}{demonstration} & 42.40(EM) & 10-shot & - & - & - & - & - & - & - & - & - \\
  & & 40.20(EM) & 5-shot  & - & - & - & - & - & - & - & - & - \\
  FliCo  & sentence & 61.80 & 5-shot  & - & - & - & - & - & - & - & - & - \\
  \multirow{2}{*}{CPC}  & \multirow{2}{*}{sentence}  & - & - & - & - & - & - & 34.90 & 3×  & 50.00 & 3×  & 1× \\
  & & - & - & - & - & - & - & 33.80 & 5×  & 49.50 & 5×  & - \\
  AdaComp  & document & 70.96 & 3.66-shot & - & - & - & - & - & - & - & - & - \\
  \midrule
  \multirow{2}{*}{LLMLingua}  & \multirow{2}{*}{demonstration -\textgreater token} & - & - & 79.08 & 5× & 70.11  & 3×  & 30.70 & 3×  & 37.40 & 3×  & 9.8×  \\
  & & 30.00 & 3.8×  & 77.41 & 14×  & 61.60  & 5×  & 27.20 & 5×  & 34.60 & 5×  & - \\
  \multirow{2}{*}{LongLLMLingua} & \multirow{2}{*}{document -\textgreater token}  & 75.50 & 3.9×  & - & - & - & - & 32.80 & 3×  & 48.80 & 3×  & 10.93×  \\
  & & - & - & - & - & - & - & 32.50 & 6×  & 48.00 & 6×  & - \\
  CoT-Influx & CoT -\textgreater token  & - & - & 73.31 & 7.7× & - & - & - & - & - & - & - \\
  \midrule
  \multirow{2}{*}{Selective Context} & \multirow{2}{*}{token, phrase, sentence}  & 43.80 & 3.7×  & 53.98 & 5× & 54.27  & 3×  & 20.70 & 3×  & 32.00 & 3×  & - \\
  & & - & - & 52.99 & 11×  & 54.02  & 5×  & 19.40 & 5×  & 24.80 & 5×  & - \\
  PROMPT-SAW & entity, relation  & 73.22(EM) & 3.86× & 72.12 & 1.49× & - & - & - & - & - & - & - \\
  \multirow{2}{*}{LLMLingua-2} & \multirow{2}{*}{token} & 71.90 & 3.9×  & 79.08 & 5× & 70.02  & 3×  & 33.50 & 3×  & 42.20 & 3×  & 0.67× \\
  & & - & - & 77.79 & 14×  & 61.94  & 5×  & 33.40 & 5×  & 39.10 & 5×  & - \\
  \bottomrule[1.5pt]
  \end{tabular}
  }
\end{table}
Here, we provide an overall empirical insight into pruning methods across different compression granularities on classical benchmarks, as shown in Table \ref{Tab_T2T_Pruning}. We observe that coarse-to-fine compression seems to be more beneficial for complex reasoning tasks while fine-grained compression is more suitable for long context tasks.

\subsubsection{Summarization} \label{Summarization} 
In essence, summarization is a semantic-level compression that may change linguistic expressions while retaining the original idea. To ensure that the LLM performance of the compressed prompt does not deviate significantly from the original prompt, there are two types of methods with and without training will be introduced in the following.

The first type of methods typically take the original output as a supervised signal to train the summarizer.
In RAG scenarios, RECOMP (Retrieve, Compress, Prepend)~\cite{Xu2023RECOMPIR} compresses the retrieved documents into textual summaries prior to in-context integration. There are two query-focus compressors summarizing multi-documents to improve LLM performance. The extractive compressor is a dual encoder model (110M) trained to select sentences with high semantic similarity to the input query by contrastive learning. The abstractive compressor is an encoder-decoder model (775M) to learn the summarization ability of the LLM by knowledge distillation, which realizes selective augmentation instead of prepending irrelevant documents.
Nano-Capsulator~\cite{Chuang2024LearningTC} is a compressor LLM to semantically compress prompts in natural language formats via summarization instructions. The response difference between the original and compressed prompt can be viewed as the reward feedback to monitor the optimization of Nano-Capsulator. There is also a strict cut-off mechanism to restrict the length of the compressed prompt. 

The second type of methods monitor whether the information of the summarized prompt is sufficient to correctly respond in real-time.
MEMWALKER~\cite{Chen2023WalkingDT} compresses long contexts by interactively prompting the LLM instead of fine-tuning. It consists of two stages where the first stage iteratively summarizes context segments to construct a memory tree and the second stage navigates from the root node to search for sufficient relevant information to respond the given query.
CompAct~\cite{Yoon2024CompActCR} mainly addresses question-answering tasks in long context scenarios. The compressor LLM sequentially summarizes each segment of retrieved documents into a compressed prompt, whose information is evaluated to determine whether it is complete to answer the given question. If incomplete, the compressed prompt will be concatenated with the subsequent segment for next-iteration summarization. CompAct can serve as a cost-efficient and flexible plug-in compressor between off-the-shelf retrievers and readers, achieving exceptionally high compression rates (47x).
Style-Compress~\cite{pu2024style} can be seen as an efficient few-shot prompting method combining automatic prompt engineering and prompt compression tricks. A smaller LLM (LLaMA-2-7B~\cite{touvron2023llama}) iteratively compresses the original prompt by prompting with diverse human-written styles and in-context learning with task-specific high-performing examples in turn. Then, the best compressed prompt evaluated by a larger LLM (LLaMA-2-13B / GPT-3.5) and its comparative advantage are added to the demonstrations pool that instructs the compressor to conduct task-specific prompt compression.
\section{Future Directions} \label{Future_directions}
Despite the remarkable advances, the area of LLM prompting still faces several major challenges, which also provide exciting opportunities for future research. We conclude our review with a conjecture of promising areas of future work, specifically highlighting the intersection with two resource-saving prompting strategies. Meanwhile, we point out some open problems in the future research direction.

\begin{itemize}[leftmargin=*]
    \item Combining two resource-saving prompting strategies as a Win-Win solution: Actually, there is a gap between optimization objectives in prompt engineering and prompt compression. The former tends to optimize shorter instructions, while the latter mainly deals with longer demonstrations. Chain-of-Thought (CoT) represents the intersection of the interests of both areas, particularly as it plays a pivotal role in reasoning, which is a key research focus in the pursuit of Artificial General Intelligence (AGI). Future research could improve the CoT prompting framework based on the Eq. \ref{eq_Prompt_Design} and \ref{eq_Hard_Prompt_Compression}. One feasible option is to consider straightforward nesting two strategies. For instance, iteratively self-improving the thought process first, and then compressing useless information from the lengthy CoT. Alternatively, both strategies could be performed synchronously in a single iteration, possibly using reinforcement learning where LLM performance serves as a reward signal to supervise compression. In addition, designing balancing factors to integrate the objective functions Eq. \ref{eq_Prompt_Design} and \ref{eq_Hard_Prompt_Compression} could facilitate analytical work, such as discussing the mutual constraints of optimization and compression levels, as well as how to balance various metrics such as model performance, computational costs, and inference speed.
    
    \item Discussion of differences between LLMs and Humans in understanding natural language: While prompt compression accelerates LLM inference and generally maintains or slightly improves performance, it often struggles to ensure the readability and interpretability of the compressed prompts. So, there is an opportunity to explore the significance of gibberish versus human-interpretable compressed prompts in the context of AGI. Is the information defined from the perspective of human comprehension a reasonable standard for measuring the effective information provided for LLMs? Investigating this distinction could yield valuable insights into optimizing prompts for better alignment with humans.

    \item Robustness of efficient prompting: Optimized instructions typically cater to specific downstream tasks, so it is worth investigating how to enhance the transferability of prompts across different data distributions during the automatic optimization process. For instance, mixing labeled and unlabeled data might facilitate the learning of a more universal prompt. Similarly, T2V-compressed prompts are typically tailored to a specific language model, making it challenging to adapt to other models without fine-tuning. One potential solution is to encapsulate soft prompts within a plug-and-play adapter, enhancing the transferability of encoded representations and achieving both high performance and general capability.

    \item Open questions for future reflection: Limited by the reasoning ability of SLMs, automatic prompt engineering heavily relies on LLMs to optimize prompts~\cite{Zhang2024RevisitingOT}. Hence, both model capabilities and computational costs need to be considered. While automatic prompt engineering significantly alleviates human labor, it still cannot completely avoid human intervention. Specifically, during the optimization of instructions, the initial iteration requires users to provide input-output examples or high-quality initial prompts; during the optimization of demonstrations (especially CoT), careful design of meta-prompt content or frameworks is necessary to correctly guide LLMs in generating appropriate responses.
\end{itemize}
\section{Conclusion}
This survey provides an extensive tour of recent advances in efficient prompting methods for LLMs through the lens of reducing resource consumption.
We comprehensively review two families of existing efficient prompting methods: automatic prompt engineering to avoid human resources and prompt compression to save computational resources. Notably, we distill the optimization objectives of each category from a mathematical perspective and expect to combine both resource-saving strategies for lightweight LLM applications in the future. We further summarize representative methods for each category with a particular emphasis on efficiency. Finally, we outline potential directions for future research and provide a list of open-source efficient prompting projects as shown in Appendix \ref{Open_Resources}.
\begin{acks}
This work was supported in part by the National Science Foundation of China (No.62276056), the Natural Science Foundation of Liaoning Province of China (2022-KF-16-01), the Fundamental Research Funds for the Central Universities (Nos. N2216016 and N2316002), the Yunnan Fundamental Research Projects (No. 202401BC070021), and the Program of Introducing Talents of Discipline to Universities, Plan 111 (No.B16009).
\end{acks}
\bibliographystyle{ACM-Reference-Format}
\bibliography{EPM}


\begin{thebibliography}{127}


\ifx \showCODEN    \undefined \def \showCODEN     #1{\unskip}     \fi
\ifx \showDOI      \undefined \def \showDOI       #1{#1}\fi
\ifx \showISBNx    \undefined \def \showISBNx     #1{\unskip}     \fi
\ifx \showISBNxiii \undefined \def \showISBNxiii  #1{\unskip}     \fi
\ifx \showISSN     \undefined \def \showISSN      #1{\unskip}     \fi
\ifx \showLCCN     \undefined \def \showLCCN      #1{\unskip}     \fi
\ifx \shownote     \undefined \def \shownote      #1{#1}          \fi
\ifx \showarticletitle \undefined \def \showarticletitle #1{#1}   \fi
\ifx \showURL      \undefined \def \showURL       {\relax}        \fi
\providecommand\bibfield[2]{#2}
\providecommand\bibinfo[2]{#2}
\providecommand\natexlab[1]{#1}
\providecommand\showeprint[2][]{arXiv:#2}

\bibitem[Achiam et~al\mbox{.}(2023)]%
        {achiam2023gpt}
\bibfield{author}{\bibinfo{person}{Josh Achiam}, \bibinfo{person}{Steven Adler}, \bibinfo{person}{Sandhini Agarwal}, \bibinfo{person}{Lama Ahmad}, \bibinfo{person}{Ilge Akkaya}, \bibinfo{person}{Florencia~Leoni Aleman}, \bibinfo{person}{Diogo Almeida}, \bibinfo{person}{Janko Altenschmidt}, \bibinfo{person}{Sam Altman}, \bibinfo{person}{Shyamal Anadkat}, {et~al\mbox{.}}} \bibinfo{year}{2023}\natexlab{}.
\newblock \showarticletitle{Gpt-4 technical report}.
\newblock \bibinfo{journal}{\emph{ArXiv preprint}}  \bibinfo{volume}{abs/2303.08774} (\bibinfo{year}{2023}).
\newblock
\urldef\tempurl%
\url{https://arxiv.org/abs/2303.08774}
\showURL{%
\tempurl}


\bibitem[Ali et~al\mbox{.}(2024)]%
        {Ali2024PROMPTSAWLR}
\bibfield{author}{\bibinfo{person}{Muhammad~Asif Ali}, \bibinfo{person}{Zhengping Li}, \bibinfo{person}{Shu Yang}, \bibinfo{person}{Keyuan Cheng}, \bibinfo{person}{Yang Cao}, \bibinfo{person}{Tianhao Huang}, \bibinfo{person}{Lijie Hu}, \bibinfo{person}{Lu Yu}, {and} \bibinfo{person}{Di Wang}.} \bibinfo{year}{2024}\natexlab{}.
\newblock \showarticletitle{PROMPT-SAW: Leveraging Relation-Aware Graphs for Textual Prompt Compression}.
\newblock \bibinfo{journal}{\emph{ArXiv preprint}}  \bibinfo{volume}{abs/2404.00489} (\bibinfo{year}{2024}).
\newblock
\urldef\tempurl%
\url{https://arxiv.org/abs/2404.00489}
\showURL{%
\tempurl}


\bibitem[Anil et~al\mbox{.}(2023)]%
        {anil2023palm}
\bibfield{author}{\bibinfo{person}{Rohan Anil}, \bibinfo{person}{Andrew~M Dai}, \bibinfo{person}{Orhan Firat}, \bibinfo{person}{Melvin Johnson}, \bibinfo{person}{Dmitry Lepikhin}, \bibinfo{person}{Alexandre Passos}, \bibinfo{person}{Siamak Shakeri}, \bibinfo{person}{Emanuel Taropa}, \bibinfo{person}{Paige Bailey}, \bibinfo{person}{Zhifeng Chen}, {et~al\mbox{.}}} \bibinfo{year}{2023}\natexlab{}.
\newblock \showarticletitle{Palm 2 technical report}.
\newblock \bibinfo{journal}{\emph{ArXiv preprint}}  \bibinfo{volume}{abs/2305.10403} (\bibinfo{year}{2023}).
\newblock
\urldef\tempurl%
\url{https://arxiv.org/abs/2305.10403}
\showURL{%
\tempurl}


\bibitem[Askell et~al\mbox{.}(2021)]%
        {Askell2021AGL}
\bibfield{author}{\bibinfo{person}{Amanda Askell}, \bibinfo{person}{Yuntao Bai}, \bibinfo{person}{Anna Chen}, \bibinfo{person}{Dawn Drain}, \bibinfo{person}{Deep Ganguli}, \bibinfo{person}{Tom Henighan}, \bibinfo{person}{Andy Jones}, \bibinfo{person}{Nicholas Joseph}, \bibinfo{person}{Benjamin Mann}, \bibinfo{person}{Nova Dassarma}, \bibinfo{person}{Nelson Elhage}, \bibinfo{person}{Zac Hatfield-Dodds}, \bibinfo{person}{Danny Hernandez}, \bibinfo{person}{John Kernion}, \bibinfo{person}{Kamal Ndousse}, \bibinfo{person}{Catherine Olsson}, \bibinfo{person}{Dario Amodei}, \bibinfo{person}{Tom~B. Brown}, \bibinfo{person}{Jack Clark}, \bibinfo{person}{Sam McCandlish}, \bibinfo{person}{Christopher Olah}, {and} \bibinfo{person}{Jared Kaplan}.} \bibinfo{year}{2021}\natexlab{}.
\newblock \showarticletitle{A General Language Assistant as a Laboratory for Alignment}.
\newblock \bibinfo{journal}{\emph{ArXiv preprint}}  \bibinfo{volume}{abs/2112.00861} (\bibinfo{year}{2021}).
\newblock
\urldef\tempurl%
\url{https://arxiv.org/abs/2112.00861}
\showURL{%
\tempurl}


\bibitem[Besta et~al\mbox{.}(2024)]%
        {Besta2023GraphOT}
\bibfield{author}{\bibinfo{person}{Maciej Besta}, \bibinfo{person}{Nils Blach}, \bibinfo{person}{Ales Kubicek}, \bibinfo{person}{Robert Gerstenberger}, \bibinfo{person}{Michal Podstawski}, \bibinfo{person}{Lukas Gianinazzi}, \bibinfo{person}{Joanna Gajda}, \bibinfo{person}{Tomasz Lehmann}, \bibinfo{person}{Hubert Niewiadomski}, \bibinfo{person}{Piotr Nyczyk}, {and} \bibinfo{person}{Torsten Hoefler}.} \bibinfo{year}{2024}\natexlab{}.
\newblock \showarticletitle{Graph of Thoughts: Solving Elaborate Problems with Large Language Models}. In \bibinfo{booktitle}{\emph{Thirty-Eighth {AAAI} Conference on Artificial Intelligence, {AAAI} 2024, Thirty-Sixth Conference on Innovative Applications of Artificial Intelligence, {IAAI} 2024, Fourteenth Symposium on Educational Advances in Artificial Intelligence, {EAAI} 2014, February 20-27, 2024, Vancouver, Canada}}, \bibfield{editor}{\bibinfo{person}{Michael~J. Wooldridge}, \bibinfo{person}{Jennifer~G. Dy}, {and} \bibinfo{person}{Sriraam Natarajan}} (Eds.). \bibinfo{publisher}{{AAAI} Press}, \bibinfo{pages}{17682--17690}.
\newblock
\urldef\tempurl%
\url{https://doi.org/10.1609/AAAI.V38I16.29720}
\showDOI{\tempurl}


\bibitem[Brown et~al\mbox{.}(2020)]%
        {brown2020language}
\bibfield{author}{\bibinfo{person}{Tom~B. Brown}, \bibinfo{person}{Benjamin Mann}, \bibinfo{person}{Nick Ryder}, \bibinfo{person}{Melanie Subbiah}, \bibinfo{person}{Jared Kaplan}, \bibinfo{person}{Prafulla Dhariwal}, \bibinfo{person}{Arvind Neelakantan}, \bibinfo{person}{Pranav Shyam}, \bibinfo{person}{Girish Sastry}, \bibinfo{person}{Amanda Askell}, \bibinfo{person}{Sandhini Agarwal}, \bibinfo{person}{Ariel Herbert{-}Voss}, \bibinfo{person}{Gretchen Krueger}, \bibinfo{person}{Tom Henighan}, \bibinfo{person}{Rewon Child}, \bibinfo{person}{Aditya Ramesh}, \bibinfo{person}{Daniel~M. Ziegler}, \bibinfo{person}{Jeffrey Wu}, \bibinfo{person}{Clemens Winter}, \bibinfo{person}{Christopher Hesse}, \bibinfo{person}{Mark Chen}, \bibinfo{person}{Eric Sigler}, \bibinfo{person}{Mateusz Litwin}, \bibinfo{person}{Scott Gray}, \bibinfo{person}{Benjamin Chess}, \bibinfo{person}{Jack Clark}, \bibinfo{person}{Christopher Berner}, \bibinfo{person}{Sam McCandlish}, \bibinfo{person}{Alec Radford}, \bibinfo{person}{Ilya Sutskever},
  {and} \bibinfo{person}{Dario Amodei}.} \bibinfo{year}{2020}\natexlab{}.
\newblock \showarticletitle{Language Models are Few-Shot Learners}. In \bibinfo{booktitle}{\emph{Advances in Neural Information Processing Systems 33: Annual Conference on Neural Information Processing Systems 2020, NeurIPS 2020, December 6-12, 2020, virtual}}, \bibfield{editor}{\bibinfo{person}{Hugo Larochelle}, \bibinfo{person}{Marc'Aurelio Ranzato}, \bibinfo{person}{Raia Hadsell}, \bibinfo{person}{Maria{-}Florina Balcan}, {and} \bibinfo{person}{Hsuan{-}Tien Lin}} (Eds.).
\newblock
\urldef\tempurl%
\url{https://proceedings.neurips.cc/paper/2020/hash/1457c0d6bfcb4967418bfb8ac142f64a-Abstract.html}
\showURL{%
\tempurl}


\bibitem[Bulatov et~al\mbox{.}(2022)]%
        {Bulatov2022RecurrentMT}
\bibfield{author}{\bibinfo{person}{Aydar Bulatov}, \bibinfo{person}{Yuri Kuratov}, {and} \bibinfo{person}{Mikhail Burtsev}.} \bibinfo{year}{2022}\natexlab{}.
\newblock \showarticletitle{Recurrent Memory Transformer}. In \bibinfo{booktitle}{\emph{Advances in Neural Information Processing Systems 35: Annual Conference on Neural Information Processing Systems 2022, NeurIPS 2022, New Orleans, LA, USA, November 28 - December 9, 2022}}, \bibfield{editor}{\bibinfo{person}{Sanmi Koyejo}, \bibinfo{person}{S.~Mohamed}, \bibinfo{person}{A.~Agarwal}, \bibinfo{person}{Danielle Belgrave}, \bibinfo{person}{K.~Cho}, {and} \bibinfo{person}{A.~Oh}} (Eds.).
\newblock
\urldef\tempurl%
\url{http://papers.nips.cc/paper\_files/paper/2022/hash/47e288629a6996a17ce50b90a056a0e1-Abstract-Conference.html}
\showURL{%
\tempurl}


\bibitem[Chen et~al\mbox{.}(2023a)]%
        {Chen2023EvoPromptingLM}
\bibfield{author}{\bibinfo{person}{Angelica Chen}, \bibinfo{person}{David Dohan}, {and} \bibinfo{person}{David~R. So}.} \bibinfo{year}{2023}\natexlab{a}.
\newblock \showarticletitle{EvoPrompting: Language Models for Code-Level Neural Architecture Search}. In \bibinfo{booktitle}{\emph{Advances in Neural Information Processing Systems 36: Annual Conference on Neural Information Processing Systems 2023, NeurIPS 2023, New Orleans, LA, USA, December 10 - 16, 2023}}, \bibfield{editor}{\bibinfo{person}{Alice Oh}, \bibinfo{person}{Tristan Naumann}, \bibinfo{person}{Amir Globerson}, \bibinfo{person}{Kate Saenko}, \bibinfo{person}{Moritz Hardt}, {and} \bibinfo{person}{Sergey Levine}} (Eds.).
\newblock
\urldef\tempurl%
\url{http://papers.nips.cc/paper\_files/paper/2023/hash/184c1e18d00d7752805324da48ad25be-Abstract-Conference.html}
\showURL{%
\tempurl}


\bibitem[Chen et~al\mbox{.}(2023b)]%
        {Chen2023WalkingDT}
\bibfield{author}{\bibinfo{person}{Howard Chen}, \bibinfo{person}{Ramakanth Pasunuru}, \bibinfo{person}{Jason Weston}, {and} \bibinfo{person}{Asli Celikyilmaz}.} \bibinfo{year}{2023}\natexlab{b}.
\newblock \showarticletitle{Walking Down the Memory Maze: Beyond Context Limit through Interactive Reading}.
\newblock \bibinfo{journal}{\emph{ArXiv preprint}}  \bibinfo{volume}{abs/2310.05029} (\bibinfo{year}{2023}).
\newblock
\urldef\tempurl%
\url{https://arxiv.org/abs/2310.05029}
\showURL{%
\tempurl}


\bibitem[Chen(2015)]%
        {chen2015xgboost}
\bibfield{author}{\bibinfo{person}{T Chen}.} \bibinfo{year}{2015}\natexlab{}.
\newblock \showarticletitle{Xgboost: extreme gradient boosting}.
\newblock \bibinfo{journal}{\emph{R package version 0.4-2}} \bibinfo{volume}{1}, \bibinfo{number}{4} (\bibinfo{year}{2015}).
\newblock


\bibitem[Chen et~al\mbox{.}(2016)]%
        {Chen2016TrainingDN}
\bibfield{author}{\bibinfo{person}{Tianqi Chen}, \bibinfo{person}{Bing Xu}, \bibinfo{person}{Chiyuan Zhang}, {and} \bibinfo{person}{Carlos Guestrin}.} \bibinfo{year}{2016}\natexlab{}.
\newblock \showarticletitle{Training Deep Nets with Sublinear Memory Cost}.
\newblock \bibinfo{journal}{\emph{ArXiv preprint}}  \bibinfo{volume}{abs/1604.06174} (\bibinfo{year}{2016}).
\newblock
\urldef\tempurl%
\url{https://arxiv.org/abs/1604.06174}
\showURL{%
\tempurl}


\bibitem[Chen et~al\mbox{.}(2024b)]%
        {Chen2024RePromptPB}
\bibfield{author}{\bibinfo{person}{Weizhe Chen}, \bibinfo{person}{Sven Koenig}, {and} \bibinfo{person}{Bistra~N. Dilkina}.} \bibinfo{year}{2024}\natexlab{b}.
\newblock \showarticletitle{RePrompt: Planning by Automatic Prompt Engineering for Large Language Models Agents}.
\newblock \bibinfo{journal}{\emph{ArXiv preprint}}  \bibinfo{volume}{abs/2406.11132} (\bibinfo{year}{2024}).
\newblock
\urldef\tempurl%
\url{https://arxiv.org/abs/2406.11132}
\showURL{%
\tempurl}


\bibitem[Chen et~al\mbox{.}(2022)]%
        {Chen2022ProgramOT}
\bibfield{author}{\bibinfo{person}{Wenhu Chen}, \bibinfo{person}{Xueguang Ma}, \bibinfo{person}{Xinyi Wang}, {and} \bibinfo{person}{William~W. Cohen}.} \bibinfo{year}{2022}\natexlab{}.
\newblock \showarticletitle{Program of Thoughts Prompting: Disentangling Computation from Reasoning for Numerical Reasoning Tasks}.
\newblock \bibinfo{journal}{\emph{Trans. Mach. Learn. Res.}}  \bibinfo{volume}{2023} (\bibinfo{year}{2022}).
\newblock
\urldef\tempurl%
\url{https://api.semanticscholar.org/CorpusID:253801709}
\showURL{%
\tempurl}


\bibitem[Chen et~al\mbox{.}(2024a)]%
        {Chen2024PRomptOI}
\bibfield{author}{\bibinfo{person}{Yongchao Chen}, \bibinfo{person}{Jacob Arkin}, \bibinfo{person}{Yilun Hao}, \bibinfo{person}{Yang Zhang}, \bibinfo{person}{Nicholas Roy}, {and} \bibinfo{person}{Chuchu Fan}.} \bibinfo{year}{2024}\natexlab{a}.
\newblock \showarticletitle{PRompt Optimization in Multi-Step Tasks (PROMST): Integrating Human Feedback and Heuristic-based Sampling}.
\newblock
\urldef\tempurl%
\url{https://api.semanticscholar.org/CorpusID:270559211}
\showURL{%
\tempurl}


\bibitem[Cheng et~al\mbox{.}(2023)]%
        {Cheng2023BlackBoxPO}
\bibfield{author}{\bibinfo{person}{Jiale Cheng}, \bibinfo{person}{Xiao Liu}, \bibinfo{person}{Kehan Zheng}, \bibinfo{person}{Pei Ke}, \bibinfo{person}{Hongning Wang}, \bibinfo{person}{Yuxiao Dong}, \bibinfo{person}{Jie Tang}, {and} \bibinfo{person}{Minlie Huang}.} \bibinfo{year}{2023}\natexlab{}.
\newblock \showarticletitle{Black-Box Prompt Optimization: Aligning Large Language Models without Model Training}.
\newblock \bibinfo{journal}{\emph{ArXiv preprint}}  \bibinfo{volume}{abs/2311.04155} (\bibinfo{year}{2023}).
\newblock
\urldef\tempurl%
\url{https://arxiv.org/abs/2311.04155}
\showURL{%
\tempurl}


\bibitem[Cheng et~al\mbox{.}(2024)]%
        {Cheng2024xRAGEC}
\bibfield{author}{\bibinfo{person}{Xin Cheng}, \bibinfo{person}{Xun Wang}, \bibinfo{person}{Xingxing Zhang}, \bibinfo{person}{Tao Ge}, \bibinfo{person}{Si-Qing Chen}, \bibinfo{person}{Furu Wei}, \bibinfo{person}{Huishuai Zhang}, {and} \bibinfo{person}{Dongyan Zhao}.} \bibinfo{year}{2024}\natexlab{}.
\newblock \showarticletitle{xRAG: Extreme Context Compression for Retrieval-augmented Generation with One Token}.
\newblock \bibinfo{journal}{\emph{ArXiv preprint}}  \bibinfo{volume}{abs/2405.13792} (\bibinfo{year}{2024}).
\newblock
\urldef\tempurl%
\url{https://arxiv.org/abs/2405.13792}
\showURL{%
\tempurl}


\bibitem[Chevalier et~al\mbox{.}(2023)]%
        {Chevalier2023AdaptingLM}
\bibfield{author}{\bibinfo{person}{Alexis Chevalier}, \bibinfo{person}{Alexander Wettig}, \bibinfo{person}{Anirudh Ajith}, {and} \bibinfo{person}{Danqi Chen}.} \bibinfo{year}{2023}\natexlab{}.
\newblock \showarticletitle{Adapting Language Models to Compress Contexts}. In \bibinfo{booktitle}{\emph{Proceedings of the 2023 Conference on Empirical Methods in Natural Language Processing}}, \bibfield{editor}{\bibinfo{person}{Houda Bouamor}, \bibinfo{person}{Juan Pino}, {and} \bibinfo{person}{Kalika Bali}} (Eds.). \bibinfo{publisher}{Association for Computational Linguistics}, \bibinfo{address}{Singapore}, \bibinfo{pages}{3829--3846}.
\newblock
\urldef\tempurl%
\url{https://doi.org/10.18653/v1/2023.emnlp-main.232}
\showDOI{\tempurl}


\bibitem[Choi et~al\mbox{.}(2022)]%
        {Choi2022PromptIP}
\bibfield{author}{\bibinfo{person}{Eunbi Choi}, \bibinfo{person}{Yongrae Jo}, \bibinfo{person}{Joel Jang}, {and} \bibinfo{person}{Minjoon Seo}.} \bibinfo{year}{2022}\natexlab{}.
\newblock \showarticletitle{Prompt Injection: Parameterization of Fixed Inputs}.
\newblock \bibinfo{journal}{\emph{ArXiv preprint}}  \bibinfo{volume}{abs/2206.11349} (\bibinfo{year}{2022}).
\newblock
\urldef\tempurl%
\url{https://arxiv.org/abs/2206.11349}
\showURL{%
\tempurl}


\bibitem[Chuang et~al\mbox{.}(2024)]%
        {Chuang2024LearningTC}
\bibfield{author}{\bibinfo{person}{Yu-Neng Chuang}, \bibinfo{person}{Tianwei Xing}, \bibinfo{person}{Chia-Yuan Chang}, \bibinfo{person}{Zirui Liu}, \bibinfo{person}{Xun Chen}, {and} \bibinfo{person}{Xia Hu}.} \bibinfo{year}{2024}\natexlab{}.
\newblock \showarticletitle{Learning to Compress Prompt in Natural Language Formats}. In \bibinfo{booktitle}{\emph{Proceedings of the 2024 Conference of the North American Chapter of the Association for Computational Linguistics: Human Language Technologies (Volume 1: Long Papers)}}, \bibfield{editor}{\bibinfo{person}{Kevin Duh}, \bibinfo{person}{Helena Gomez}, {and} \bibinfo{person}{Steven Bethard}} (Eds.). \bibinfo{publisher}{Association for Computational Linguistics}, \bibinfo{address}{Mexico City, Mexico}, \bibinfo{pages}{7756--7767}.
\newblock
\urldef\tempurl%
\url{https://aclanthology.org/2024.naacl-long.429}
\showURL{%
\tempurl}


\bibitem[Cui et~al\mbox{.}(2024)]%
        {cui2024phaseevo}
\bibfield{author}{\bibinfo{person}{Wendi Cui}, \bibinfo{person}{Jiaxin Zhang}, \bibinfo{person}{Zhuohang Li}, \bibinfo{person}{Hao Sun}, \bibinfo{person}{Damien Lopez}, \bibinfo{person}{Kamalika Das}, \bibinfo{person}{Bradley Malin}, {and} \bibinfo{person}{Sricharan Kumar}.} \bibinfo{year}{2024}\natexlab{}.
\newblock \showarticletitle{PhaseEvo: Towards Unified In-Context Prompt Optimization for Large Language Models}.
\newblock \bibinfo{journal}{\emph{ArXiv preprint}}  \bibinfo{volume}{abs/2402.11347} (\bibinfo{year}{2024}).
\newblock
\urldef\tempurl%
\url{https://arxiv.org/abs/2402.11347}
\showURL{%
\tempurl}


\bibitem[Deng et~al\mbox{.}(2022)]%
        {Deng2022RLPromptOD}
\bibfield{author}{\bibinfo{person}{Mingkai Deng}, \bibinfo{person}{Jianyu Wang}, \bibinfo{person}{Cheng-Ping Hsieh}, \bibinfo{person}{Yihan Wang}, \bibinfo{person}{Han Guo}, \bibinfo{person}{Tianmin Shu}, \bibinfo{person}{Meng Song}, \bibinfo{person}{Eric Xing}, {and} \bibinfo{person}{Zhiting Hu}.} \bibinfo{year}{2022}\natexlab{}.
\newblock \showarticletitle{{RLP}rompt: Optimizing Discrete Text Prompts with Reinforcement Learning}. In \bibinfo{booktitle}{\emph{Proceedings of the 2022 Conference on Empirical Methods in Natural Language Processing}}, \bibfield{editor}{\bibinfo{person}{Yoav Goldberg}, \bibinfo{person}{Zornitsa Kozareva}, {and} \bibinfo{person}{Yue Zhang}} (Eds.). \bibinfo{publisher}{Association for Computational Linguistics}, \bibinfo{address}{Abu Dhabi, United Arab Emirates}, \bibinfo{pages}{3369--3391}.
\newblock
\urldef\tempurl%
\url{https://doi.org/10.18653/v1/2022.emnlp-main.222}
\showDOI{\tempurl}


\bibitem[Devlin et~al\mbox{.}(2019)]%
        {Devlin2019BERTPO}
\bibfield{author}{\bibinfo{person}{Jacob Devlin}, \bibinfo{person}{Ming-Wei Chang}, \bibinfo{person}{Kenton Lee}, {and} \bibinfo{person}{Kristina Toutanova}.} \bibinfo{year}{2019}\natexlab{}.
\newblock \showarticletitle{{BERT}: Pre-training of Deep Bidirectional Transformers for Language Understanding}. In \bibinfo{booktitle}{\emph{Proceedings of the 2019 Conference of the North {A}merican Chapter of the Association for Computational Linguistics: Human Language Technologies, Volume 1 (Long and Short Papers)}}, \bibfield{editor}{\bibinfo{person}{Jill Burstein}, \bibinfo{person}{Christy Doran}, {and} \bibinfo{person}{Thamar Solorio}} (Eds.). \bibinfo{publisher}{Association for Computational Linguistics}, \bibinfo{address}{Minneapolis, Minnesota}, \bibinfo{pages}{4171--4186}.
\newblock
\urldef\tempurl%
\url{https://doi.org/10.18653/v1/N19-1423}
\showDOI{\tempurl}


\bibitem[Dong et~al\mbox{.}(2023)]%
        {dong2023pace}
\bibfield{author}{\bibinfo{person}{Yihong Dong}, \bibinfo{person}{Kangcheng Luo}, \bibinfo{person}{Xue Jiang}, \bibinfo{person}{Zhi Jin}, {and} \bibinfo{person}{Ge Li}.} \bibinfo{year}{2023}\natexlab{}.
\newblock \showarticletitle{PACE: Improving Prompt with Actor-Critic Editing for Large Language Model}.
\newblock \bibinfo{journal}{\emph{ArXiv preprint}}  \bibinfo{volume}{abs/2308.10088} (\bibinfo{year}{2023}).
\newblock
\urldef\tempurl%
\url{https://arxiv.org/abs/2308.10088}
\showURL{%
\tempurl}


\bibitem[Fernando et~al\mbox{.}(2023)]%
        {Fernando2023PromptbreederSS}
\bibfield{author}{\bibinfo{person}{Chrisantha Fernando}, \bibinfo{person}{Dylan Banarse}, \bibinfo{person}{Henryk Michalewski}, \bibinfo{person}{Simon Osindero}, {and} \bibinfo{person}{Tim Rockt{\"a}schel}.} \bibinfo{year}{2023}\natexlab{}.
\newblock \showarticletitle{Promptbreeder: Self-Referential Self-Improvement Via Prompt Evolution}.
\newblock \bibinfo{journal}{\emph{ArXiv preprint}}  \bibinfo{volume}{abs/2309.16797} (\bibinfo{year}{2023}).
\newblock
\urldef\tempurl%
\url{https://arxiv.org/abs/2309.16797}
\showURL{%
\tempurl}


\bibitem[Gao et~al\mbox{.}(2024)]%
        {Gao2024SelfCPCO}
\bibfield{author}{\bibinfo{person}{Jun Gao}, \bibinfo{person}{Ziqiang Cao}, {and} \bibinfo{person}{Wenjie Li}.} \bibinfo{year}{2024}\natexlab{}.
\newblock \showarticletitle{SelfCP: Compressing over-limit prompt via the frozen large language model itself}.
\newblock \bibinfo{journal}{\emph{Information Processing \& Management}} (\bibinfo{year}{2024}).
\newblock
\urldef\tempurl%
\url{https://api.semanticscholar.org/CorpusID:270063106}
\showURL{%
\tempurl}


\bibitem[Ge et~al\mbox{.}(2023)]%
        {Ge2023IncontextAF}
\bibfield{author}{\bibinfo{person}{Tao Ge}, \bibinfo{person}{Jing Hu}, \bibinfo{person}{Xun Wang}, \bibinfo{person}{Si-Qing Chen}, {and} \bibinfo{person}{Furu Wei}.} \bibinfo{year}{2023}\natexlab{}.
\newblock \showarticletitle{In-context Autoencoder for Context Compression in a Large Language Model}.
\newblock \bibinfo{journal}{\emph{ArXiv preprint}}  \bibinfo{volume}{abs/2307.06945} (\bibinfo{year}{2023}).
\newblock
\urldef\tempurl%
\url{https://arxiv.org/abs/2307.06945}
\showURL{%
\tempurl}


\bibitem[Gr{\"u}nwald(2007)]%
        {Grnwald2007TheMD}
\bibfield{author}{\bibinfo{person}{Peter~D. Gr{\"u}nwald}.} \bibinfo{year}{2007}\natexlab{}.
\newblock \showarticletitle{The Minimum Description Length Principle (Adaptive Computation and Machine Learning)}.
\newblock
\urldef\tempurl%
\url{https://api.semanticscholar.org/CorpusID:119390683}
\showURL{%
\tempurl}


\bibitem[Guo et~al\mbox{.}(2023)]%
        {Guo2023ConnectingLL}
\bibfield{author}{\bibinfo{person}{Qingyan Guo}, \bibinfo{person}{Rui Wang}, \bibinfo{person}{Junliang Guo}, \bibinfo{person}{Bei Li}, \bibinfo{person}{Kaitao Song}, \bibinfo{person}{Xu Tan}, \bibinfo{person}{Guoqing Liu}, \bibinfo{person}{Jiang Bian}, \bibinfo{person}{Yujiu Yang}, \bibinfo{person}{Tsinghua University}, {and} \bibinfo{person}{Microsoft Research}.} \bibinfo{year}{2023}\natexlab{}.
\newblock \showarticletitle{Connecting Large Language Models with Evolutionary Algorithms Yields Powerful Prompt Optimizers}.
\newblock \bibinfo{journal}{\emph{ArXiv preprint}}  \bibinfo{volume}{abs/2309.08532} (\bibinfo{year}{2023}).
\newblock
\urldef\tempurl%
\url{https://arxiv.org/abs/2309.08532}
\showURL{%
\tempurl}


\bibitem[Hinton et~al\mbox{.}(2015)]%
        {Hinton2015DistillingTK}
\bibfield{author}{\bibinfo{person}{Geoffrey~E. Hinton}, \bibinfo{person}{Oriol Vinyals}, {and} \bibinfo{person}{Jeffrey Dean}.} \bibinfo{year}{2015}\natexlab{}.
\newblock \showarticletitle{Distilling the Knowledge in a Neural Network}.
\newblock \bibinfo{journal}{\emph{ArXiv preprint}}  \bibinfo{volume}{abs/1503.02531} (\bibinfo{year}{2015}).
\newblock
\urldef\tempurl%
\url{https://arxiv.org/abs/1503.02531}
\showURL{%
\tempurl}


\bibitem[Hsieh et~al\mbox{.}(2023b)]%
        {hsieh2023automatic}
\bibfield{author}{\bibinfo{person}{Cho-Jui Hsieh}, \bibinfo{person}{Si Si}, \bibinfo{person}{Felix~X Yu}, {and} \bibinfo{person}{Inderjit~S Dhillon}.} \bibinfo{year}{2023}\natexlab{b}.
\newblock \showarticletitle{Automatic engineering of long prompts}.
\newblock \bibinfo{journal}{\emph{ArXiv preprint}}  \bibinfo{volume}{abs/2311.10117} (\bibinfo{year}{2023}).
\newblock
\urldef\tempurl%
\url{https://arxiv.org/abs/2311.10117}
\showURL{%
\tempurl}


\bibitem[Hsieh et~al\mbox{.}(2023a)]%
        {Hsieh2023DistillingSO}
\bibfield{author}{\bibinfo{person}{Cheng-Yu Hsieh}, \bibinfo{person}{Chun-Liang Li}, \bibinfo{person}{Chih-kuan Yeh}, \bibinfo{person}{Hootan Nakhost}, \bibinfo{person}{Yasuhisa Fujii}, \bibinfo{person}{Alex Ratner}, \bibinfo{person}{Ranjay Krishna}, \bibinfo{person}{Chen-Yu Lee}, {and} \bibinfo{person}{Tomas Pfister}.} \bibinfo{year}{2023}\natexlab{a}.
\newblock \showarticletitle{Distilling Step-by-Step! Outperforming Larger Language Models with Less Training Data and Smaller Model Sizes}. In \bibinfo{booktitle}{\emph{Findings of the Association for Computational Linguistics: ACL 2023}}, \bibfield{editor}{\bibinfo{person}{Anna Rogers}, \bibinfo{person}{Jordan Boyd-Graber}, {and} \bibinfo{person}{Naoaki Okazaki}} (Eds.). \bibinfo{publisher}{Association for Computational Linguistics}, \bibinfo{address}{Toronto, Canada}, \bibinfo{pages}{8003--8017}.
\newblock
\urldef\tempurl%
\url{https://doi.org/10.18653/v1/2023.findings-acl.507}
\showDOI{\tempurl}


\bibitem[Huang et~al\mbox{.}(2023a)]%
        {huang2022large}
\bibfield{author}{\bibinfo{person}{Jiaxin Huang}, \bibinfo{person}{Shixiang Gu}, \bibinfo{person}{Le Hou}, \bibinfo{person}{Yuexin Wu}, \bibinfo{person}{Xuezhi Wang}, \bibinfo{person}{Hongkun Yu}, {and} \bibinfo{person}{Jiawei Han}.} \bibinfo{year}{2023}\natexlab{a}.
\newblock \showarticletitle{Large Language Models Can Self-Improve}. In \bibinfo{booktitle}{\emph{Proceedings of the 2023 Conference on Empirical Methods in Natural Language Processing}}, \bibfield{editor}{\bibinfo{person}{Houda Bouamor}, \bibinfo{person}{Juan Pino}, {and} \bibinfo{person}{Kalika Bali}} (Eds.). \bibinfo{publisher}{Association for Computational Linguistics}, \bibinfo{address}{Singapore}, \bibinfo{pages}{1051--1068}.
\newblock
\urldef\tempurl%
\url{https://doi.org/10.18653/v1/2023.emnlp-main.67}
\showDOI{\tempurl}


\bibitem[Huang et~al\mbox{.}(2023b)]%
        {Huang2023FewerIM}
\bibfield{author}{\bibinfo{person}{Xijie Huang}, \bibinfo{person}{Li~Lyna Zhang}, \bibinfo{person}{Kwang-Ting Cheng}, \bibinfo{person}{Fan Yang}, {and} \bibinfo{person}{Mao Yang}.} \bibinfo{year}{2023}\natexlab{b}.
\newblock \showarticletitle{Fewer is More: Boosting LLM Reasoning with Reinforced Context Pruning}.
\newblock
\urldef\tempurl%
\url{https://api.semanticscholar.org/CorpusID:266210460}
\showURL{%
\tempurl}


\bibitem[Jiang et~al\mbox{.}(2023a)]%
        {Jiang2023LLMLinguaCP}
\bibfield{author}{\bibinfo{person}{Huiqiang Jiang}, \bibinfo{person}{Qianhui Wu}, \bibinfo{person}{Chin-Yew Lin}, \bibinfo{person}{Yuqing Yang}, {and} \bibinfo{person}{Lili Qiu}.} \bibinfo{year}{2023}\natexlab{a}.
\newblock \showarticletitle{{LLML}ingua: Compressing Prompts for Accelerated Inference of Large Language Models}. In \bibinfo{booktitle}{\emph{Proceedings of the 2023 Conference on Empirical Methods in Natural Language Processing}}, \bibfield{editor}{\bibinfo{person}{Houda Bouamor}, \bibinfo{person}{Juan Pino}, {and} \bibinfo{person}{Kalika Bali}} (Eds.). \bibinfo{publisher}{Association for Computational Linguistics}, \bibinfo{address}{Singapore}, \bibinfo{pages}{13358--13376}.
\newblock
\urldef\tempurl%
\url{https://doi.org/10.18653/v1/2023.emnlp-main.825}
\showDOI{\tempurl}


\bibitem[Jiang et~al\mbox{.}(2023b)]%
        {Jiang2023LongLLMLinguaAA}
\bibfield{author}{\bibinfo{person}{Huiqiang Jiang}, \bibinfo{person}{Qianhui Wu}, \bibinfo{person}{Xufang Luo}, \bibinfo{person}{Dongsheng Li}, \bibinfo{person}{Chin-Yew Lin}, \bibinfo{person}{Yuqing Yang}, {and} \bibinfo{person}{Lili Qiu}.} \bibinfo{year}{2023}\natexlab{b}.
\newblock \showarticletitle{LongLLMLingua: Accelerating and Enhancing LLMs in Long Context Scenarios via Prompt Compression}.
\newblock \bibinfo{journal}{\emph{ArXiv preprint}}  \bibinfo{volume}{abs/2310.06839} (\bibinfo{year}{2023}).
\newblock
\urldef\tempurl%
\url{https://arxiv.org/abs/2310.06839}
\showURL{%
\tempurl}


\bibitem[Jiang et~al\mbox{.}(2020)]%
        {jiang2020can}
\bibfield{author}{\bibinfo{person}{Zhengbao Jiang}, \bibinfo{person}{Frank~F. Xu}, \bibinfo{person}{Jun Araki}, {and} \bibinfo{person}{Graham Neubig}.} \bibinfo{year}{2020}\natexlab{}.
\newblock \showarticletitle{How Can We Know What Language Models Know?}
\newblock \bibinfo{journal}{\emph{Transactions of the Association for Computational Linguistics}}  \bibinfo{volume}{8} (\bibinfo{year}{2020}), \bibinfo{pages}{423--438}.
\newblock
\urldef\tempurl%
\url{https://doi.org/10.1162/tacl_a_00324}
\showDOI{\tempurl}


\bibitem[Jin et~al\mbox{.}(2024)]%
        {jin2024apeer}
\bibfield{author}{\bibinfo{person}{Can Jin}, \bibinfo{person}{Hongwu Peng}, \bibinfo{person}{Shiyu Zhao}, \bibinfo{person}{Zhenting Wang}, \bibinfo{person}{Wujiang Xu}, \bibinfo{person}{Ligong Han}, \bibinfo{person}{Jiahui Zhao}, \bibinfo{person}{Kai Zhong}, \bibinfo{person}{Sanguthevar Rajasekaran}, {and} \bibinfo{person}{Dimitris~N Metaxas}.} \bibinfo{year}{2024}\natexlab{}.
\newblock \showarticletitle{APEER: Automatic Prompt Engineering Enhances Large Language Model Reranking}.
\newblock \bibinfo{journal}{\emph{ArXiv preprint}}  \bibinfo{volume}{abs/2406.14449} (\bibinfo{year}{2024}).
\newblock
\urldef\tempurl%
\url{https://arxiv.org/abs/2406.14449}
\showURL{%
\tempurl}


\bibitem[Juneja et~al\mbox{.}(2024)]%
        {juneja2024task}
\bibfield{author}{\bibinfo{person}{Gurusha Juneja}, \bibinfo{person}{Nagarajan Natarajan}, \bibinfo{person}{Hua Li}, \bibinfo{person}{Jian Jiao}, {and} \bibinfo{person}{Amit Sharma}.} \bibinfo{year}{2024}\natexlab{}.
\newblock \showarticletitle{Task Facet Learning: A Structured Approach to Prompt Optimization}.
\newblock \bibinfo{journal}{\emph{ArXiv preprint}}  \bibinfo{volume}{abs/2406.10504} (\bibinfo{year}{2024}).
\newblock
\urldef\tempurl%
\url{https://arxiv.org/abs/2406.10504}
\showURL{%
\tempurl}


\bibitem[Jung and Kim(2023)]%
        {Jung2023DiscretePC}
\bibfield{author}{\bibinfo{person}{Hoyoun Jung} {and} \bibinfo{person}{Kyung-Joong Kim}.} \bibinfo{year}{2023}\natexlab{}.
\newblock \showarticletitle{Discrete Prompt Compression With Reinforcement Learning}.
\newblock \bibinfo{journal}{\emph{IEEE Access}}  \bibinfo{volume}{12} (\bibinfo{year}{2023}), \bibinfo{pages}{72578--72587}.
\newblock
\urldef\tempurl%
\url{https://api.semanticscholar.org/CorpusID:261030884}
\showURL{%
\tempurl}


\bibitem[Kojima et~al\mbox{.}(2022)]%
        {Kojima2022LargeLM}
\bibfield{author}{\bibinfo{person}{Takeshi Kojima}, \bibinfo{person}{Shixiang~Shane Gu}, \bibinfo{person}{Machel Reid}, \bibinfo{person}{Yutaka Matsuo}, {and} \bibinfo{person}{Yusuke Iwasawa}.} \bibinfo{year}{2022}\natexlab{}.
\newblock \showarticletitle{Large Language Models are Zero-Shot Reasoners}. In \bibinfo{booktitle}{\emph{Advances in Neural Information Processing Systems 35: Annual Conference on Neural Information Processing Systems 2022, NeurIPS 2022, New Orleans, LA, USA, November 28 - December 9, 2022}}, \bibfield{editor}{\bibinfo{person}{Sanmi Koyejo}, \bibinfo{person}{S.~Mohamed}, \bibinfo{person}{A.~Agarwal}, \bibinfo{person}{Danielle Belgrave}, \bibinfo{person}{K.~Cho}, {and} \bibinfo{person}{A.~Oh}} (Eds.).
\newblock
\urldef\tempurl%
\url{http://papers.nips.cc/paper\_files/paper/2022/hash/8bb0d291acd4acf06ef112099c16f326-Abstract-Conference.html}
\showURL{%
\tempurl}


\bibitem[Kong et~al\mbox{.}(2024)]%
        {kong2024prewrite}
\bibfield{author}{\bibinfo{person}{Weize Kong}, \bibinfo{person}{Spurthi~Amba Hombaiah}, \bibinfo{person}{Mingyang Zhang}, \bibinfo{person}{Qiaozhu Mei}, {and} \bibinfo{person}{Michael Bendersky}.} \bibinfo{year}{2024}\natexlab{}.
\newblock \showarticletitle{PRewrite: Prompt Rewriting with Reinforcement Learning}.
\newblock \bibinfo{journal}{\emph{ArXiv preprint}}  \bibinfo{volume}{abs/2401.08189} (\bibinfo{year}{2024}).
\newblock
\urldef\tempurl%
\url{https://arxiv.org/abs/2401.08189}
\showURL{%
\tempurl}


\bibitem[Kumar et~al\mbox{.}(2024)]%
        {kumar2024sculpt}
\bibfield{author}{\bibinfo{person}{Shanu Kumar}, \bibinfo{person}{Akhila~Yesantarao Venkata}, \bibinfo{person}{Shubhanshu Khandelwal}, \bibinfo{person}{Bishal Santra}, \bibinfo{person}{Parag Agrawal}, {and} \bibinfo{person}{Manish Gupta}.} \bibinfo{year}{2024}\natexlab{}.
\newblock \showarticletitle{SCULPT: Systematic Tuning of Long Prompts}.
\newblock \bibinfo{journal}{\emph{ArXiv preprint}}  \bibinfo{volume}{abs/2410.20788} (\bibinfo{year}{2024}).
\newblock
\urldef\tempurl%
\url{https://arxiv.org/abs/2410.20788}
\showURL{%
\tempurl}


\bibitem[Lan et~al\mbox{.}(2020)]%
        {Lan2019ALBERTAL}
\bibfield{author}{\bibinfo{person}{Zhenzhong Lan}, \bibinfo{person}{Mingda Chen}, \bibinfo{person}{Sebastian Goodman}, \bibinfo{person}{Kevin Gimpel}, \bibinfo{person}{Piyush Sharma}, {and} \bibinfo{person}{Radu Soricut}.} \bibinfo{year}{2020}\natexlab{}.
\newblock \showarticletitle{{ALBERT:} {A} Lite {BERT} for Self-supervised Learning of Language Representations}. In \bibinfo{booktitle}{\emph{8th International Conference on Learning Representations, {ICLR} 2020, Addis Ababa, Ethiopia, April 26-30, 2020}}. \bibinfo{publisher}{OpenReview.net}.
\newblock
\urldef\tempurl%
\url{https://openreview.net/forum?id=H1eA7AEtvS}
\showURL{%
\tempurl}


\bibitem[Lester et~al\mbox{.}(2021)]%
        {Lester2021ThePO}
\bibfield{author}{\bibinfo{person}{Brian Lester}, \bibinfo{person}{Rami Al-Rfou}, {and} \bibinfo{person}{Noah Constant}.} \bibinfo{year}{2021}\natexlab{}.
\newblock \showarticletitle{The Power of Scale for Parameter-Efficient Prompt Tuning}. In \bibinfo{booktitle}{\emph{Proceedings of the 2021 Conference on Empirical Methods in Natural Language Processing}}, \bibfield{editor}{\bibinfo{person}{Marie-Francine Moens}, \bibinfo{person}{Xuanjing Huang}, \bibinfo{person}{Lucia Specia}, {and} \bibinfo{person}{Scott Wen-tau Yih}} (Eds.). \bibinfo{publisher}{Association for Computational Linguistics}, \bibinfo{address}{Online and Punta Cana, Dominican Republic}, \bibinfo{pages}{3045--3059}.
\newblock
\urldef\tempurl%
\url{https://doi.org/10.18653/v1/2021.emnlp-main.243}
\showDOI{\tempurl}


\bibitem[Li et~al\mbox{.}(2023d)]%
        {Li2023DeliberateTG}
\bibfield{author}{\bibinfo{person}{Bei Li}, \bibinfo{person}{Rui Wang}, \bibinfo{person}{Junliang Guo}, \bibinfo{person}{Kaitao Song}, \bibinfo{person}{Xuejiao Tan}, \bibinfo{person}{Hany Hassan}, \bibinfo{person}{Arul Menezes}, \bibinfo{person}{Tong Xiao}, \bibinfo{person}{Jiang Bian}, {and} \bibinfo{person}{Jingbo Zhu}.} \bibinfo{year}{2023}\natexlab{d}.
\newblock \showarticletitle{Deliberate then Generate: Enhanced Prompting Framework for Text Generation}.
\newblock \bibinfo{journal}{\emph{ArXiv preprint}}  \bibinfo{volume}{abs/2305.19835} (\bibinfo{year}{2023}).
\newblock
\urldef\tempurl%
\url{https://arxiv.org/abs/2305.19835}
\showURL{%
\tempurl}


\bibitem[Li et~al\mbox{.}(2023e)]%
        {Li2023PromptDF}
\bibfield{author}{\bibinfo{person}{Lei Li}, \bibinfo{person}{Yongfeng Zhang}, {and} \bibinfo{person}{Li Chen}.} \bibinfo{year}{2023}\natexlab{e}.
\newblock \showarticletitle{Prompt Distillation for Efficient LLM-based Recommendation}.
\newblock \bibinfo{journal}{\emph{Proceedings of the 32nd ACM International Conference on Information and Knowledge Management}} (\bibinfo{year}{2023}).
\newblock
\urldef\tempurl%
\url{https://api.semanticscholar.org/CorpusID:264350121}
\showURL{%
\tempurl}


\bibitem[Li et~al\mbox{.}(2023c)]%
        {Li2023RobustPO}
\bibfield{author}{\bibinfo{person}{Moxin Li}, \bibinfo{person}{Wenjie Wang}, \bibinfo{person}{Fuli Feng}, \bibinfo{person}{Yixin Cao}, \bibinfo{person}{Jizhi Zhang}, {and} \bibinfo{person}{Tat-Seng Chua}.} \bibinfo{year}{2023}\natexlab{c}.
\newblock \showarticletitle{Robust Prompt Optimization for Large Language Models Against Distribution Shifts}. In \bibinfo{booktitle}{\emph{Proceedings of the 2023 Conference on Empirical Methods in Natural Language Processing}}, \bibfield{editor}{\bibinfo{person}{Houda Bouamor}, \bibinfo{person}{Juan Pino}, {and} \bibinfo{person}{Kalika Bali}} (Eds.). \bibinfo{publisher}{Association for Computational Linguistics}, \bibinfo{address}{Singapore}, \bibinfo{pages}{1539--1554}.
\newblock
\urldef\tempurl%
\url{https://doi.org/10.18653/v1/2023.emnlp-main.95}
\showDOI{\tempurl}


\bibitem[Li et~al\mbox{.}(2024a)]%
        {Li2024SayMW}
\bibfield{author}{\bibinfo{person}{Xinze Li}, \bibinfo{person}{Zhenghao Liu}, \bibinfo{person}{Chenyan Xiong}, \bibinfo{person}{Shi Yu}, \bibinfo{person}{Yukun Yan}, \bibinfo{person}{Shuo Wang}, {and} \bibinfo{person}{Ge Yu}.} \bibinfo{year}{2024}\natexlab{a}.
\newblock \showarticletitle{Say More with Less: Understanding Prompt Learning Behaviors through Gist Compression}.
\newblock \bibinfo{journal}{\emph{ArXiv preprint}}  \bibinfo{volume}{abs/2402.16058} (\bibinfo{year}{2024}).
\newblock
\urldef\tempurl%
\url{https://arxiv.org/abs/2402.16058}
\showURL{%
\tempurl}


\bibitem[Li and Liang(2021)]%
        {Li2021PrefixTuningOC}
\bibfield{author}{\bibinfo{person}{Xiang~Lisa Li} {and} \bibinfo{person}{Percy Liang}.} \bibinfo{year}{2021}\natexlab{}.
\newblock \showarticletitle{Prefix-Tuning: Optimizing Continuous Prompts for Generation}. In \bibinfo{booktitle}{\emph{Proceedings of the 59th Annual Meeting of the Association for Computational Linguistics and the 11th International Joint Conference on Natural Language Processing (Volume 1: Long Papers)}}, \bibfield{editor}{\bibinfo{person}{Chengqing Zong}, \bibinfo{person}{Fei Xia}, \bibinfo{person}{Wenjie Li}, {and} \bibinfo{person}{Roberto Navigli}} (Eds.). \bibinfo{publisher}{Association for Computational Linguistics}, \bibinfo{address}{Online}, \bibinfo{pages}{4582--4597}.
\newblock
\urldef\tempurl%
\url{https://doi.org/10.18653/v1/2021.acl-long.353}
\showDOI{\tempurl}


\bibitem[Li et~al\mbox{.}(2023a)]%
        {Li2023CompressingCT}
\bibfield{author}{\bibinfo{person}{Yucheng Li}, \bibinfo{person}{Bo Dong}, \bibinfo{person}{Frank Guerin}, {and} \bibinfo{person}{Chenghua Lin}.} \bibinfo{year}{2023}\natexlab{a}.
\newblock \showarticletitle{Compressing Context to Enhance Inference Efficiency of Large Language Models}. In \bibinfo{booktitle}{\emph{Proceedings of the 2023 Conference on Empirical Methods in Natural Language Processing}}, \bibfield{editor}{\bibinfo{person}{Houda Bouamor}, \bibinfo{person}{Juan Pino}, {and} \bibinfo{person}{Kalika Bali}} (Eds.). \bibinfo{publisher}{Association for Computational Linguistics}, \bibinfo{address}{Singapore}, \bibinfo{pages}{6342--6353}.
\newblock
\urldef\tempurl%
\url{https://doi.org/10.18653/v1/2023.emnlp-main.391}
\showDOI{\tempurl}


\bibitem[Li et~al\mbox{.}(2023b)]%
        {li2024guiding}
\bibfield{author}{\bibinfo{person}{Zekun Li}, \bibinfo{person}{Baolin Peng}, \bibinfo{person}{Pengcheng He}, \bibinfo{person}{Michel Galley}, \bibinfo{person}{Jianfeng Gao}, {and} \bibinfo{person}{Xifeng Yan}.} \bibinfo{year}{2023}\natexlab{b}.
\newblock \showarticletitle{Guiding Large Language Models via Directional Stimulus Prompting}. In \bibinfo{booktitle}{\emph{Advances in Neural Information Processing Systems 36: Annual Conference on Neural Information Processing Systems 2023, NeurIPS 2023, New Orleans, LA, USA, December 10 - 16, 2023}}, \bibfield{editor}{\bibinfo{person}{Alice Oh}, \bibinfo{person}{Tristan Naumann}, \bibinfo{person}{Amir Globerson}, \bibinfo{person}{Kate Saenko}, \bibinfo{person}{Moritz Hardt}, {and} \bibinfo{person}{Sergey Levine}} (Eds.).
\newblock
\urldef\tempurl%
\url{http://papers.nips.cc/paper\_files/paper/2023/hash/c5601d99ed028448f29d1dae2e4a926d-Abstract-Conference.html}
\showURL{%
\tempurl}


\bibitem[Li et~al\mbox{.}(2024b)]%
        {Li2024500xCompressorGP}
\bibfield{author}{\bibinfo{person}{Zongqian Li}, \bibinfo{person}{Yixuan Su}, {and} \bibinfo{person}{Nigel Collier}.} \bibinfo{year}{2024}\natexlab{b}.
\newblock \showarticletitle{500xCompressor: Generalized Prompt Compression for Large Language Models}.
\newblock \bibinfo{journal}{\emph{ArXiv preprint}}  \bibinfo{volume}{abs/2408.03094} (\bibinfo{year}{2024}).
\newblock
\urldef\tempurl%
\url{https://arxiv.org/abs/2408.03094}
\showURL{%
\tempurl}


\bibitem[Lin et~al\mbox{.}(2024)]%
        {lin2024prompt}
\bibfield{author}{\bibinfo{person}{Xiaoqiang Lin}, \bibinfo{person}{Zhongxiang Dai}, \bibinfo{person}{Arun Verma}, \bibinfo{person}{See-Kiong Ng}, \bibinfo{person}{Patrick Jaillet}, {and} \bibinfo{person}{Bryan Kian~Hsiang Low}.} \bibinfo{year}{2024}\natexlab{}.
\newblock \showarticletitle{Prompt Optimization with Human Feedback}.
\newblock \bibinfo{journal}{\emph{ArXiv preprint}}  \bibinfo{volume}{abs/2405.17346} (\bibinfo{year}{2024}).
\newblock
\urldef\tempurl%
\url{https://arxiv.org/abs/2405.17346}
\showURL{%
\tempurl}


\bibitem[Liskavets et~al\mbox{.}(2024)]%
        {Liskavets2024PromptCW}
\bibfield{author}{\bibinfo{person}{Barys Liskavets}, \bibinfo{person}{Maxim Ushakov}, \bibinfo{person}{Shuvendu Roy}, \bibinfo{person}{Mark Klibanov}, \bibinfo{person}{Ali Etemad}, {and} \bibinfo{person}{Shane Luke}.} \bibinfo{year}{2024}\natexlab{}.
\newblock \showarticletitle{Prompt Compression with Context-Aware Sentence Encoding for Fast and Improved LLM Inference}.
\newblock
\urldef\tempurl%
\url{https://api.semanticscholar.org/CorpusID:272367247}
\showURL{%
\tempurl}


\bibitem[Liu(2019)]%
        {liu2019roberta}
\bibfield{author}{\bibinfo{person}{Yinhan Liu}.} \bibinfo{year}{2019}\natexlab{}.
\newblock \showarticletitle{Roberta: A robustly optimized bert pretraining approach}.
\newblock \bibinfo{journal}{\emph{ArXiv preprint}}  \bibinfo{volume}{abs/1907.11692} (\bibinfo{year}{2019}).
\newblock
\urldef\tempurl%
\url{https://arxiv.org/abs/1907.11692}
\showURL{%
\tempurl}


\bibitem[Liu et~al\mbox{.}(2019)]%
        {Liu2019RoBERTaAR}
\bibfield{author}{\bibinfo{person}{Yinhan Liu}, \bibinfo{person}{Myle Ott}, \bibinfo{person}{Naman Goyal}, \bibinfo{person}{Jingfei Du}, \bibinfo{person}{Mandar Joshi}, \bibinfo{person}{Danqi Chen}, \bibinfo{person}{Omer Levy}, \bibinfo{person}{Mike Lewis}, \bibinfo{person}{Luke Zettlemoyer}, {and} \bibinfo{person}{Veselin Stoyanov}.} \bibinfo{year}{2019}\natexlab{}.
\newblock \showarticletitle{RoBERTa: A Robustly Optimized BERT Pretraining Approach}.
\newblock \bibinfo{journal}{\emph{ArXiv preprint}}  \bibinfo{volume}{abs/1907.11692} (\bibinfo{year}{2019}).
\newblock
\urldef\tempurl%
\url{https://arxiv.org/abs/1907.11692}
\showURL{%
\tempurl}


\bibitem[Lu et~al\mbox{.}(2024)]%
        {lu2024fipo}
\bibfield{author}{\bibinfo{person}{Junru Lu}, \bibinfo{person}{Siyu An}, \bibinfo{person}{Min Zhang}, \bibinfo{person}{Yulan He}, \bibinfo{person}{Di Yin}, {and} \bibinfo{person}{Xing Sun}.} \bibinfo{year}{2024}\natexlab{}.
\newblock \showarticletitle{FIPO: Free-form Instruction-oriented Prompt Optimization with Preference Dataset and Modular Fine-tuning Schema}.
\newblock \bibinfo{journal}{\emph{ArXiv preprint}}  \bibinfo{volume}{abs/2402.11811} (\bibinfo{year}{2024}).
\newblock
\urldef\tempurl%
\url{https://arxiv.org/abs/2402.11811}
\showURL{%
\tempurl}


\bibitem[Lu et~al\mbox{.}(2023)]%
        {lu2022dynamic}
\bibfield{author}{\bibinfo{person}{Pan Lu}, \bibinfo{person}{Liang Qiu}, \bibinfo{person}{Kai{-}Wei Chang}, \bibinfo{person}{Ying~Nian Wu}, \bibinfo{person}{Song{-}Chun Zhu}, \bibinfo{person}{Tanmay Rajpurohit}, \bibinfo{person}{Peter Clark}, {and} \bibinfo{person}{Ashwin Kalyan}.} \bibinfo{year}{2023}\natexlab{}.
\newblock \showarticletitle{Dynamic Prompt Learning via Policy Gradient for Semi-structured Mathematical Reasoning}. In \bibinfo{booktitle}{\emph{The Eleventh International Conference on Learning Representations, {ICLR} 2023, Kigali, Rwanda, May 1-5, 2023}}. \bibinfo{publisher}{OpenReview.net}.
\newblock
\urldef\tempurl%
\url{https://openreview.net/pdf?id=DHyHRBwJUTN}
\showURL{%
\tempurl}


\bibitem[Ma et~al\mbox{.}(2024)]%
        {ma2024large}
\bibfield{author}{\bibinfo{person}{Ruotian Ma}, \bibinfo{person}{Xiaolei Wang}, \bibinfo{person}{Xin Zhou}, \bibinfo{person}{Jian Li}, \bibinfo{person}{Nan Du}, \bibinfo{person}{Tao Gui}, \bibinfo{person}{Qi Zhang}, {and} \bibinfo{person}{Xuanjing Huang}.} \bibinfo{year}{2024}\natexlab{}.
\newblock \showarticletitle{Are Large Language Models Good Prompt Optimizers?}
\newblock \bibinfo{journal}{\emph{ArXiv preprint}}  \bibinfo{volume}{abs/2402.02101} (\bibinfo{year}{2024}).
\newblock
\urldef\tempurl%
\url{https://arxiv.org/abs/2402.02101}
\showURL{%
\tempurl}


\bibitem[Madaan et~al\mbox{.}(2023)]%
        {madaan2024self}
\bibfield{author}{\bibinfo{person}{Aman Madaan}, \bibinfo{person}{Niket Tandon}, \bibinfo{person}{Prakhar Gupta}, \bibinfo{person}{Skyler Hallinan}, \bibinfo{person}{Luyu Gao}, \bibinfo{person}{Sarah Wiegreffe}, \bibinfo{person}{Uri Alon}, \bibinfo{person}{Nouha Dziri}, \bibinfo{person}{Shrimai Prabhumoye}, \bibinfo{person}{Yiming Yang}, \bibinfo{person}{Shashank Gupta}, \bibinfo{person}{Bodhisattwa~Prasad Majumder}, \bibinfo{person}{Katherine Hermann}, \bibinfo{person}{Sean Welleck}, \bibinfo{person}{Amir Yazdanbakhsh}, {and} \bibinfo{person}{Peter Clark}.} \bibinfo{year}{2023}\natexlab{}.
\newblock \showarticletitle{Self-Refine: Iterative Refinement with Self-Feedback}. In \bibinfo{booktitle}{\emph{Advances in Neural Information Processing Systems 36: Annual Conference on Neural Information Processing Systems 2023, NeurIPS 2023, New Orleans, LA, USA, December 10 - 16, 2023}}, \bibfield{editor}{\bibinfo{person}{Alice Oh}, \bibinfo{person}{Tristan Naumann}, \bibinfo{person}{Amir Globerson}, \bibinfo{person}{Kate Saenko}, \bibinfo{person}{Moritz Hardt}, {and} \bibinfo{person}{Sergey Levine}} (Eds.).
\newblock
\urldef\tempurl%
\url{http://papers.nips.cc/paper\_files/paper/2023/hash/91edff07232fb1b55a505a9e9f6c0ff3-Abstract-Conference.html}
\showURL{%
\tempurl}


\bibitem[Mu et~al\mbox{.}(2023)]%
        {Mu2023LearningTC}
\bibfield{author}{\bibinfo{person}{Jesse Mu}, \bibinfo{person}{Xiang Li}, {and} \bibinfo{person}{Noah~D. Goodman}.} \bibinfo{year}{2023}\natexlab{}.
\newblock \showarticletitle{Learning to Compress Prompts with Gist Tokens}. In \bibinfo{booktitle}{\emph{Advances in Neural Information Processing Systems 36: Annual Conference on Neural Information Processing Systems 2023, NeurIPS 2023, New Orleans, LA, USA, December 10 - 16, 2023}}, \bibfield{editor}{\bibinfo{person}{Alice Oh}, \bibinfo{person}{Tristan Naumann}, \bibinfo{person}{Amir Globerson}, \bibinfo{person}{Kate Saenko}, \bibinfo{person}{Moritz Hardt}, {and} \bibinfo{person}{Sergey Levine}} (Eds.).
\newblock
\urldef\tempurl%
\url{http://papers.nips.cc/paper\_files/paper/2023/hash/3d77c6dcc7f143aa2154e7f4d5e22d68-Abstract-Conference.html}
\showURL{%
\tempurl}


\bibitem[{OpenAI}(2024a)]%
        {o1}
\bibfield{author}{\bibinfo{person}{{OpenAI}}.} \bibinfo{year}{2024}\natexlab{a}.
\newblock \bibinfo{title}{{Introducing OpenAI o1-preview}}.
\newblock
\urldef\tempurl%
\url{https://openai.com/index/introducing-openai-o1-preview/}
\showURL{%
\tempurl}


\bibitem[{OpenAI}(2024b)]%
        {o1-mini}
\bibfield{author}{\bibinfo{person}{{OpenAI}}.} \bibinfo{year}{2024}\natexlab{b}.
\newblock \bibinfo{title}{{OpenAI o1-mini}}.
\newblock
\urldef\tempurl%
\url{https://openai.com/index/openai-o1-mini-advancing-cost-efficient-reasoning/}
\showURL{%
\tempurl}


\bibitem[Ouyang et~al\mbox{.}(2022)]%
        {ouyang2022training}
\bibfield{author}{\bibinfo{person}{Long Ouyang}, \bibinfo{person}{Jeffrey Wu}, \bibinfo{person}{Xu Jiang}, \bibinfo{person}{Diogo Almeida}, \bibinfo{person}{Carroll~L. Wainwright}, \bibinfo{person}{Pamela Mishkin}, \bibinfo{person}{Chong Zhang}, \bibinfo{person}{Sandhini Agarwal}, \bibinfo{person}{Katarina Slama}, \bibinfo{person}{Alex Ray}, \bibinfo{person}{John Schulman}, \bibinfo{person}{Jacob Hilton}, \bibinfo{person}{Fraser Kelton}, \bibinfo{person}{Luke Miller}, \bibinfo{person}{Maddie Simens}, \bibinfo{person}{Amanda Askell}, \bibinfo{person}{Peter Welinder}, \bibinfo{person}{Paul~F. Christiano}, \bibinfo{person}{Jan Leike}, {and} \bibinfo{person}{Ryan Lowe}.} \bibinfo{year}{2022}\natexlab{}.
\newblock \showarticletitle{Training language models to follow instructions with human feedback}. In \bibinfo{booktitle}{\emph{Advances in Neural Information Processing Systems 35: Annual Conference on Neural Information Processing Systems 2022, NeurIPS 2022, New Orleans, LA, USA, November 28 - December 9, 2022}}, \bibfield{editor}{\bibinfo{person}{Sanmi Koyejo}, \bibinfo{person}{S.~Mohamed}, \bibinfo{person}{A.~Agarwal}, \bibinfo{person}{Danielle Belgrave}, \bibinfo{person}{K.~Cho}, {and} \bibinfo{person}{A.~Oh}} (Eds.).
\newblock
\urldef\tempurl%
\url{http://papers.nips.cc/paper\_files/paper/2022/hash/b1efde53be364a73914f58805a001731-Abstract-Conference.html}
\showURL{%
\tempurl}


\bibitem[Pan et~al\mbox{.}(2023)]%
        {Pan2023PlumPL}
\bibfield{author}{\bibinfo{person}{Rui Pan}, \bibinfo{person}{Shuo Xing}, \bibinfo{person}{Shizhe Diao}, \bibinfo{person}{Xiang Liu}, \bibinfo{person}{Kashun Shum}, \bibinfo{person}{Jipeng Zhang}, {and} \bibinfo{person}{Tong Zhang}.} \bibinfo{year}{2023}\natexlab{}.
\newblock \showarticletitle{Plum: Prompt Learning using Metaheuristic}.
\newblock \bibinfo{journal}{\emph{ArXiv preprint}}  \bibinfo{volume}{abs/2311.08364} (\bibinfo{year}{2023}).
\newblock
\urldef\tempurl%
\url{https://arxiv.org/abs/2311.08364}
\showURL{%
\tempurl}


\bibitem[Pan et~al\mbox{.}(2024)]%
        {Pan2024LLMLingua2DD}
\bibfield{author}{\bibinfo{person}{Zhuoshi Pan}, \bibinfo{person}{Qianhui Wu}, \bibinfo{person}{Huiqiang Jiang}, \bibinfo{person}{Menglin Xia}, \bibinfo{person}{Xufang Luo}, \bibinfo{person}{Jue Zhang}, \bibinfo{person}{Qingwei Lin}, \bibinfo{person}{Victor R{\"u}hle}, \bibinfo{person}{Yuqing Yang}, \bibinfo{person}{Chin-Yew Lin}, \bibinfo{person}{H.~Vicky Zhao}, \bibinfo{person}{Lili Qiu}, \bibinfo{person}{Dongmei Zhang}, \bibinfo{person}{Karl Cobbe}, \bibinfo{person}{Vineet Kosaraju}, \bibinfo{person}{Mo Bavarian}, \bibinfo{person}{Mark Chen}, \bibinfo{person}{Heewoo Jun}, \bibinfo{person}{Lukasz Kaiser}, \bibinfo{person}{Matthias Plappert}, \bibinfo{person}{Jerry Tworek}, \bibinfo{person}{Jacob Hilton}, {and} \bibinfo{person}{Reiichiro Nakano}.} \bibinfo{year}{2024}\natexlab{}.
\newblock \showarticletitle{LLMLingua-2: Data Distillation for Efficient and Faithful Task-Agnostic Prompt Compression}. In \bibinfo{booktitle}{\emph{Annual Meeting of the Association for Computational Linguistics}}.
\newblock
\urldef\tempurl%
\url{https://api.semanticscholar.org/CorpusID:268531237}
\showURL{%
\tempurl}


\bibitem[Paranjape et~al\mbox{.}(2023)]%
        {Paranjape2023ARTAM}
\bibfield{author}{\bibinfo{person}{Bhargavi Paranjape}, \bibinfo{person}{Scott~M. Lundberg}, \bibinfo{person}{Sameer Singh}, \bibinfo{person}{Hannaneh Hajishirzi}, \bibinfo{person}{Luke Zettlemoyer}, {and} \bibinfo{person}{Marco~Tulio Ribeiro}.} \bibinfo{year}{2023}\natexlab{}.
\newblock \showarticletitle{ART: Automatic multi-step reasoning and tool-use for large language models}.
\newblock \bibinfo{journal}{\emph{ArXiv preprint}}  \bibinfo{volume}{abs/2303.09014} (\bibinfo{year}{2023}).
\newblock
\urldef\tempurl%
\url{https://arxiv.org/abs/2303.09014}
\showURL{%
\tempurl}


\bibitem[Phang et~al\mbox{.}(2023)]%
        {Phang2022HyperTuningTA}
\bibfield{author}{\bibinfo{person}{Jason Phang}, \bibinfo{person}{Yi Mao}, \bibinfo{person}{Pengcheng He}, {and} \bibinfo{person}{Weizhu Chen}.} \bibinfo{year}{2023}\natexlab{}.
\newblock \showarticletitle{HyperTuning: Toward Adapting Large Language Models without Back-propagation}. In \bibinfo{booktitle}{\emph{International Conference on Machine Learning, {ICML} 2023, 23-29 July 2023, Honolulu, Hawaii, {USA}}} \emph{(\bibinfo{series}{Proceedings of Machine Learning Research}, Vol.~\bibinfo{volume}{202})}, \bibfield{editor}{\bibinfo{person}{Andreas Krause}, \bibinfo{person}{Emma Brunskill}, \bibinfo{person}{Kyunghyun Cho}, \bibinfo{person}{Barbara Engelhardt}, \bibinfo{person}{Sivan Sabato}, {and} \bibinfo{person}{Jonathan Scarlett}} (Eds.). \bibinfo{publisher}{{PMLR}}, \bibinfo{pages}{27854--27875}.
\newblock
\urldef\tempurl%
\url{https://proceedings.mlr.press/v202/phang23a.html}
\showURL{%
\tempurl}


\bibitem[Pitis et~al\mbox{.}(2023)]%
        {pitis2023boosted}
\bibfield{author}{\bibinfo{person}{Silviu Pitis}, \bibinfo{person}{Michael~R Zhang}, \bibinfo{person}{Andrew Wang}, {and} \bibinfo{person}{Jimmy Ba}.} \bibinfo{year}{2023}\natexlab{}.
\newblock \showarticletitle{Boosted prompt ensembles for large language models}.
\newblock \bibinfo{journal}{\emph{ArXiv preprint}}  \bibinfo{volume}{abs/2304.05970} (\bibinfo{year}{2023}).
\newblock
\urldef\tempurl%
\url{https://arxiv.org/abs/2304.05970}
\showURL{%
\tempurl}


\bibitem[Prasad et~al\mbox{.}(2023)]%
        {Prasad2022GrIPSGE}
\bibfield{author}{\bibinfo{person}{Archiki Prasad}, \bibinfo{person}{Peter Hase}, \bibinfo{person}{Xiang Zhou}, {and} \bibinfo{person}{Mohit Bansal}.} \bibinfo{year}{2023}\natexlab{}.
\newblock \showarticletitle{{G}r{IPS}: Gradient-free, Edit-based Instruction Search for Prompting Large Language Models}. In \bibinfo{booktitle}{\emph{Proceedings of the 17th Conference of the European Chapter of the Association for Computational Linguistics}}, \bibfield{editor}{\bibinfo{person}{Andreas Vlachos} {and} \bibinfo{person}{Isabelle Augenstein}} (Eds.). \bibinfo{publisher}{Association for Computational Linguistics}, \bibinfo{address}{Dubrovnik, Croatia}, \bibinfo{pages}{3845--3864}.
\newblock
\urldef\tempurl%
\url{https://doi.org/10.18653/v1/2023.eacl-main.277}
\showDOI{\tempurl}


\bibitem[Press et~al\mbox{.}(2023)]%
        {press2022measuring}
\bibfield{author}{\bibinfo{person}{Ofir Press}, \bibinfo{person}{Muru Zhang}, \bibinfo{person}{Sewon Min}, \bibinfo{person}{Ludwig Schmidt}, \bibinfo{person}{Noah Smith}, {and} \bibinfo{person}{Mike Lewis}.} \bibinfo{year}{2023}\natexlab{}.
\newblock \showarticletitle{Measuring and Narrowing the Compositionality Gap in Language Models}. In \bibinfo{booktitle}{\emph{Findings of the Association for Computational Linguistics: EMNLP 2023}}, \bibfield{editor}{\bibinfo{person}{Houda Bouamor}, \bibinfo{person}{Juan Pino}, {and} \bibinfo{person}{Kalika Bali}} (Eds.). \bibinfo{publisher}{Association for Computational Linguistics}, \bibinfo{address}{Singapore}, \bibinfo{pages}{5687--5711}.
\newblock
\urldef\tempurl%
\url{https://doi.org/10.18653/v1/2023.findings-emnlp.378}
\showDOI{\tempurl}


\bibitem[Pryzant et~al\mbox{.}(2023)]%
        {Pryzant2023AutomaticPO}
\bibfield{author}{\bibinfo{person}{Reid Pryzant}, \bibinfo{person}{Dan Iter}, \bibinfo{person}{Jerry Li}, \bibinfo{person}{Yin Lee}, \bibinfo{person}{Chenguang Zhu}, {and} \bibinfo{person}{Michael Zeng}.} \bibinfo{year}{2023}\natexlab{}.
\newblock \showarticletitle{Automatic Prompt Optimization with {``}Gradient Descent{''} and Beam Search}. In \bibinfo{booktitle}{\emph{Proceedings of the 2023 Conference on Empirical Methods in Natural Language Processing}}, \bibfield{editor}{\bibinfo{person}{Houda Bouamor}, \bibinfo{person}{Juan Pino}, {and} \bibinfo{person}{Kalika Bali}} (Eds.). \bibinfo{publisher}{Association for Computational Linguistics}, \bibinfo{address}{Singapore}, \bibinfo{pages}{7957--7968}.
\newblock
\urldef\tempurl%
\url{https://doi.org/10.18653/v1/2023.emnlp-main.494}
\showDOI{\tempurl}


\bibitem[Pu et~al\mbox{.}(2024)]%
        {pu2024style}
\bibfield{author}{\bibinfo{person}{Xiao Pu}, \bibinfo{person}{Tianxing He}, {and} \bibinfo{person}{Xiaojun Wan}.} \bibinfo{year}{2024}\natexlab{}.
\newblock \showarticletitle{Style-Compress: An LLM-Based Prompt Compression Framework Considering Task-Specific Styles}. In \bibinfo{booktitle}{\emph{Findings of the Association for Computational Linguistics: EMNLP 2024}}. \bibinfo{pages}{14533--14549}.
\newblock


\bibitem[Qin et~al\mbox{.}(2023)]%
        {qin2023toolllm}
\bibfield{author}{\bibinfo{person}{Yujia Qin}, \bibinfo{person}{Shihao Liang}, \bibinfo{person}{Yining Ye}, \bibinfo{person}{Kunlun Zhu}, \bibinfo{person}{Lan Yan}, \bibinfo{person}{Yaxi Lu}, \bibinfo{person}{Yankai Lin}, \bibinfo{person}{Xin Cong}, \bibinfo{person}{Xiangru Tang}, \bibinfo{person}{Bill Qian}, {et~al\mbox{.}}} \bibinfo{year}{2023}\natexlab{}.
\newblock \showarticletitle{Toolllm: Facilitating large language models to master 16000+ real-world apis}.
\newblock \bibinfo{journal}{\emph{ArXiv preprint}}  \bibinfo{volume}{abs/2307.16789} (\bibinfo{year}{2023}).
\newblock
\urldef\tempurl%
\url{https://arxiv.org/abs/2307.16789}
\showURL{%
\tempurl}


\bibitem[Radford et~al\mbox{.}(2021)]%
        {Linearprobe2021LearningTV}
\bibfield{author}{\bibinfo{person}{Alec Radford}, \bibinfo{person}{Jong~Wook Kim}, \bibinfo{person}{Chris Hallacy}, \bibinfo{person}{Aditya Ramesh}, \bibinfo{person}{Gabriel Goh}, \bibinfo{person}{Sandhini Agarwal}, \bibinfo{person}{Girish Sastry}, \bibinfo{person}{Amanda Askell}, \bibinfo{person}{Pamela Mishkin}, \bibinfo{person}{Jack Clark}, \bibinfo{person}{Gretchen Krueger}, {and} \bibinfo{person}{Ilya Sutskever}.} \bibinfo{year}{2021}\natexlab{}.
\newblock \showarticletitle{Learning Transferable Visual Models From Natural Language Supervision}. In \bibinfo{booktitle}{\emph{Proceedings of the 38th International Conference on Machine Learning, {ICML} 2021, 18-24 July 2021, Virtual Event}} \emph{(\bibinfo{series}{Proceedings of Machine Learning Research}, Vol.~\bibinfo{volume}{139})}, \bibfield{editor}{\bibinfo{person}{Marina Meila} {and} \bibinfo{person}{Tong Zhang}} (Eds.). \bibinfo{publisher}{{PMLR}}, \bibinfo{pages}{8748--8763}.
\newblock
\urldef\tempurl%
\url{http://proceedings.mlr.press/v139/radford21a.html}
\showURL{%
\tempurl}


\bibitem[Radford and Narasimhan(2018)]%
        {Radford2018ImprovingLU}
\bibfield{author}{\bibinfo{person}{Alec Radford} {and} \bibinfo{person}{Karthik Narasimhan}.} \bibinfo{year}{2018}\natexlab{}.
\newblock \showarticletitle{Improving Language Understanding by Generative Pre-Training}.
\newblock
\urldef\tempurl%
\url{https://api.semanticscholar.org/CorpusID:49313245}
\showURL{%
\tempurl}


\bibitem[Radford et~al\mbox{.}(2019)]%
        {Radford2019LanguageMA}
\bibfield{author}{\bibinfo{person}{Alec Radford}, \bibinfo{person}{Jeff Wu}, \bibinfo{person}{Rewon Child}, \bibinfo{person}{David Luan}, \bibinfo{person}{Dario Amodei}, {and} \bibinfo{person}{Ilya Sutskever}.} \bibinfo{year}{2019}\natexlab{}.
\newblock \showarticletitle{Language Models are Unsupervised Multitask Learners}.
\newblock
\urldef\tempurl%
\url{https://api.semanticscholar.org/CorpusID:160025533}
\showURL{%
\tempurl}


\bibitem[Rafailov et~al\mbox{.}(2023)]%
        {Rafailov2023DirectPO}
\bibfield{author}{\bibinfo{person}{Rafael Rafailov}, \bibinfo{person}{Archit Sharma}, \bibinfo{person}{Eric Mitchell}, \bibinfo{person}{Christopher~D. Manning}, \bibinfo{person}{Stefano Ermon}, {and} \bibinfo{person}{Chelsea Finn}.} \bibinfo{year}{2023}\natexlab{}.
\newblock \showarticletitle{Direct Preference Optimization: Your Language Model is Secretly a Reward Model}. In \bibinfo{booktitle}{\emph{Advances in Neural Information Processing Systems 36: Annual Conference on Neural Information Processing Systems 2023, NeurIPS 2023, New Orleans, LA, USA, December 10 - 16, 2023}}, \bibfield{editor}{\bibinfo{person}{Alice Oh}, \bibinfo{person}{Tristan Naumann}, \bibinfo{person}{Amir Globerson}, \bibinfo{person}{Kate Saenko}, \bibinfo{person}{Moritz Hardt}, {and} \bibinfo{person}{Sergey Levine}} (Eds.).
\newblock
\urldef\tempurl%
\url{http://papers.nips.cc/paper\_files/paper/2023/hash/a85b405ed65c6477a4fe8302b5e06ce7-Abstract-Conference.html}
\showURL{%
\tempurl}


\bibitem[Raffel et~al\mbox{.}(2020)]%
        {raffel2020exploring}
\bibfield{author}{\bibinfo{person}{Colin Raffel}, \bibinfo{person}{Noam Shazeer}, \bibinfo{person}{Adam Roberts}, \bibinfo{person}{Katherine Lee}, \bibinfo{person}{Sharan Narang}, \bibinfo{person}{Michael Matena}, \bibinfo{person}{Yanqi Zhou}, \bibinfo{person}{Wei Li}, {and} \bibinfo{person}{Peter~J. Liu}.} \bibinfo{year}{2020}\natexlab{}.
\newblock \showarticletitle{Exploring the Limits of Transfer Learning with a Unified Text-to-Text Transformer}.
\newblock \bibinfo{journal}{\emph{J. Mach. Learn. Res.}}  \bibinfo{volume}{21} (\bibinfo{year}{2020}), \bibinfo{pages}{140:1--140:67}.
\newblock
\urldef\tempurl%
\url{http://jmlr.org/papers/v21/20-074.html}
\showURL{%
\tempurl}


\bibitem[Reimers and Gurevych(2019)]%
        {Reimers2019SentenceBERTSE}
\bibfield{author}{\bibinfo{person}{Nils Reimers} {and} \bibinfo{person}{Iryna Gurevych}.} \bibinfo{year}{2019}\natexlab{}.
\newblock \showarticletitle{Sentence-{BERT}: Sentence Embeddings using {S}iamese {BERT}-Networks}. In \bibinfo{booktitle}{\emph{Proceedings of the 2019 Conference on Empirical Methods in Natural Language Processing and the 9th International Joint Conference on Natural Language Processing (EMNLP-IJCNLP)}}, \bibfield{editor}{\bibinfo{person}{Kentaro Inui}, \bibinfo{person}{Jing Jiang}, \bibinfo{person}{Vincent Ng}, {and} \bibinfo{person}{Xiaojun Wan}} (Eds.). \bibinfo{publisher}{Association for Computational Linguistics}, \bibinfo{address}{Hong Kong, China}, \bibinfo{pages}{3982--3992}.
\newblock
\urldef\tempurl%
\url{https://doi.org/10.18653/v1/D19-1410}
\showDOI{\tempurl}


\bibitem[Schulman et~al\mbox{.}(2017)]%
        {Schulman2017ProximalPO}
\bibfield{author}{\bibinfo{person}{John Schulman}, \bibinfo{person}{Filip Wolski}, \bibinfo{person}{Prafulla Dhariwal}, \bibinfo{person}{Alec Radford}, {and} \bibinfo{person}{Oleg Klimov}.} \bibinfo{year}{2017}\natexlab{}.
\newblock \showarticletitle{Proximal Policy Optimization Algorithms}.
\newblock \bibinfo{journal}{\emph{ArXiv preprint}}  \bibinfo{volume}{abs/1707.06347} (\bibinfo{year}{2017}).
\newblock
\urldef\tempurl%
\url{https://arxiv.org/abs/1707.06347}
\showURL{%
\tempurl}


\bibitem[Sclar et~al\mbox{.}(2023)]%
        {sclar2023quantifying}
\bibfield{author}{\bibinfo{person}{Melanie Sclar}, \bibinfo{person}{Yejin Choi}, \bibinfo{person}{Yulia Tsvetkov}, {and} \bibinfo{person}{Alane Suhr}.} \bibinfo{year}{2023}\natexlab{}.
\newblock \showarticletitle{Quantifying Language Models' Sensitivity to Spurious Features in Prompt Design or: How I learned to start worrying about prompt formatting}.
\newblock \bibinfo{journal}{\emph{ArXiv preprint}}  \bibinfo{volume}{abs/2310.11324} (\bibinfo{year}{2023}).
\newblock
\urldef\tempurl%
\url{https://arxiv.org/abs/2310.11324}
\showURL{%
\tempurl}


\bibitem[Shannon(1948)]%
        {Shannon1948AMT}
\bibfield{author}{\bibinfo{person}{Claude~E. Shannon}.} \bibinfo{year}{1948}\natexlab{}.
\newblock \showarticletitle{A mathematical theory of communication}.
\newblock \bibinfo{journal}{\emph{Bell Syst. Tech. J.}}  \bibinfo{volume}{27} (\bibinfo{year}{1948}), \bibinfo{pages}{623--656}.
\newblock
\urldef\tempurl%
\url{https://api.semanticscholar.org/CorpusID:55379485}
\showURL{%
\tempurl}


\bibitem[Shi et~al\mbox{.}(2024)]%
        {shi2024chain}
\bibfield{author}{\bibinfo{person}{Zhengliang Shi}, \bibinfo{person}{Shen Gao}, \bibinfo{person}{Xiuyi Chen}, \bibinfo{person}{Yue Feng}, \bibinfo{person}{Lingyong Yan}, \bibinfo{person}{Haibo Shi}, \bibinfo{person}{Dawei Yin}, \bibinfo{person}{Zhumin Chen}, \bibinfo{person}{Suzan Verberne}, {and} \bibinfo{person}{Zhaochun Ren}.} \bibinfo{year}{2024}\natexlab{}.
\newblock \showarticletitle{Chain of Tools: Large Language Model is an Automatic Multi-tool Learner}.
\newblock \bibinfo{journal}{\emph{ArXiv preprint}}  \bibinfo{volume}{abs/2405.16533} (\bibinfo{year}{2024}).
\newblock
\urldef\tempurl%
\url{https://arxiv.org/abs/2405.16533}
\showURL{%
\tempurl}


\bibitem[Shinn et~al\mbox{.}(2023)]%
        {Shinn2023ReflexionLA}
\bibfield{author}{\bibinfo{person}{Noah Shinn}, \bibinfo{person}{Federico Cassano}, \bibinfo{person}{Ashwin Gopinath}, \bibinfo{person}{Karthik Narasimhan}, {and} \bibinfo{person}{Shunyu Yao}.} \bibinfo{year}{2023}\natexlab{}.
\newblock \showarticletitle{Reflexion: language agents with verbal reinforcement learning}. In \bibinfo{booktitle}{\emph{Advances in Neural Information Processing Systems 36: Annual Conference on Neural Information Processing Systems 2023, NeurIPS 2023, New Orleans, LA, USA, December 10 - 16, 2023}}, \bibfield{editor}{\bibinfo{person}{Alice Oh}, \bibinfo{person}{Tristan Naumann}, \bibinfo{person}{Amir Globerson}, \bibinfo{person}{Kate Saenko}, \bibinfo{person}{Moritz Hardt}, {and} \bibinfo{person}{Sergey Levine}} (Eds.).
\newblock
\urldef\tempurl%
\url{http://papers.nips.cc/paper\_files/paper/2023/hash/1b44b878bb782e6954cd888628510e90-Abstract-Conference.html}
\showURL{%
\tempurl}


\bibitem[Snell et~al\mbox{.}(2022)]%
        {Snell2022LearningBD}
\bibfield{author}{\bibinfo{person}{Charles~Burton Snell}, \bibinfo{person}{Dan Klein}, {and} \bibinfo{person}{Ruiqi Zhong}.} \bibinfo{year}{2022}\natexlab{}.
\newblock \showarticletitle{Learning by Distilling Context}.
\newblock \bibinfo{journal}{\emph{ArXiv preprint}}  \bibinfo{volume}{abs/2209.15189} (\bibinfo{year}{2022}).
\newblock
\urldef\tempurl%
\url{https://arxiv.org/abs/2209.15189}
\showURL{%
\tempurl}


\bibitem[Sun et~al\mbox{.}(2023b)]%
        {sun2023query}
\bibfield{author}{\bibinfo{person}{Hao Sun}, \bibinfo{person}{Alihan H{\"u}y{\"u}k}, {and} \bibinfo{person}{Mihaela van~der Schaar}.} \bibinfo{year}{2023}\natexlab{b}.
\newblock \showarticletitle{Query-dependent prompt evaluation and optimization with offline inverse RL}. In \bibinfo{booktitle}{\emph{The Twelfth International Conference on Learning Representations}}.
\newblock


\bibitem[Sun et~al\mbox{.}(2023c)]%
        {sun2023autohint}
\bibfield{author}{\bibinfo{person}{Hong Sun}, \bibinfo{person}{Xue Li}, \bibinfo{person}{Yinchuan Xu}, \bibinfo{person}{Youkow Homma}, \bibinfo{person}{Qi Cao}, \bibinfo{person}{Min Wu}, \bibinfo{person}{Jian Jiao}, {and} \bibinfo{person}{Denis Charles}.} \bibinfo{year}{2023}\natexlab{c}.
\newblock \showarticletitle{Autohint: Automatic prompt optimization with hint generation}.
\newblock \bibinfo{journal}{\emph{ArXiv preprint}}  \bibinfo{volume}{abs/2307.07415} (\bibinfo{year}{2023}).
\newblock
\urldef\tempurl%
\url{https://arxiv.org/abs/2307.07415}
\showURL{%
\tempurl}


\bibitem[Sun et~al\mbox{.}(2023a)]%
        {Sun2023InstructionDM}
\bibfield{author}{\bibinfo{person}{Weiwei Sun}, \bibinfo{person}{Zheng Chen}, \bibinfo{person}{Xinyu Ma}, \bibinfo{person}{Lingyong Yan}, \bibinfo{person}{Shuaiqiang Wang}, \bibinfo{person}{Pengjie Ren}, \bibinfo{person}{Zhumin Chen}, \bibinfo{person}{Dawei Yin}, {and} \bibinfo{person}{Zhaochun Ren}.} \bibinfo{year}{2023}\natexlab{a}.
\newblock \showarticletitle{Instruction Distillation Makes Large Language Models Efficient Zero-shot Rankers}.
\newblock \bibinfo{journal}{\emph{ArXiv preprint}}  \bibinfo{volume}{abs/2311.01555} (\bibinfo{year}{2023}).
\newblock
\urldef\tempurl%
\url{https://arxiv.org/abs/2311.01555}
\showURL{%
\tempurl}


\bibitem[Suzgun et~al\mbox{.}(2023)]%
        {Suzgun2022ChallengingBT}
\bibfield{author}{\bibinfo{person}{Mirac Suzgun}, \bibinfo{person}{Nathan Scales}, \bibinfo{person}{Nathanael Sch{\"a}rli}, \bibinfo{person}{Sebastian Gehrmann}, \bibinfo{person}{Yi Tay}, \bibinfo{person}{Hyung~Won Chung}, \bibinfo{person}{Aakanksha Chowdhery}, \bibinfo{person}{Quoc Le}, \bibinfo{person}{Ed Chi}, \bibinfo{person}{Denny Zhou}, {and} \bibinfo{person}{Jason Wei}.} \bibinfo{year}{2023}\natexlab{}.
\newblock \showarticletitle{Challenging {BIG}-Bench Tasks and Whether Chain-of-Thought Can Solve Them}. In \bibinfo{booktitle}{\emph{Findings of the Association for Computational Linguistics: ACL 2023}}, \bibfield{editor}{\bibinfo{person}{Anna Rogers}, \bibinfo{person}{Jordan Boyd-Graber}, {and} \bibinfo{person}{Naoaki Okazaki}} (Eds.). \bibinfo{publisher}{Association for Computational Linguistics}, \bibinfo{address}{Toronto, Canada}, \bibinfo{pages}{13003--13051}.
\newblock
\urldef\tempurl%
\url{https://doi.org/10.18653/v1/2023.findings-acl.824}
\showDOI{\tempurl}


\bibitem[Tan et~al\mbox{.}(2024)]%
        {Tan2024LLoCOLL}
\bibfield{author}{\bibinfo{person}{Sijun Tan}, \bibinfo{person}{Xiuyu Li}, \bibinfo{person}{Shishir~G. Patil}, \bibinfo{person}{Ziyang Wu}, \bibinfo{person}{Tianjun Zhang}, \bibinfo{person}{Kurt Keutzer}, \bibinfo{person}{Joseph~E. Gonzalez}, {and} \bibinfo{person}{Raluca~A. Popa}.} \bibinfo{year}{2024}\natexlab{}.
\newblock \showarticletitle{LLoCO: Learning Long Contexts Offline}.
\newblock \bibinfo{journal}{\emph{ArXiv preprint}}  \bibinfo{volume}{abs/2404.07979} (\bibinfo{year}{2024}).
\newblock
\urldef\tempurl%
\url{https://arxiv.org/abs/2404.07979}
\showURL{%
\tempurl}


\bibitem[Tang et~al\mbox{.}(2024)]%
        {tang2024unleashing}
\bibfield{author}{\bibinfo{person}{Xinyu Tang}, \bibinfo{person}{Xiaolei Wang}, \bibinfo{person}{Wayne~Xin Zhao}, \bibinfo{person}{Siyuan Lu}, \bibinfo{person}{Yaliang Li}, {and} \bibinfo{person}{Ji-Rong Wen}.} \bibinfo{year}{2024}\natexlab{}.
\newblock \showarticletitle{Unleashing the Potential of Large Language Models as Prompt Optimizers: An Analogical Analysis with Gradient-based Model Optimizers}.
\newblock \bibinfo{journal}{\emph{ArXiv preprint}}  \bibinfo{volume}{abs/2402.17564} (\bibinfo{year}{2024}).
\newblock
\urldef\tempurl%
\url{https://arxiv.org/abs/2402.17564}
\showURL{%
\tempurl}


\bibitem[Touvron et~al\mbox{.}(2023)]%
        {touvron2023llama}
\bibfield{author}{\bibinfo{person}{Hugo Touvron}, \bibinfo{person}{Louis Martin}, \bibinfo{person}{Kevin Stone}, \bibinfo{person}{Peter Albert}, \bibinfo{person}{Amjad Almahairi}, \bibinfo{person}{Yasmine Babaei}, \bibinfo{person}{Nikolay Bashlykov}, \bibinfo{person}{Soumya Batra}, \bibinfo{person}{Prajjwal Bhargava}, \bibinfo{person}{Shruti Bhosale}, {et~al\mbox{.}}} \bibinfo{year}{2023}\natexlab{}.
\newblock \showarticletitle{Llama 2: Open foundation and fine-tuned chat models}.
\newblock \bibinfo{journal}{\emph{ArXiv preprint}}  \bibinfo{volume}{abs/2307.09288} (\bibinfo{year}{2023}).
\newblock
\urldef\tempurl%
\url{https://arxiv.org/abs/2307.09288}
\showURL{%
\tempurl}


\bibitem[Vaswani et~al\mbox{.}(2017)]%
        {Vaswani2017AttentionIA}
\bibfield{author}{\bibinfo{person}{Ashish Vaswani}, \bibinfo{person}{Noam Shazeer}, \bibinfo{person}{Niki Parmar}, \bibinfo{person}{Jakob Uszkoreit}, \bibinfo{person}{Llion Jones}, \bibinfo{person}{Aidan~N. Gomez}, \bibinfo{person}{Lukasz Kaiser}, {and} \bibinfo{person}{Illia Polosukhin}.} \bibinfo{year}{2017}\natexlab{}.
\newblock \showarticletitle{Attention is All you Need}. In \bibinfo{booktitle}{\emph{Advances in Neural Information Processing Systems 30: Annual Conference on Neural Information Processing Systems 2017, December 4-9, 2017, Long Beach, CA, {USA}}}, \bibfield{editor}{\bibinfo{person}{Isabelle Guyon}, \bibinfo{person}{Ulrike von Luxburg}, \bibinfo{person}{Samy Bengio}, \bibinfo{person}{Hanna~M. Wallach}, \bibinfo{person}{Rob Fergus}, \bibinfo{person}{S.~V.~N. Vishwanathan}, {and} \bibinfo{person}{Roman Garnett}} (Eds.). \bibinfo{pages}{5998--6008}.
\newblock
\urldef\tempurl%
\url{https://proceedings.neurips.cc/paper/2017/hash/3f5ee243547dee91fbd053c1c4a845aa-Abstract.html}
\showURL{%
\tempurl}


\bibitem[Wan et~al\mbox{.}(2023a)]%
        {Wan2023BetterZR}
\bibfield{author}{\bibinfo{person}{Xingchen Wan}, \bibinfo{person}{Ruoxi Sun}, \bibinfo{person}{Hanjun Dai}, \bibinfo{person}{Sercan Arik}, {and} \bibinfo{person}{Tomas Pfister}.} \bibinfo{year}{2023}\natexlab{a}.
\newblock \showarticletitle{Better Zero-Shot Reasoning with Self-Adaptive Prompting}. In \bibinfo{booktitle}{\emph{Findings of the Association for Computational Linguistics: ACL 2023}}, \bibfield{editor}{\bibinfo{person}{Anna Rogers}, \bibinfo{person}{Jordan Boyd-Graber}, {and} \bibinfo{person}{Naoaki Okazaki}} (Eds.). \bibinfo{publisher}{Association for Computational Linguistics}, \bibinfo{address}{Toronto, Canada}, \bibinfo{pages}{3493--3514}.
\newblock
\urldef\tempurl%
\url{https://doi.org/10.18653/v1/2023.findings-acl.216}
\showDOI{\tempurl}


\bibitem[Wan et~al\mbox{.}(2023b)]%
        {wan2023universal}
\bibfield{author}{\bibinfo{person}{Xingchen Wan}, \bibinfo{person}{Ruoxi Sun}, \bibinfo{person}{Hootan Nakhost}, \bibinfo{person}{Hanjun Dai}, \bibinfo{person}{Julian Eisenschlos}, \bibinfo{person}{Sercan Arik}, {and} \bibinfo{person}{Tomas Pfister}.} \bibinfo{year}{2023}\natexlab{b}.
\newblock \showarticletitle{Universal Self-Adaptive Prompting}. In \bibinfo{booktitle}{\emph{Proceedings of the 2023 Conference on Empirical Methods in Natural Language Processing}}, \bibfield{editor}{\bibinfo{person}{Houda Bouamor}, \bibinfo{person}{Juan Pino}, {and} \bibinfo{person}{Kalika Bali}} (Eds.). \bibinfo{publisher}{Association for Computational Linguistics}, \bibinfo{address}{Singapore}, \bibinfo{pages}{7437--7462}.
\newblock
\urldef\tempurl%
\url{https://doi.org/10.18653/v1/2023.emnlp-main.461}
\showDOI{\tempurl}


\bibitem[Wang et~al\mbox{.}(2024a)]%
        {wang2024one}
\bibfield{author}{\bibinfo{person}{Ruochen Wang}, \bibinfo{person}{Sohyun An}, \bibinfo{person}{Minhao Cheng}, \bibinfo{person}{Tianyi Zhou}, \bibinfo{person}{Sung~Ju Hwang}, {and} \bibinfo{person}{Cho-Jui Hsieh}.} \bibinfo{year}{2024}\natexlab{a}.
\newblock \showarticletitle{One Prompt is not Enough: Automated Construction of a Mixture-of-Expert Prompts}.
\newblock \bibinfo{journal}{\emph{ArXiv preprint}}  \bibinfo{volume}{abs/2407.00256} (\bibinfo{year}{2024}).
\newblock
\urldef\tempurl%
\url{https://arxiv.org/abs/2407.00256}
\showURL{%
\tempurl}


\bibitem[Wang et~al\mbox{.}(2024b)]%
        {Wang2024RDRecRD}
\bibfield{author}{\bibinfo{person}{Xinfeng Wang}, \bibinfo{person}{Jin Cui}, \bibinfo{person}{Yoshimi Suzuki}, {and} \bibinfo{person}{Fumiyo Fukumoto}.} \bibinfo{year}{2024}\natexlab{b}.
\newblock \showarticletitle{RDRec: Rationale Distillation for LLM-based Recommendation}. In \bibinfo{booktitle}{\emph{Annual Meeting of the Association for Computational Linguistics}}.
\newblock
\urldef\tempurl%
\url{https://api.semanticscholar.org/CorpusID:269899491}
\showURL{%
\tempurl}


\bibitem[Wang et~al\mbox{.}(2023c)]%
        {Wang2023PromptAgentSP}
\bibfield{author}{\bibinfo{person}{Xinyuan Wang}, \bibinfo{person}{Chenxi Li}, \bibinfo{person}{Zhen Wang}, \bibinfo{person}{Fan Bai}, \bibinfo{person}{Haotian Luo}, \bibinfo{person}{Jiayou Zhang}, \bibinfo{person}{Nebojsa Jojic}, \bibinfo{person}{Eric~P. Xing}, {and} \bibinfo{person}{Zhiting Hu}.} \bibinfo{year}{2023}\natexlab{c}.
\newblock \showarticletitle{PromptAgent: Strategic Planning with Language Models Enables Expert-level Prompt Optimization}.
\newblock \bibinfo{journal}{\emph{ArXiv preprint}}  \bibinfo{volume}{abs/2310.16427} (\bibinfo{year}{2023}).
\newblock
\urldef\tempurl%
\url{https://arxiv.org/abs/2310.16427}
\showURL{%
\tempurl}


\bibitem[Wang et~al\mbox{.}(2023d)]%
        {wang2022self2}
\bibfield{author}{\bibinfo{person}{Xuezhi Wang}, \bibinfo{person}{Jason Wei}, \bibinfo{person}{Dale Schuurmans}, \bibinfo{person}{Quoc~V. Le}, \bibinfo{person}{Ed~H. Chi}, \bibinfo{person}{Sharan Narang}, \bibinfo{person}{Aakanksha Chowdhery}, {and} \bibinfo{person}{Denny Zhou}.} \bibinfo{year}{2023}\natexlab{d}.
\newblock \showarticletitle{Self-Consistency Improves Chain of Thought Reasoning in Language Models}. In \bibinfo{booktitle}{\emph{The Eleventh International Conference on Learning Representations, {ICLR} 2023, Kigali, Rwanda, May 1-5, 2023}}. \bibinfo{publisher}{OpenReview.net}.
\newblock
\urldef\tempurl%
\url{https://openreview.net/pdf?id=1PL1NIMMrw}
\showURL{%
\tempurl}


\bibitem[Wang et~al\mbox{.}(2023b)]%
        {wang2022self}
\bibfield{author}{\bibinfo{person}{Yizhong Wang}, \bibinfo{person}{Yeganeh Kordi}, \bibinfo{person}{Swaroop Mishra}, \bibinfo{person}{Alisa Liu}, \bibinfo{person}{Noah~A. Smith}, \bibinfo{person}{Daniel Khashabi}, {and} \bibinfo{person}{Hannaneh Hajishirzi}.} \bibinfo{year}{2023}\natexlab{b}.
\newblock \showarticletitle{Self-Instruct: Aligning Language Models with Self-Generated Instructions}. In \bibinfo{booktitle}{\emph{Proceedings of the 61st Annual Meeting of the Association for Computational Linguistics (Volume 1: Long Papers)}}, \bibfield{editor}{\bibinfo{person}{Anna Rogers}, \bibinfo{person}{Jordan Boyd-Graber}, {and} \bibinfo{person}{Naoaki Okazaki}} (Eds.). \bibinfo{publisher}{Association for Computational Linguistics}, \bibinfo{address}{Toronto, Canada}, \bibinfo{pages}{13484--13508}.
\newblock
\urldef\tempurl%
\url{https://doi.org/10.18653/v1/2023.acl-long.754}
\showDOI{\tempurl}


\bibitem[Wang et~al\mbox{.}(2023a)]%
        {Wang2023LearningTF}
\bibfield{author}{\bibinfo{person}{Zhiruo Wang}, \bibinfo{person}{Jun Araki}, \bibinfo{person}{Zhengbao Jiang}, \bibinfo{person}{Md.~Rizwan Parvez}, {and} \bibinfo{person}{Graham Neubig}.} \bibinfo{year}{2023}\natexlab{a}.
\newblock \showarticletitle{Learning to Filter Context for Retrieval-Augmented Generation}.
\newblock \bibinfo{journal}{\emph{ArXiv preprint}}  \bibinfo{volume}{abs/2311.08377} (\bibinfo{year}{2023}).
\newblock
\urldef\tempurl%
\url{https://arxiv.org/abs/2311.08377}
\showURL{%
\tempurl}


\bibitem[Wei et~al\mbox{.}(2022a)]%
        {Wei2022EmergentAO}
\bibfield{author}{\bibinfo{person}{Jason Wei}, \bibinfo{person}{Yi Tay}, \bibinfo{person}{Rishi Bommasani}, \bibinfo{person}{Colin Raffel}, \bibinfo{person}{Barret Zoph}, \bibinfo{person}{Sebastian Borgeaud}, \bibinfo{person}{Dani Yogatama}, \bibinfo{person}{Maarten Bosma}, \bibinfo{person}{Denny Zhou}, \bibinfo{person}{Donald Metzler}, \bibinfo{person}{Ed~Huai hsin Chi}, \bibinfo{person}{Tatsunori Hashimoto}, \bibinfo{person}{Oriol Vinyals}, \bibinfo{person}{Percy Liang}, \bibinfo{person}{Jeff Dean}, {and} \bibinfo{person}{William Fedus}.} \bibinfo{year}{2022}\natexlab{a}.
\newblock \showarticletitle{Emergent Abilities of Large Language Models}.
\newblock \bibinfo{journal}{\emph{ArXiv preprint}}  \bibinfo{volume}{abs/2206.07682} (\bibinfo{year}{2022}).
\newblock
\urldef\tempurl%
\url{https://arxiv.org/abs/2206.07682}
\showURL{%
\tempurl}


\bibitem[Wei et~al\mbox{.}(2022b)]%
        {Wei2022ChainOT}
\bibfield{author}{\bibinfo{person}{Jason Wei}, \bibinfo{person}{Xuezhi Wang}, \bibinfo{person}{Dale Schuurmans}, \bibinfo{person}{Maarten Bosma}, \bibinfo{person}{Brian Ichter}, \bibinfo{person}{Fei Xia}, \bibinfo{person}{Ed~H. Chi}, \bibinfo{person}{Quoc~V. Le}, {and} \bibinfo{person}{Denny Zhou}.} \bibinfo{year}{2022}\natexlab{b}.
\newblock \showarticletitle{Chain-of-Thought Prompting Elicits Reasoning in Large Language Models}. In \bibinfo{booktitle}{\emph{Advances in Neural Information Processing Systems 35: Annual Conference on Neural Information Processing Systems 2022, NeurIPS 2022, New Orleans, LA, USA, November 28 - December 9, 2022}}, \bibfield{editor}{\bibinfo{person}{Sanmi Koyejo}, \bibinfo{person}{S.~Mohamed}, \bibinfo{person}{A.~Agarwal}, \bibinfo{person}{Danielle Belgrave}, \bibinfo{person}{K.~Cho}, {and} \bibinfo{person}{A.~Oh}} (Eds.).
\newblock
\urldef\tempurl%
\url{http://papers.nips.cc/paper\_files/paper/2022/hash/9d5609613524ecf4f15af0f7b31abca4-Abstract-Conference.html}
\showURL{%
\tempurl}


\bibitem[Wingate et~al\mbox{.}(2022)]%
        {Wingate2022PromptCA}
\bibfield{author}{\bibinfo{person}{David Wingate}, \bibinfo{person}{Mohammad Shoeybi}, {and} \bibinfo{person}{Taylor Sorensen}.} \bibinfo{year}{2022}\natexlab{}.
\newblock \showarticletitle{Prompt Compression and Contrastive Conditioning for Controllability and Toxicity Reduction in Language Models}. In \bibinfo{booktitle}{\emph{Findings of the Association for Computational Linguistics: EMNLP 2022}}, \bibfield{editor}{\bibinfo{person}{Yoav Goldberg}, \bibinfo{person}{Zornitsa Kozareva}, {and} \bibinfo{person}{Yue Zhang}} (Eds.). \bibinfo{publisher}{Association for Computational Linguistics}, \bibinfo{address}{Abu Dhabi, United Arab Emirates}, \bibinfo{pages}{5621--5634}.
\newblock
\urldef\tempurl%
\url{https://doi.org/10.18653/v1/2022.findings-emnlp.412}
\showDOI{\tempurl}


\bibitem[Xu et~al\mbox{.}(2023b)]%
        {Xu2023RECOMPIR}
\bibfield{author}{\bibinfo{person}{Fangyuan Xu}, \bibinfo{person}{Weijia Shi}, {and} \bibinfo{person}{Eunsol Choi}.} \bibinfo{year}{2023}\natexlab{b}.
\newblock \showarticletitle{RECOMP: Improving Retrieval-Augmented LMs with Compression and Selective Augmentation}.
\newblock \bibinfo{journal}{\emph{ArXiv preprint}}  \bibinfo{volume}{abs/2310.04408} (\bibinfo{year}{2023}).
\newblock
\urldef\tempurl%
\url{https://arxiv.org/abs/2310.04408}
\showURL{%
\tempurl}


\bibitem[Xu et~al\mbox{.}(2022)]%
        {xu2022gps}
\bibfield{author}{\bibinfo{person}{Hanwei Xu}, \bibinfo{person}{Yujun Chen}, \bibinfo{person}{Yulun Du}, \bibinfo{person}{Nan Shao}, \bibinfo{person}{Wang Yanggang}, \bibinfo{person}{Haiyu Li}, {and} \bibinfo{person}{Zhilin Yang}.} \bibinfo{year}{2022}\natexlab{}.
\newblock \showarticletitle{{GPS}: Genetic Prompt Search for Efficient Few-Shot Learning}. In \bibinfo{booktitle}{\emph{Proceedings of the 2022 Conference on Empirical Methods in Natural Language Processing}}, \bibfield{editor}{\bibinfo{person}{Yoav Goldberg}, \bibinfo{person}{Zornitsa Kozareva}, {and} \bibinfo{person}{Yue Zhang}} (Eds.). \bibinfo{publisher}{Association for Computational Linguistics}, \bibinfo{address}{Abu Dhabi, United Arab Emirates}, \bibinfo{pages}{8162--8171}.
\newblock
\urldef\tempurl%
\url{https://doi.org/10.18653/v1/2022.emnlp-main.559}
\showDOI{\tempurl}


\bibitem[Xu et~al\mbox{.}(2023a)]%
        {Xu2023RepromptingAC}
\bibfield{author}{\bibinfo{person}{Weijia Xu}, \bibinfo{person}{Andrzej Banburski-Fahey}, {and} \bibinfo{person}{Nebojsa Jojic}.} \bibinfo{year}{2023}\natexlab{a}.
\newblock \showarticletitle{Reprompting: Automated Chain-of-Thought Prompt Inference Through Gibbs Sampling}.
\newblock \bibinfo{journal}{\emph{ArXiv preprint}}  \bibinfo{volume}{abs/2305.09993} (\bibinfo{year}{2023}).
\newblock
\urldef\tempurl%
\url{https://arxiv.org/abs/2305.09993}
\showURL{%
\tempurl}


\bibitem[Yang et~al\mbox{.}(2023)]%
        {Yang2023LargeLM}
\bibfield{author}{\bibinfo{person}{Chengrun Yang}, \bibinfo{person}{Xuezhi Wang}, \bibinfo{person}{Yifeng Lu}, \bibinfo{person}{Hanxiao Liu}, \bibinfo{person}{Quoc~V. Le}, \bibinfo{person}{Denny Zhou}, {and} \bibinfo{person}{Xinyun Chen}.} \bibinfo{year}{2023}\natexlab{}.
\newblock \showarticletitle{Large Language Models as Optimizers}.
\newblock \bibinfo{journal}{\emph{ArXiv preprint}}  \bibinfo{volume}{abs/2309.03409} (\bibinfo{year}{2023}).
\newblock
\urldef\tempurl%
\url{https://arxiv.org/abs/2309.03409}
\showURL{%
\tempurl}


\bibitem[Yang et~al\mbox{.}(2024)]%
        {yang2024ampo}
\bibfield{author}{\bibinfo{person}{Sheng Yang}, \bibinfo{person}{Yurong Wu}, \bibinfo{person}{Yan Gao}, \bibinfo{person}{Zineng Zhou}, \bibinfo{person}{Bin~Benjamin Zhu}, \bibinfo{person}{Xiaodi Sun}, \bibinfo{person}{Jian-Guang Lou}, \bibinfo{person}{Zhiming Ding}, \bibinfo{person}{Anbang Hu}, \bibinfo{person}{Yuan Fang}, {et~al\mbox{.}}} \bibinfo{year}{2024}\natexlab{}.
\newblock \showarticletitle{AMPO: Automatic Multi-Branched Prompt Optimization}.
\newblock \bibinfo{journal}{\emph{ArXiv preprint}}  \bibinfo{volume}{abs/2410.08696} (\bibinfo{year}{2024}).
\newblock
\urldef\tempurl%
\url{https://arxiv.org/abs/2410.08696}
\showURL{%
\tempurl}


\bibitem[Yao et~al\mbox{.}(2023a)]%
        {Yao2023TreeOT}
\bibfield{author}{\bibinfo{person}{Shunyu Yao}, \bibinfo{person}{Dian Yu}, \bibinfo{person}{Jeffrey Zhao}, \bibinfo{person}{Izhak Shafran}, \bibinfo{person}{Tom Griffiths}, \bibinfo{person}{Yuan Cao}, {and} \bibinfo{person}{Karthik Narasimhan}.} \bibinfo{year}{2023}\natexlab{a}.
\newblock \showarticletitle{Tree of Thoughts: Deliberate Problem Solving with Large Language Models}. In \bibinfo{booktitle}{\emph{Advances in Neural Information Processing Systems 36: Annual Conference on Neural Information Processing Systems 2023, NeurIPS 2023, New Orleans, LA, USA, December 10 - 16, 2023}}, \bibfield{editor}{\bibinfo{person}{Alice Oh}, \bibinfo{person}{Tristan Naumann}, \bibinfo{person}{Amir Globerson}, \bibinfo{person}{Kate Saenko}, \bibinfo{person}{Moritz Hardt}, {and} \bibinfo{person}{Sergey Levine}} (Eds.).
\newblock
\urldef\tempurl%
\url{http://papers.nips.cc/paper\_files/paper/2023/hash/271db9922b8d1f4dd7aaef84ed5ac703-Abstract-Conference.html}
\showURL{%
\tempurl}


\bibitem[Yao et~al\mbox{.}(2023b)]%
        {Yao2022ReActSR}
\bibfield{author}{\bibinfo{person}{Shunyu Yao}, \bibinfo{person}{Jeffrey Zhao}, \bibinfo{person}{Dian Yu}, \bibinfo{person}{Nan Du}, \bibinfo{person}{Izhak Shafran}, \bibinfo{person}{Karthik~R. Narasimhan}, {and} \bibinfo{person}{Yuan Cao}.} \bibinfo{year}{2023}\natexlab{b}.
\newblock \showarticletitle{ReAct: Synergizing Reasoning and Acting in Language Models}. In \bibinfo{booktitle}{\emph{The Eleventh International Conference on Learning Representations, {ICLR} 2023, Kigali, Rwanda, May 1-5, 2023}}. \bibinfo{publisher}{OpenReview.net}.
\newblock
\urldef\tempurl%
\url{https://openreview.net/pdf?id=WE\_vluYUL-X}
\showURL{%
\tempurl}


\bibitem[Ye et~al\mbox{.}(2023)]%
        {ye2023prompt}
\bibfield{author}{\bibinfo{person}{Qinyuan Ye}, \bibinfo{person}{Maxamed Axmed}, \bibinfo{person}{Reid Pryzant}, {and} \bibinfo{person}{Fereshte Khani}.} \bibinfo{year}{2023}\natexlab{}.
\newblock \showarticletitle{Prompt engineering a prompt engineer}.
\newblock \bibinfo{journal}{\emph{ArXiv preprint}}  \bibinfo{volume}{abs/2311.05661} (\bibinfo{year}{2023}).
\newblock
\urldef\tempurl%
\url{https://arxiv.org/abs/2311.05661}
\showURL{%
\tempurl}


\bibitem[Yoon et~al\mbox{.}(2024)]%
        {Yoon2024CompActCR}
\bibfield{author}{\bibinfo{person}{Chanwoong Yoon}, \bibinfo{person}{Taewhoo Lee}, \bibinfo{person}{Hyeon Hwang}, \bibinfo{person}{Minbyul Jeong}, {and} \bibinfo{person}{Jaewoo Kang}.} \bibinfo{year}{2024}\natexlab{}.
\newblock \showarticletitle{CompAct: Compressing Retrieved Documents Actively for Question Answering}.
\newblock \bibinfo{journal}{\emph{ArXiv preprint}}  \bibinfo{volume}{abs/2407.09014} (\bibinfo{year}{2024}).
\newblock
\urldef\tempurl%
\url{https://arxiv.org/abs/2407.09014}
\showURL{%
\tempurl}


\bibitem[Zhang et~al\mbox{.}(2020)]%
        {Zhang2020FastIB}
\bibfield{author}{\bibinfo{person}{Biao Zhang}, \bibinfo{person}{Ivan Titov}, {and} \bibinfo{person}{Rico Sennrich}.} \bibinfo{year}{2020}\natexlab{}.
\newblock \showarticletitle{Fast Interleaved Bidirectional Sequence Generation}. In \bibinfo{booktitle}{\emph{Proceedings of the Fifth Conference on Machine Translation}}, \bibfield{editor}{\bibinfo{person}{Lo{\"\i}c Barrault}, \bibinfo{person}{Ond{\v{r}}ej Bojar}, \bibinfo{person}{Fethi Bougares}, \bibinfo{person}{Rajen Chatterjee}, \bibinfo{person}{Marta~R. Costa-juss{\`a}}, \bibinfo{person}{Christian Federmann}, \bibinfo{person}{Mark Fishel}, \bibinfo{person}{Alexander Fraser}, \bibinfo{person}{Yvette Graham}, \bibinfo{person}{Paco Guzman}, \bibinfo{person}{Barry Haddow}, \bibinfo{person}{Matthias Huck}, \bibinfo{person}{Antonio~Jimeno Yepes}, \bibinfo{person}{Philipp Koehn}, \bibinfo{person}{Andr{\'e} Martins}, \bibinfo{person}{Makoto Morishita}, \bibinfo{person}{Christof Monz}, \bibinfo{person}{Masaaki Nagata}, \bibinfo{person}{Toshiaki Nakazawa}, {and} \bibinfo{person}{Matteo Negri}} (Eds.). \bibinfo{publisher}{Association for Computational Linguistics}, \bibinfo{address}{Online},
  \bibinfo{pages}{503--515}.
\newblock
\urldef\tempurl%
\url{https://aclanthology.org/2020.wmt-1.62}
\showURL{%
\tempurl}


\bibitem[Zhang et~al\mbox{.}(2024b)]%
        {zhang2024prefer}
\bibfield{author}{\bibinfo{person}{Chenrui Zhang}, \bibinfo{person}{Lin Liu}, \bibinfo{person}{Chuyuan Wang}, \bibinfo{person}{Xiao Sun}, \bibinfo{person}{Hongyu Wang}, \bibinfo{person}{Jinpeng Wang}, {and} \bibinfo{person}{Mingchen Cai}.} \bibinfo{year}{2024}\natexlab{b}.
\newblock \showarticletitle{{PREFER:} Prompt Ensemble Learning via Feedback-Reflect-Refine}. In \bibinfo{booktitle}{\emph{Thirty-Eighth {AAAI} Conference on Artificial Intelligence, {AAAI} 2024, Thirty-Sixth Conference on Innovative Applications of Artificial Intelligence, {IAAI} 2024, Fourteenth Symposium on Educational Advances in Artificial Intelligence, {EAAI} 2014, February 20-27, 2024, Vancouver, Canada}}, \bibfield{editor}{\bibinfo{person}{Michael~J. Wooldridge}, \bibinfo{person}{Jennifer~G. Dy}, {and} \bibinfo{person}{Sriraam Natarajan}} (Eds.). \bibinfo{publisher}{{AAAI} Press}, \bibinfo{pages}{19525--19532}.
\newblock
\urldef\tempurl%
\url{https://doi.org/10.1609/AAAI.V38I17.29924}
\showDOI{\tempurl}


\bibitem[Zhang et~al\mbox{.}(2024a)]%
        {zhang2024sprig}
\bibfield{author}{\bibinfo{person}{Lechen Zhang}, \bibinfo{person}{Tolga Ergen}, \bibinfo{person}{Lajanugen Logeswaran}, \bibinfo{person}{Moontae Lee}, {and} \bibinfo{person}{David Jurgens}.} \bibinfo{year}{2024}\natexlab{a}.
\newblock \showarticletitle{SPRIG: Improving Large Language Model Performance by System Prompt Optimization}.
\newblock \bibinfo{journal}{\emph{ArXiv preprint}}  \bibinfo{volume}{abs/2410.14826} (\bibinfo{year}{2024}).
\newblock
\urldef\tempurl%
\url{https://arxiv.org/abs/2410.14826}
\showURL{%
\tempurl}


\bibitem[Zhang et~al\mbox{.}(2024c)]%
        {Zhang2024CompressingLC}
\bibfield{author}{\bibinfo{person}{Peitian Zhang}, \bibinfo{person}{Zheng Liu}, \bibinfo{person}{Shitao Xiao}, \bibinfo{person}{Ninglu Shao}, \bibinfo{person}{Qiwei Ye}, {and} \bibinfo{person}{Zhicheng Dou}.} \bibinfo{year}{2024}\natexlab{c}.
\newblock \showarticletitle{Compressing Lengthy Context With UltraGist}.
\newblock \bibinfo{journal}{\emph{ArXiv preprint}}  \bibinfo{volume}{abs/2405.16635} (\bibinfo{year}{2024}).
\newblock
\urldef\tempurl%
\url{https://arxiv.org/abs/2405.16635}
\showURL{%
\tempurl}


\bibitem[Zhang et~al\mbox{.}(2024e)]%
        {zhang2024adacomp}
\bibfield{author}{\bibinfo{person}{Qianchi Zhang}, \bibinfo{person}{Hainan Zhang}, \bibinfo{person}{Liang Pang}, \bibinfo{person}{Hongwei Zheng}, {and} \bibinfo{person}{Zhiming Zheng}.} \bibinfo{year}{2024}\natexlab{e}.
\newblock \showarticletitle{AdaComp: Extractive Context Compression with Adaptive Predictor for Retrieval-Augmented Large Language Models}.
\newblock \bibinfo{journal}{\emph{ArXiv preprint}}  \bibinfo{volume}{abs/2409.01579} (\bibinfo{year}{2024}).
\newblock
\urldef\tempurl%
\url{https://arxiv.org/abs/2409.01579}
\showURL{%
\tempurl}


\bibitem[Zhang et~al\mbox{.}(2022)]%
        {Zhang2022TEMPERATP}
\bibfield{author}{\bibinfo{person}{Tianjun Zhang}, \bibinfo{person}{Xuezhi Wang}, \bibinfo{person}{Denny Zhou}, \bibinfo{person}{Dale Schuurmans}, {and} \bibinfo{person}{Joseph~E. Gonzalez}.} \bibinfo{year}{2022}\natexlab{}.
\newblock \showarticletitle{TEMPERA: Test-Time Prompting via Reinforcement Learning}.
\newblock \bibinfo{journal}{\emph{ArXiv preprint}}  \bibinfo{volume}{abs/2211.11890} (\bibinfo{year}{2022}).
\newblock
\urldef\tempurl%
\url{https://arxiv.org/abs/2211.11890}
\showURL{%
\tempurl}


\bibitem[Zhang et~al\mbox{.}(2024d)]%
        {Zhang2024RevisitingOT}
\bibfield{author}{\bibinfo{person}{Tuo Zhang}, \bibinfo{person}{Jinyue Yuan}, {and} \bibinfo{person}{Amir~Salman Avestimehr}.} \bibinfo{year}{2024}\natexlab{d}.
\newblock \showarticletitle{Revisiting OPRO: The Limitations of Small-Scale LLMs as Optimizers}. In \bibinfo{booktitle}{\emph{Annual Meeting of the Association for Computational Linguistics}}.
\newblock
\urldef\tempurl%
\url{https://api.semanticscholar.org/CorpusID:269791383}
\showURL{%
\tempurl}


\bibitem[Zhang et~al\mbox{.}(2023)]%
        {Zhang2022AutomaticCO}
\bibfield{author}{\bibinfo{person}{Zhuosheng Zhang}, \bibinfo{person}{Aston Zhang}, \bibinfo{person}{Mu Li}, {and} \bibinfo{person}{Alex Smola}.} \bibinfo{year}{2023}\natexlab{}.
\newblock \showarticletitle{Automatic Chain of Thought Prompting in Large Language Models}. In \bibinfo{booktitle}{\emph{The Eleventh International Conference on Learning Representations, {ICLR} 2023, Kigali, Rwanda, May 1-5, 2023}}. \bibinfo{publisher}{OpenReview.net}.
\newblock
\urldef\tempurl%
\url{https://openreview.net/pdf?id=5NTt8GFjUHkr}
\showURL{%
\tempurl}


\bibitem[Zhao et~al\mbox{.}(2023)]%
        {zhao2023verify}
\bibfield{author}{\bibinfo{person}{Ruochen Zhao}, \bibinfo{person}{Xingxuan Li}, \bibinfo{person}{Shafiq Joty}, \bibinfo{person}{Chengwei Qin}, {and} \bibinfo{person}{Lidong Bing}.} \bibinfo{year}{2023}\natexlab{}.
\newblock \showarticletitle{Verify-and-Edit: A Knowledge-Enhanced Chain-of-Thought Framework}. In \bibinfo{booktitle}{\emph{Proceedings of the 61st Annual Meeting of the Association for Computational Linguistics (Volume 1: Long Papers)}}, \bibfield{editor}{\bibinfo{person}{Anna Rogers}, \bibinfo{person}{Jordan Boyd-Graber}, {and} \bibinfo{person}{Naoaki Okazaki}} (Eds.). \bibinfo{publisher}{Association for Computational Linguistics}, \bibinfo{address}{Toronto, Canada}, \bibinfo{pages}{5823--5840}.
\newblock
\urldef\tempurl%
\url{https://doi.org/10.18653/v1/2023.acl-long.320}
\showDOI{\tempurl}


\bibitem[Zhao et~al\mbox{.}(2021)]%
        {zhao2021calibrate}
\bibfield{author}{\bibinfo{person}{Zihao Zhao}, \bibinfo{person}{Eric Wallace}, \bibinfo{person}{Shi Feng}, \bibinfo{person}{Dan Klein}, {and} \bibinfo{person}{Sameer Singh}.} \bibinfo{year}{2021}\natexlab{}.
\newblock \showarticletitle{Calibrate Before Use: Improving Few-shot Performance of Language Models}. In \bibinfo{booktitle}{\emph{Proceedings of the 38th International Conference on Machine Learning, {ICML} 2021, 18-24 July 2021, Virtual Event}} \emph{(\bibinfo{series}{Proceedings of Machine Learning Research}, Vol.~\bibinfo{volume}{139})}, \bibfield{editor}{\bibinfo{person}{Marina Meila} {and} \bibinfo{person}{Tong Zhang}} (Eds.). \bibinfo{publisher}{{PMLR}}, \bibinfo{pages}{12697--12706}.
\newblock
\urldef\tempurl%
\url{http://proceedings.mlr.press/v139/zhao21c.html}
\showURL{%
\tempurl}


\bibitem[Zhou et~al\mbox{.}(2023a)]%
        {Zhou2023EfficientPV}
\bibfield{author}{\bibinfo{person}{Wangchunshu Zhou}, \bibinfo{person}{Yuchen Jiang}, \bibinfo{person}{Ryan Cotterell}, {and} \bibinfo{person}{Mrinmaya Sachan}.} \bibinfo{year}{2023}\natexlab{a}.
\newblock \showarticletitle{Efficient Prompting via Dynamic In-Context Learning}.
\newblock \bibinfo{journal}{\emph{ArXiv preprint}}  \bibinfo{volume}{abs/2305.11170} (\bibinfo{year}{2023}).
\newblock
\urldef\tempurl%
\url{https://arxiv.org/abs/2305.11170}
\showURL{%
\tempurl}


\bibitem[Zhou et~al\mbox{.}(2023b)]%
        {Zhou2022LargeLM}
\bibfield{author}{\bibinfo{person}{Yongchao Zhou}, \bibinfo{person}{Andrei~Ioan Muresanu}, \bibinfo{person}{Ziwen Han}, \bibinfo{person}{Keiran Paster}, \bibinfo{person}{Silviu Pitis}, \bibinfo{person}{Harris Chan}, {and} \bibinfo{person}{Jimmy Ba}.} \bibinfo{year}{2023}\natexlab{b}.
\newblock \showarticletitle{Large Language Models are Human-Level Prompt Engineers}. In \bibinfo{booktitle}{\emph{The Eleventh International Conference on Learning Representations, {ICLR} 2023, Kigali, Rwanda, May 1-5, 2023}}. \bibinfo{publisher}{OpenReview.net}.
\newblock
\urldef\tempurl%
\url{https://openreview.net/pdf?id=92gvk82DE-}
\showURL{%
\tempurl}


\bibitem[Zou et~al\mbox{.}(2023)]%
        {Zou2023MetaCoTGC}
\bibfield{author}{\bibinfo{person}{Anni Zou}, \bibinfo{person}{Zhuosheng Zhang}, \bibinfo{person}{Hai Zhao}, {and} \bibinfo{person}{Xiangru Tang}.} \bibinfo{year}{2023}\natexlab{}.
\newblock \showarticletitle{Meta-CoT: Generalizable Chain-of-Thought Prompting in Mixed-task Scenarios with Large Language Models}.
\newblock \bibinfo{journal}{\emph{ArXiv preprint}}  \bibinfo{volume}{abs/2310.06692} (\bibinfo{year}{2023}).
\newblock
\urldef\tempurl%
\url{https://arxiv.org/abs/2310.06692}
\showURL{%
\tempurl}


\end{thebibliography}
\appendix
\section{Appendix}
\subsection{Open Resources} \label{Open_Resources}
\begin{table}[!ht]
    \centering
    \caption{Open resources of efficient prompting methods in Prompt Compression}
    \label{Tab_open_resource2}
    \begin{tabularx}{\textwidth}{lX}
    \toprule
    \textbf{Method} &  \textbf{Link}\\
    \midrule
    \cellcolor{OOrange!30} Prompt Injection & \cellcolor{OOrange!30} \url{https://github.com/unbiarirang/Fixed-Input-Parameterization} \\
    \cellcolor{OOrange!30} Distilling Context & \cellcolor{OOrange!30} \url{https://en.wikipedia.org/wiki/Declarative_learning} \\
    \cellcolor{OOrange!30} Instrcution Distillation & \cellcolor{OOrange!30} 
    \url{www.github.com/sunnweiwei/RankGPT} \\
    \cellcolor{OOrange!30} xRAG & \cellcolor{OOrange!30} 
    \url{https://github.com/Hannibal046/xRAG} \\

    \cellcolor{OOrange!15} Gisting & \cellcolor{OOrange!15} \url{https://github.com/jayelm/gisting} \\
    \cellcolor{OOrange!15} UltraGist & \cellcolor{OOrange!15} \url{https://github.com/namespace-Pt/UltraGist} \\
    \cellcolor{OOrange!15} AutoCompressor & \cellcolor{OOrange!15} \url{https://github.com/princeton-nlp/AutoCompressors} \\
    \cellcolor{OOrange!15} LLoCO & \cellcolor{OOrange!15} \url{https://github.com/jeffreysijuntan/lloco} \\
    \cellcolor{OOrange!15} ICAE & \cellcolor{OOrange!15} \url{https://github.com/getao/icae} \\
    \cellcolor{OOrange!15} SelfCP & \cellcolor{OOrange!15} \url{https://github.com/jungao1106/SelfCP} \\
    \cellcolor{OOrange!15} 500xCompressor & \cellcolor{OOrange!15} \url{https://github.com/ZongqianLi/500xCompressor} \\
    \cellcolor{OOrange!15} POD & \cellcolor{OOrange!15} \url{https://github.com/lileipisces/POD} \\
    \cellcolor{OOrange!15} RDRec & \cellcolor{OOrange!15} \url{https://github.com/WangXFng/RDRec} \\
    \cellcolor{GGreen!30} FilCo & \cellcolor{GGreen!30} \url{https://github.com/zorazrw/filco} \\
    \cellcolor{GGreen!30} CPC & \cellcolor{GGreen!30} \url{https://github.com/Workday/cpc} \\
    \cellcolor{GGreen!30} AdaComp & \cellcolor{GGreen!30} \url{https://anonymous.4open.science/r/AdaComp-8C0C/} \\
    \cellcolor{GGreen!30} LLMLingua & \cellcolor{GGreen!30} \url{https://huggingface.co/spaces/microsoft/LLMLingua} \\
    \cellcolor{GGreen!30} LongLLMLingua & \cellcolor{GGreen!30} \url{https://aka.ms/LongLLMLingua} \\
    \cellcolor{GGreen!30} Selective Context & \cellcolor{GGreen!30} \url{https://github.com/liyucheng09/Selective_Context} \\
    \cellcolor{GGreen!30} PCRL & \cellcolor{GGreen!30} \url{https://github.com/nenomigami/PromptCompressor} \\
    \cellcolor{GGreen!30} LLMLingua-2 & \cellcolor{GGreen!30} \url{https://aka.ms/LLMLingua-2} \\

    \cellcolor{GGreen!10} RECOMP & \cellcolor{GGreen!10} \url{https://github.com/carriex/recomp} \\
    \cellcolor{GGreen!10} CompAct & \cellcolor{GGreen!10} \url{https://github.com/dmis-lab/CompAct} \\
    \bottomrule
    \end{tabularx}
\end{table}

\begin{table}[!ht]
    \centering
    \caption{Open resources of efficient prompting methods in Automatic Prompt Engineering}
    \label{open_resource1}
    \begin{tabularx}{\textwidth}{lX}
    \toprule
    \textbf{Method} &  \textbf{Link}\\
    \midrule
    \cellcolor{BBlue!40} APE  & \cellcolor{BBlue!40} \url{https://github.com/keirp/automatic_prompt_engineer} \\ 
    \cellcolor{BBlue!40} OPRO  & \cellcolor{BBlue!40} \url{https://github.com/google-deepmind/opro} \\
    \cellcolor{BBlue!40} PromptAgent & \cellcolor{BBlue!40} \url{https://github.com/XinyuanWangCS/PromptAgent} \\
    \cellcolor{BBlue!40} EvoPrompt & \cellcolor{BBlue!40} \url{https://github.com/beeevita/EvoPrompt} \\
    \cellcolor{BBlue!25} RLPrompt & \cellcolor{BBlue!25} \url{https://github.com/mingkaid/rl-prompt} \\
    \cellcolor{BBlue!25} DSP & \cellcolor{BBlue!25} \url{https://github.com/Leezekun/Directional-Stimulus-Prompting} \\
    \cellcolor{BBlue!25} ProTeGi & \cellcolor{BBlue!25} \url{https://github.com/microsoft/LMOps/tree/main/prompt_optimization} \\
    \cellcolor{BBlue!25} PE2 & \cellcolor{BBlue!25} \url{https://www.promptingguide.ai/introduction} \\
    \cellcolor{BBlue!25} PREFER & \cellcolor{BBlue!25} \url{https://github.com/zcrwind/PREFER} \\
    \cellcolor{BBlue!25} UniPrompt & \cellcolor{BBlue!25} \url{https://aka.ms/uniprompt} \\
    \cellcolor{BBlue!25} APOHF & \cellcolor{BBlue!25} \url{https://github.com/xqlin98/APOHF} \\
    \cellcolor{BBlue!25} BPO & \cellcolor{BBlue!25} \url{https://github.com/thu-coai/BPO} \\
    \cellcolor{BBlue!25} APEER & \cellcolor{BBlue!25} \url{https://github.com/jincan333/APEER} \\
    \cellcolor{BBlue!25} FIPO & \cellcolor{BBlue!25} \url{https://github.com/LuJunru/FIPO_Project} \\
    \cellcolor{BBlue!10} GrIPS & \cellcolor{BBlue!10} \url{https://github.com/archiki/GrIPS} \\
    \cellcolor{BBlue!10} Plum & \cellcolor{BBlue!10} \url{https://github.com/research4pan/Plum} \\
    \cellcolor{BBlue!10} TEMPERA & \cellcolor{BBlue!10} \url{https://github.com/tianjunz/TEMPERA} \\
    \cellcolor{YYellow!40} Auto-CoT & \cellcolor{YYellow!40} \url{https://github.com/amazon-research/auto-cot} \\
    \cellcolor{YYellow!40} Boosted Prompting & \cellcolor{YYellow!40} \url{https://github.com/awwang10/llmpromptboosting} \\
    \cellcolor{YYellow!25} Self-refine & \cellcolor{YYellow!25} \url{https://selfrefine.info/} \\
    \cellcolor{YYellow!25} PromptPG & \cellcolor{YYellow!25} \url{https://promptpg.github.io/}\\
    \cellcolor{YYellow!25} Prompt-OIRL & \cellcolor{YYellow!25} \url{ https://github.com/holarissun/Prompt-OIRL}\\
    \cellcolor{YYellow!25} Reflection & \cellcolor{YYellow!25} \url{https://github.com/noahshinn024/reflexion} \\
    \cellcolor{YYellow!25} PROMST & \cellcolor{YYellow!25} \url{https://yongchao98.github.io/MIT-REALM-PROMST/} \\
    \cellcolor{YYellow!10} ReAct & \cellcolor{YYellow!10} \url{https://react-lm.github.io/} \\
    \cellcolor{YYellow!10} Verify-and-Edit & \cellcolor{YYellow!10} \url{https://github.com/RuochenZhao/Verify-and-Edit} \\
    \cellcolor{YYellow!10} ART & \cellcolor{YYellow!10} \url{https://github.com/bhargaviparanjape/language-programmes/} \\
    \cellcolor{YYellow!10} Self-ask & \cellcolor{YYellow!10} \url{https://github.com/ofirpress/self-ask} \\
    \cellcolor{YYellow!10} ToolLLM & \cellcolor{YYellow!10} \url{https://github.com/OpenBMB/ToolBench} \\

    \bottomrule
    \end{tabularx}
\end{table}

\end{document}